\documentclass[10pt,twocolumn,letterpaper]{article}

\usepackage{wacv}
\makeatletter\let\suppmaketitle\@maketitle\makeatother

\usepackage[utf8]{inputenc}
\usepackage{amsmath}
\usepackage{graphicx}
\usepackage{booktabs}
\usepackage{multirow}
\usepackage{changepage}
\usepackage{makecell}
\usepackage{algorithm}
\usepackage{algorithmic}
\usepackage[table]{xcolor}
\usepackage{colortbl}
\usepackage{amsmath,amssymb}
\usepackage{enumitem}
\usepackage{dblfloatfix}
\usepackage{tikz}
\usepackage{array}
\usepackage{wrapfig}
\usepackage{microtype}
\usepackage{tabularx}
\newcolumntype{Y}{>{\centering\arraybackslash}X}

\definecolor{wacvblue}{rgb}{0.21,0.49,0.74}
\usepackage[pagebackref,breaklinks,colorlinks,allcolors=wacvblue]{hyperref}

\def\wacvPaperID{3550}
\def\confName{WACV}
\def\confYear{2027}

\title{Gaussians Can Subtract: Neural Dissolve Fields for Hybrid Scene Rendering}

\author{
Yiyan Wang$^{1,*}$ \qquad
Haohan Liu$^{2,*}$ \qquad
Tetsuo Shibuya$^{1}$\\
$^{1}$The University of Tokyo, Japan \qquad
$^{2}$Fuzhou University, China\qquad
$^{*}$Equal contribution \\
{\tt\small
5770716834@utac.u-tokyo.ac.jp \quad 852203208@fzu.edu.cn
\quad tshibuya@hgc.jp}}

\setlist[itemize]{leftmargin=1.1em,topsep=1pt,itemsep=1pt,parsep=0pt,partopsep=0pt}

\definecolor{best}{HTML}{FFD1CC}   %
\definecolor{second}{HTML}{FFE6CC} %
\definecolor{third}{HTML}{FFF5CC}  %
\newcommand{\gaussolve}{Gaussolve}
\newcommand{\ndf}{Neural Dissolve Field}

\newcommand{\placeholderfig}[2]{%
  \begin{figure}[t]
  \centering
  \fbox{\parbox[c][0.17\textheight][c]{0.92\linewidth}{\centering \small Missing original figure/table fragment.\\#2}}
  \caption{Placeholder for a missing original figure/table fragment. Replace by copying the corresponding original fragment into the project.}
  \end{figure}%
}
\newcommand{\maybeinput}[2]{%
  \IfFileExists{#1.tex}{\input{#1}}{\placeholderfig{#1}{#2}}%
}
\definecolor{lightcyanborder}{RGB}{120,210,235}
\newlength{\meshcellw}

\begin{document}
\maketitle

\begin{figure*}[t]
    \centering
    \includegraphics[width=\textwidth]{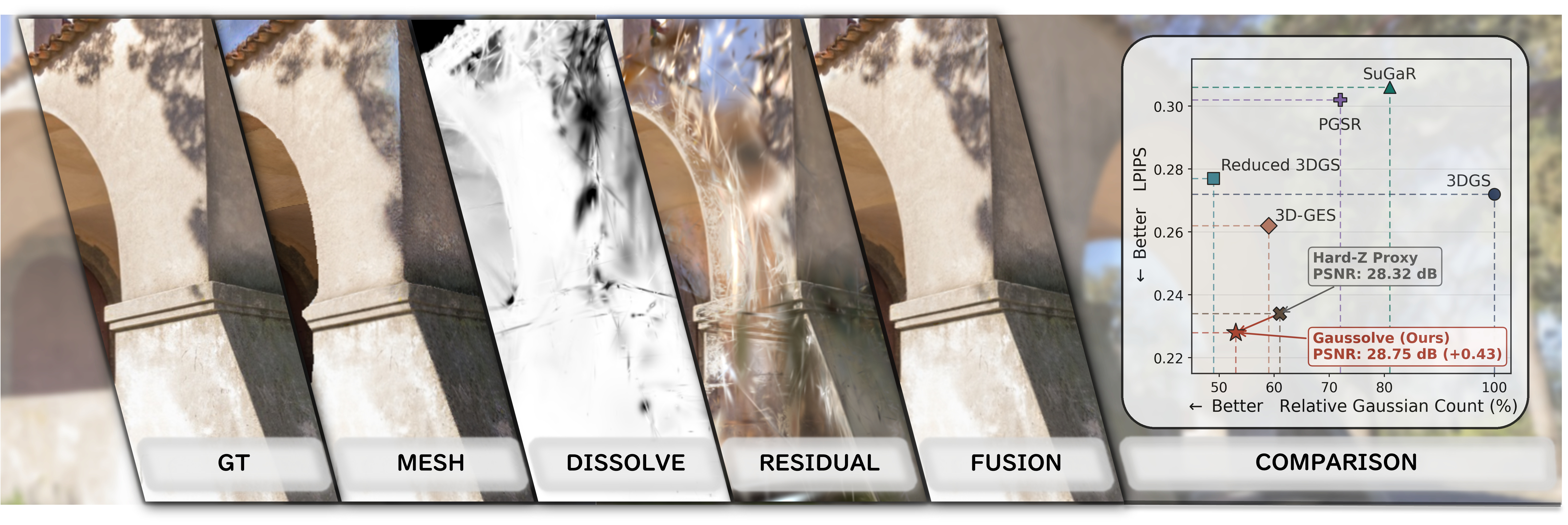}
    \caption{
    \textbf{Gaussians Can Subtract.}
    Existing hybrids primarily compensate for mesh errors by adding Gaussian residuals. \textbf{Gaussolve} instead introduces a subtractive Neural Dissolve Field that selectively dissolves inconsistent mesh content and lets residual Gaussians refill the revealed regions.
    This explicit subtract-and-refill decomposition improves visual quality while retaining a compact hybrid representation.
    Compared with existing methods, Gaussolve achieves a more favorable quality--compactness trade-off by allowing meshes and Gaussians to cooperate rather than compete.
    }
    \label{fig:teaser}
\end{figure*}

\begin{abstract}
Hybrid Mesh--Gaussian rendering combines compact textured surfaces with flexible 3D Gaussian primitives, offering an efficient alternative to purely Gaussian scene representations for structured environments. Existing scene-reconstruction hybrids commonly follow an additive formulation, where Gaussian residuals compensate for errors left by the mesh. However, not every Mesh--Gaussian conflict is caused by missing content: residual geometry errors, occlusion mismatch, texture-projection artifacts, and view-dependent appearance can make the same mesh content consistent in one view but inconsistent in another. We therefore formulate hybrid rendering as a multi-view inconsistency problem and present \gaussolve, a subtract-and-refill framework that selectively dissolves inconsistent mesh content before residual Gaussians refill the revealed regions. We introduce a lightweight \ndf{}, parameterized by scalar spherical-harmonic Gaussians, to learn where and from which viewing directions mesh content should be dissolved. A continuous transmittance fusion rule progressively exposes residual Gaussians behind the dissolved mesh, preserving their reconstruction gradients and avoiding hard visibility discontinuities. Experiments on structured and semi-structured scenes demonstrate higher visual fidelity with fewer Gaussian primitives and faster rendering. Project page: \href{https://ooosn.github.io/gaussians-can-subtract/}{https://ooosn.github.io/gaussians-can-subtract/}
\end{abstract}

\section{Introduction}
\label{sec:intro}

3D Gaussian Splatting (3DGS) has become a strong representation for real-time novel-view synthesis due to explicit primitives and efficient differentiable rasterization~\cite{kerbl2023gaussian}. However, practical environments contain broad regular structures, such as road surfaces with lane markings and large architectural surfaces~\cite{huang2025hybridmeshgaussian}. Representing these large regular regions purely with Gaussian primitives can require a large primitive budget, increasing storage and rasterization cost~\cite{huang2025hybridmeshgaussian,lee2024compact3dgs,fan2024lightgaussian}. Textured meshes provide a complementary representation: they compactly anchor broad regular surfaces, while Gaussians remain effective for thin structures, local residual geometry, and view-dependent appearance~\cite{kerbl2023gaussian,choi2024meshgs, huang2025hybridmeshgaussian}. However, existing hybrid scene-reconstruction pipelines still inherit limitations in both how the mesh is constructed and how the two representations are combined.

First, we revisit how the textured mesh is constructed. Existing methods typically extract the mesh from a previously optimized NeRF or Gaussian representation and then bake the reconstructed appearance onto the mesh~\cite{choi2024meshgs,huang2025hybridmeshgaussian}, making the resulting texture inherit the constraints of such reconstruction. In particular, enforcing cleaner and more regular geometry can reduce the flexibility of these representations to fit the observed appearance, leading to a trade-off between geometric alignment and rendering quality~\cite{choi2024meshgs,held2026meshsplatting,zhou2025mgsr}. Instead, we reconstruct the geometry independently and project the training images directly onto the mesh, thereby preserving the observed appearance when the geometry is reliable.

More importantly, we revisit how the mesh and Gaussians are combined. Even with such a textured mesh, the fundamental Mesh--Gaussian conflict remains. Conceptually, existing hybrids largely follow an additive formulation: the mesh accounts for the regular scene content, while Gaussians compensate for what the mesh fails to represent. But not every Mesh--Gaussian conflict is caused by missing content. Such conflicts may arise from residual geometry errors, occlusion mismatch, texture-projection artifacts, or view-dependent appearance, and importantly, the same mesh content can agree with one view while becoming inconsistent with another. We therefore reinterpret the remaining Mesh--Gaussian conflict as a multi-view inconsistency problem, which requires selectively suppressing inconsistent mesh content rather than only adding Gaussian residuals.

This reformulation exposes a fundamental mismatch between the problem and existing conflict-handling strategies. Existing mesh--Gaussian hybrids often resolve representation conflicts through hard visibility or assignment decisions. For example, prior methods remove inaccurate mesh regions to make room for Gaussians~\cite{huang2025hybridmeshgaussian}, enforce hard assignments between mesh and Gaussian representations~\cite{kim2024integrating}, or discard Gaussians occluded by an opaque mesh surface~\cite{choi2024meshgs,kim2024integrating}. Although these designs differ in implementation, they reduce the conflict to fixed removal, hard assignment, or binary visibility, none of which captures inconsistencies that vary across viewpoints. In particular, hard visibility allows the mesh to completely suppress residual Gaussians behind it. When the mesh conflicts with the current observation, this suppression also blocks the reconstruction gradients required for the Gaussian branch to correct the inconsistency. Moreover, small depth changes can abruptly switch the mesh--Gaussian visibility state.

We propose \gaussolve, a hybrid Mesh--Gaussian framework that resolves these inconsistencies through a subtract-and-refill process (Fig.~\ref{fig:teaser}). Instead of treating the mesh as a fixed explanation and only adding Gaussian residuals around it, \gaussolve \ allows inconsistent mesh content to be selectively dissolved before residual Gaussians refill the revealed regions. To this end, we introduce a \ndf{} parameterized by scalar spherical-harmonic Gaussians, which learns where and from which viewing directions the mesh should be attenuated under the visual reconstruction objective. We further integrate mesh dissolution and Gaussian accumulation through continuous transmittance-based fusion, so that suppressing the mesh progressively exposes the residual Gaussians behind it rather than abruptly switching visibility. This preserves their reconstruction gradients and allows them to correct the inconsistencies exposed by mesh dissolution, while mesh regions that remain consistent with the current view are preserved.

Our contributions are summarized as follows:
\begin{itemize}
\item We reformulate hybrid Mesh--Gaussian rendering as a multi-view inconsistency problem, moving beyond the purely additive formulation by selectively subtracting inconsistent mesh content before Gaussian refill.
\item We introduce a view-dependent \ndf{} that learns where and from which viewing directions mesh content should be dissolved, while preserving it in views where it remains consistent.
\item We propose a continuous transmittance fusion rule that avoids hard visibility discontinuities, preserves reconstruction gradients for residual Gaussians, and enables them to refill the content revealed by mesh dissolution.
\item We introduce a progressive optimization curriculum with
loss-driven dual densification, further improving optimization
stability and performance.
\end{itemize}

\begin{figure*}[t]
  \centering
  \IfFileExists{images/pipeline.pdf}{%
    \includegraphics[width=\linewidth]{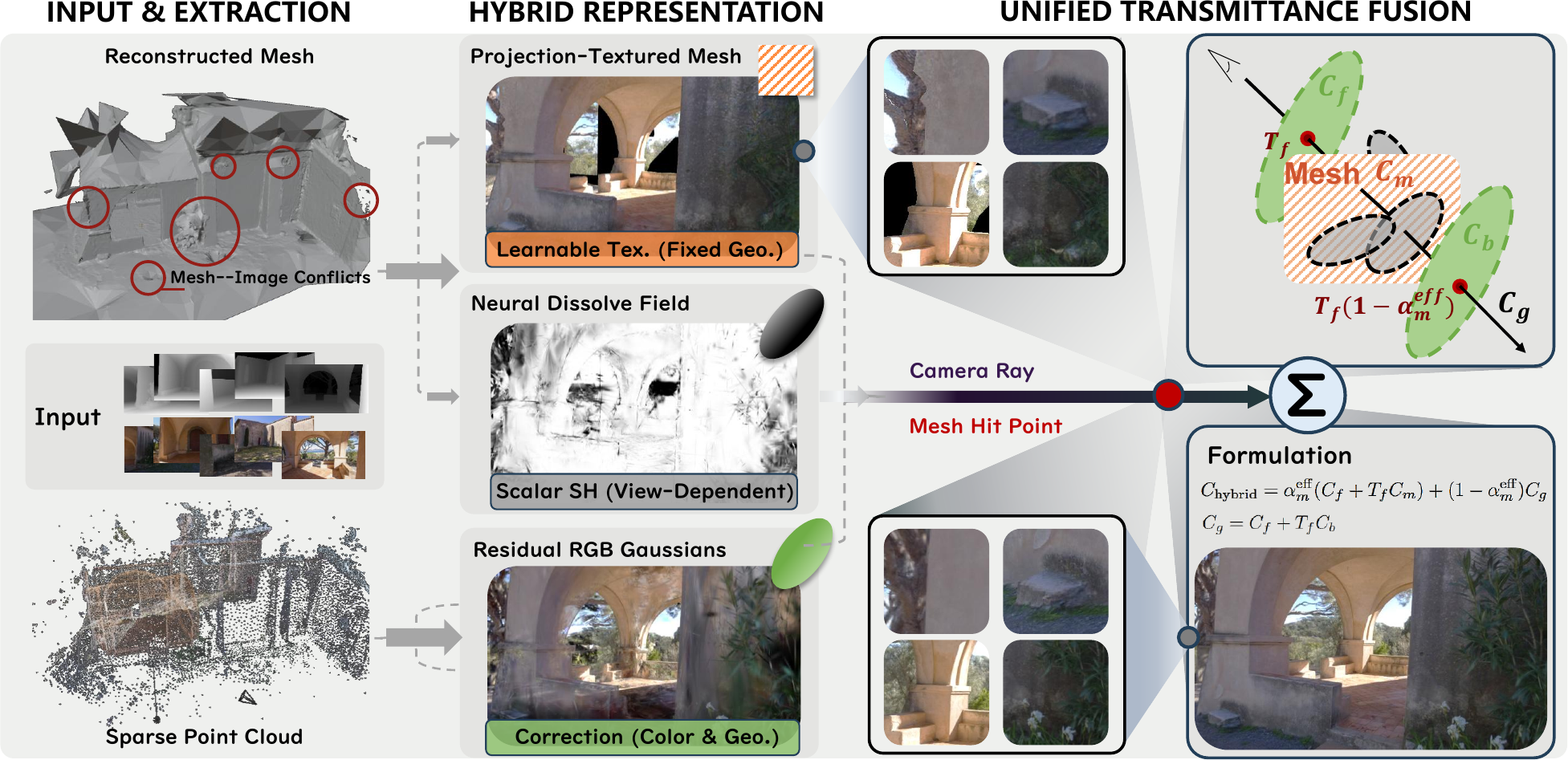}
  }{%
    \fbox{\parbox[c][0.20\textheight][c]{0.92\linewidth}{\centering Missing \texttt{images/pipeline.png}}}
  }
\caption{\textbf{Overview of \gaussolve.}
The projection-textured mesh represents broad regular content, while the Neural Dissolve Field selectively suppresses mesh content that becomes inconsistent across views. Residual Gaussians refill the revealed regions, and all components are combined through continuous transmittance fusion.}
  \label{fig:pipeline}
\end{figure*}

\section{Related Work}

\subsection{3D Gaussian Splatting}
3D Gaussian Splatting (3DGS)~\cite{kerbl2023gaussian} represents scenes using explicit 3D Gaussian primitives and achieves efficient optimization and rendering via differentiable rasterization, offering a strong quality--speed trade-off for novel-view synthesis. Subsequent works improve the Gaussian-only pipeline through densification, pruning, sorting, compression, or reduced memory footprints (e.g., StopThePop~\cite{radl2024stopthepop}, Revising Densification~\cite{rotabulo2024revising}, Taming 3DGS~\cite{mallick2024taming}, LightGaussian~\cite{fan2024lightgaussian}, Mini-Splatting~\cite{fang2024minisplatting}, Scaffold-GS~\cite{lu2024scaffoldgs}, Octree-GS~\cite{ren2024octreegs}, and Reduced 3DGS~\cite{papantonakis2024reduced3dgs}), and recent surveys summarize these branches~\cite{bao2024survey3dgs}.

\subsection{Mesh Priors and Surface Reconstruction}
SfM--MVS pipelines such as COLMAP~\cite{schoenberger2016sfm,schoenberger2016mvs}, as well as LiDAR and photogrammetry workflows, are widely used to produce point clouds, depth maps, and mesh priors for downstream rendering, editing, and interaction. Surface reconstruction from these inputs (e.g., Poisson reconstruction~\cite{kazhdan2006poisson}) remains standard engineering practice, but real-world capture conditions (noise, occlusions, non-uniform sampling) inevitably introduce local geometric errors and topological inconsistencies. As a result, reconstructed meshes often contain holes, self-intersections, or non-manifold connections, which mesh-repair methods have long targeted~\cite {Attene2010LightweightMeshRepair}. Such mesh priors can still disagree with image observations due to
geometry errors, occlusion mismatch, or appearance inconsistencies. We therefore treat the mesh as a useful structural prior but not a perfect and strict occluder.

\subsection{Hybrid Mesh--Gaussian Rendering}
Hybrid representations typically employ a surface mesh to anchor large-scale structures and use Gaussians or surfels to capture thin geometry and appearance residuals. Such designs have been explored for general scene rendering (e.g., MeshGS~\cite{choi2024meshgs}, CityGo~\cite{liu2025citygo}, HybridMeshGS~\cite{huang2025hybridmeshgaussian}, MeshSplats~\cite{tobiasz2025meshsplats}, OMeGa~\cite{cao2025omega}, UniMGS~\cite{xiao2026unimgs}, and When Gaussian Meets Surfel~\cite{ye2025whengaussianmeetssurfel}) and for controllable avatars and heads (e.g., SplattingAvatar~\cite{shao2024splattingavatar}, GaussianAvatars~\cite{qian2024gaussianavatars}, HAHA~\cite{svitov2024haha}, MeGA~\cite{wang2025mega}, HERA~\cite{cai2025hera}, Gaussian Pixel Codec Avatars~\cite{gupta2025gpica}, and relightable Gaussian avatars~\cite{saito2023relightablegca,wang2025relightablefullbody,zhan2024interactive}). Related works such as SuGaR~\cite{guedon2024sugar}, MILo~\cite{guedon2025milo}, and GaMeS~\cite{waczynska2024games} further demonstrate the benefits of explicit surfaces for geometric stability and mesh-in-the-loop optimization. However, many existing scene-reconstruction hybrids rely on two-pass rendering, hard depth-based splitting~\cite{liu2025citygo}, or heuristically pruned hard occluders~\cite{huang2025hybridmeshgaussian}, which can suppress residual Gaussians, block their reconstruction gradients, or abruptly switch the mesh--Gaussian visibility state. Gaussolve instead formulates visibility fusion as a single continuous transmittance accumulation with a volumetric Neural Dissolve Field, enabling continuous fusion and selective suppression of inconsistent mesh content.

\section{Method}
\label{sec:method}
As shown in Fig.~\ref{fig:pipeline}, given posed multi-view images and a textured mesh prior, \gaussolve{} constructs a hybrid representation with three branches: a textured mesh branch, residual RGB Gaussians, and a Neural Dissolve Field (NDF). The mesh branch provides compact structural and texture evidence for broad regular regions, while residual RGB Gaussians model content that cannot be consistently represented by the mesh, including fine details and view-dependent appearance. Instead of treating the mesh as a hard occluder, \gaussolve{} uses the NDF to selectively dissolve inconsistent mesh content and continuously exposes residual Gaussians through transmittance-based fusion.

\subsection{Textured Mesh Prior}
\label{sec:textured_mesh_prior}

\begin{figure}[t]
  \centering
  \includegraphics[width=\columnwidth]{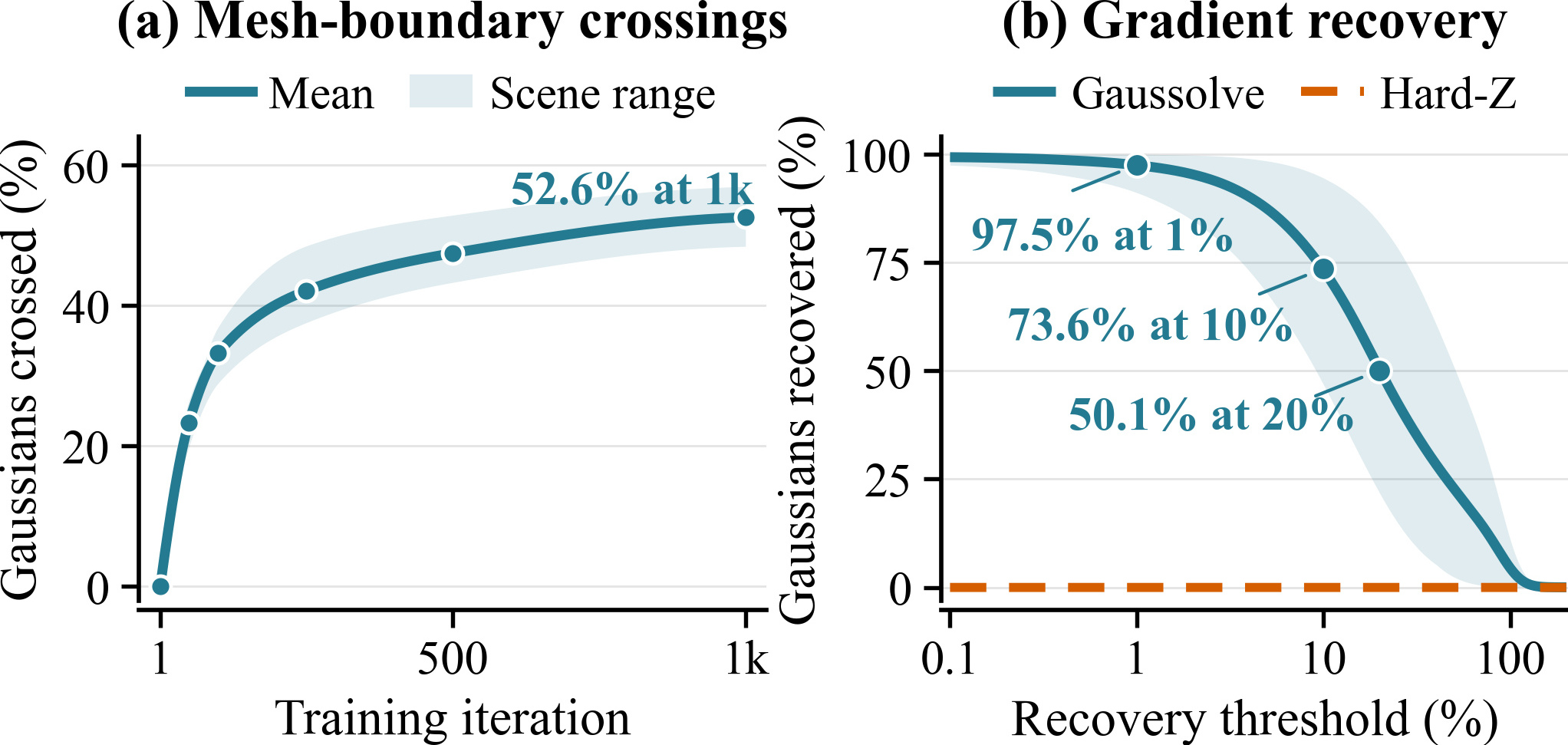}
\caption{\textbf{Hard-Z brittleness and gradient recovery.}
(a) Gaussians initially in front of the mesh frequently cross behind the Hard-Z visibility boundary during training.
(b) For Gaussians whose reconstruction gradients are blocked by Hard-Z, Gaussolve recovers a substantial fraction of the ungated reference gradient.
Curves and shading denote the four-scene mean and range.}
  \label{fig:hardz-gradient-evidence}
\end{figure}

\gaussolve{} uses a textured mesh prior $\mathcal{M}$ as the mesh branch. We reconstruct the geometry with RealityCapture (RC)~\cite{realitycapture} and directly project the training images onto the mesh to obtain its texture. Compared with baking reconstructed appearance from NeRF or Gaussian representations~\cite{choi2024meshgs,huang2025hybridmeshgaussian}, direct projection better preserves observed high-frequency details, as shown in Fig.~\ref{fig:mesh_prior_sanity}. To avoid test-view leakage, only training views are used for both MVS reconstruction and texturing. Unless otherwise stated, the default meshes are uniformly decimated to 100k faces. The resulting projected texture map defines the mesh color $C_m$ used during fusion.

\subsection{Continuous Transmittance Fusion}
\label{sec:continuous_fusion}

Hard depth splitting treats the mesh as a binary occluder, removing Gaussian contributions behind it. This cannot resolve multi-view mesh inconsistencies and also blocks corrective gradients to occluded residual Gaussians. Fig.~\ref{fig:hardz-gradient-evidence} quantifies both the frequent visibility changes and the resulting gradient starvation. We therefore replace hard visibility with a continuous fusion rule.

For exposition, we describe the fusion rule using a ray-wise front/back decomposition and omit repeated pixel and view arguments. Let $(C_f,T_f)$ denote the accumulated Gaussian color and transmittance before the mesh depth along a ray, and let $C_b$ denote the accumulated Gaussian color behind the mesh depth, excluding the front transmittance. The full Gaussian rendering is
\begin{equation}
C_g = C_f + T_f C_b.
\label{eq:decomposition}
\end{equation}
A direct soft fusion between the mesh color $C_m$ and the Gaussian back segment is
\begin{equation}
C_{\mathrm{hybrid}} = C_f + T_f\left(\alpha_m^{\mathrm{eff}} C_m + (1-\alpha_m^{\mathrm{eff}})C_b\right),
\label{eq:naive_fusion}
\end{equation}
where $\alpha_m^{\mathrm{eff}}\in[0,1]$ is the effective mesh opacity. Rather than explicitly recovering $C_b$ through division by $T_f$, we substitute $T_fC_b=C_g-C_f$ and obtain the division-free form
\begin{equation}
C_{\mathrm{hybrid}} = \alpha_m^{\mathrm{eff}}(C_f+T_fC_m) + (1-\alpha_m^{\mathrm{eff}})C_g.
\label{eq:continuous_fusion}
\end{equation}
Thus $\alpha_m^{\mathrm{eff}}=1$ preserves the mesh evidence, while $\alpha_m^{\mathrm{eff}}=0$ falls back to Gaussian rendering. In practice, the rasterized mesh depth is used only to compute the front/back Gaussian partition; dissolving the mesh changes only the final blending weight, not the mesh geometry. 

\subsection{Neural Dissolve Field}
\label{sec:neural_dissolve_field}

\begin{table*}[t]
\centering
\scriptsize
\setlength{\tabcolsep}{2.1pt}
\renewcommand{\arraystretch}{1.08}
\caption{\textbf{Main comparison on structured and semi-structured scenes.}
The target regime contains broad regular structures where a fixed-budget textured mesh can replace redundant Gaussian primitives.
\textbf{Bold} indicates the best performance, and cell colors highlight the \colorbox{best}{top}, \colorbox{second}{second-best}
, and \colorbox{third}{third-best} results
.}
\label{tab:main_results}

\begin{tabularx}{\textwidth}{@{}p{2.3cm}|cc|YYYYY|YYYYY@{}}
\toprule
\multirow{2}{*}{Method}
& \multirow{2}{*}{Mesh}
& \multirow{2}{*}{Hybrid}
& \multicolumn{5}{c|}{ML-Hypersim-AI (Structured)}
& \multicolumn{5}{c}{Deep Blending (Semi-Structured)} \\
\cmidrule(lr){4-8}\cmidrule(lr){9-13}
& &
& PSNR$\uparrow$ & SSIM$\uparrow$ & LPIPS$\downarrow$ & $\Delta$GS$\uparrow$ & Aux.\%
& PSNR$\uparrow$ & SSIM$\uparrow$ & LPIPS$\downarrow$ & $\Delta$GS$\uparrow$ & Aux.\% \\
\midrule

3DGS
& $\times$ & $\times$
& 34.39 & 0.959 & 0.155 & 0\% & --
& 28.50 & \cellcolor{second}0.877 & 0.272 & 0\% & -- \\

Reduced 3DGS
& $\times$ & $\times$
& 34.04 & 0.957 & 0.153 & \cellcolor{third}62\% & --
& 28.43 & 0.875 & 0.277 & \cellcolor{second}51\% & -- \\

3DGS-MCMC
& $\times$ & $\times$
& \cellcolor{second}35.82 & \cellcolor{third}0.960 & 0.140 & -1\% & --
& 28.56 & 0.875 & 0.268 & -12\% & -- \\

\midrule

SuGaR
& $\checkmark$ & $\times$
& 32.29 & 0.950 & 0.149 & 33\% & --
& 27.58 & 0.858 & 0.306 & 19\% & -- \\

PGSR
& $\checkmark$ & $\times$
& 32.95 & 0.958 & 0.147 & 61\% & --
& 28.44 & 0.863 & 0.302 & 28\% & -- \\

MeshSplatting
& $\checkmark$ & $\times$
& 28.47 & 0.903 & 0.236 & -- & --
& 24.71 & 0.809 & 0.359 & -- & -- \\

\midrule

3D-GES
& $\times$ & $\checkmark$
& 34.87 & \cellcolor{second}0.961 & 0.130 & \cellcolor{second}63\% & 13\%
& 28.42 & 0.875 & 0.262 & 41\% & 17\% \\

\midrule

Hard-Z Proxy
& $\checkmark$ & $\checkmark$
& 34.31 & 0.958 & 0.119 & 27\% & --
& 28.32 & 0.872 & 0.234 & 39\% & -- \\

\rowcolor{gray!10}
\textbf{Ours (3DGS)}
& $\checkmark$ & $\checkmark$
& \cellcolor{third}35.35 & \cellcolor{best}\textbf{0.962} & \cellcolor{best}\textbf{0.108} & 30\% & 7\%
& \cellcolor{second}28.75 & \cellcolor{third}0.876 & \cellcolor{second}0.228 & \cellcolor{third}47\% & 8\% \\

\rowcolor{gray!10}
\textbf{Ours (Reduced 3DGS)}
& $\checkmark$ & $\checkmark$
& 35.29 & 0.959 & \cellcolor{third}0.113 & \cellcolor{best}\textbf{67\%} & 8\%
& \cellcolor{third}28.67 & \cellcolor{second}0.877 & \cellcolor{third}0.232 & \cellcolor{best}\textbf{65\%} & 13\% \\

\rowcolor{gray!10}
\textbf{Ours (3DGS-MCMC)}
& $\checkmark$ & $\checkmark$
& \cellcolor{best}\textbf{36.89} & \cellcolor{second}0.961 & \cellcolor{second}0.110 & -6\% & 4\%
& \cellcolor{best}\textbf{28.93} & \cellcolor{best}\textbf{0.880} & \cellcolor{best}\textbf{0.227} & -3\% & 4\% \\

\bottomrule
\end{tabularx}
\end{table*}

\begin{table*}[t]
\centering
\scriptsize
\setlength{\tabcolsep}{2.1pt}
\renewcommand{\arraystretch}{1.08}
\caption{\textbf{Comparison with MeshGS~\cite{choi2024meshgs} and Hybrid Mesh-Gaussian~\cite{huang2025hybridmeshgaussian} on the Deep Blending dataset.}
For the baselines, we report the \textbf{3DGS-based} results directly from the respective papers.
}
\label{tab:hybrid_comparison}

\begin{tabularx}{\textwidth}{@{}p{2.3cm}|cc|YYYY|YYYY|YYYY@{}}
\toprule
\multirow{2}{*}{\makecell[l]{Method}}
& \multirow{2}{*}{Mesh}
& \multirow{2}{*}{Hybrid}
& \multicolumn{4}{c|}{DrJohnson}
& \multicolumn{4}{c|}{Playroom}
& \multicolumn{4}{c}{Average} \\
\cmidrule(lr){4-7}
\cmidrule(lr){8-11}
\cmidrule(lr){12-15}
& &
& PSNR$\uparrow$ & SSIM$\uparrow$ & LPIPS$\downarrow$ & $\Delta$GS$\uparrow$
& PSNR$\uparrow$ & SSIM$\uparrow$ & LPIPS$\downarrow$ & $\Delta$GS$\uparrow$
& PSNR$\uparrow$ & SSIM$\uparrow$ & LPIPS$\downarrow$ & $\Delta$GS$\uparrow$ \\
\midrule

MeshGS
& $\checkmark$ & $\checkmark$
& 28.80 & \cellcolor{best}\textbf{0.900} & \cellcolor{second}0.240 & --
& \cellcolor{second}30.50 & \cellcolor{second}0.900 & \cellcolor{second}0.250 & --
& \cellcolor{second}29.65 & \cellcolor{best}\textbf{0.900} & \cellcolor{second}0.245 & $\sim$30\% \\

Hybrid Mesh-Gaussian
& $\checkmark$ & $\checkmark$
& \cellcolor{second}29.01 & 0.888 & 0.262 & \cellcolor{second}43\%
& 30.11 & \cellcolor{best}\textbf{0.901} & 0.255 & \cellcolor{second}30\%
& 29.56 & 0.895 & 0.259 & \cellcolor{second}36\% \\

\midrule

\rowcolor{gray!10}
\textbf{Ours}
& $\checkmark$ & $\checkmark$
& \cellcolor{best}\textbf{29.69} & \cellcolor{second}0.894 & \cellcolor{best}\textbf{0.240} & \cellcolor{best}\textbf{58\%}
& \cellcolor{best}\textbf{30.93} & 0.893 & \cellcolor{best}\textbf{0.223} & \cellcolor{best}\textbf{61\%}
& \cellcolor{best}\textbf{30.31} & \cellcolor{second}0.894 & \cellcolor{best}\textbf{0.232} & \cellcolor{best}\textbf{59.5\%} \\

\bottomrule
\end{tabularx}
\end{table*}

Eq.~\eqref{eq:continuous_fusion} requires estimating the effective mesh opacity $\alpha_m^{\mathrm{eff}}(\mathbf{u})$. A natural alternative is to learn this opacity directly on the mesh surface, either as per-vertex alpha or as a texture-map alpha. However, such surface-alpha fields are limited in two ways.

First, surface alpha is view-invariant, while mesh inconsistencies vary with viewpoint. In geometry and occlusion conflicts, a false protrusion, hole, or inaccurate boundary may block the correct visual content only from certain viewpoints. Similarly, the textured mesh branch has fixed appearance and cannot selectively hand off highlights and reflection-like residuals to residual RGB Gaussians from specific directions.

Second, surface alpha is limited by its parameterization. Per-vertex alpha is tied to mesh tessellation, while texture-map alpha requires indirect view-space-to-texture-space optimization and can produce fragmented masks. We therefore model these inconsistencies with a volumetric, view-dependent Neural Dissolve Field.

The NDF is represented by dissolve Gaussians $\mathcal{D}$. Each dissolve Gaussian stores standard Gaussian geometry and opacity parameters, but predicts a scalar dissolve value instead of RGB color. For viewing direction $\mathbf{v}$, the $i$-th dissolve Gaussian predicts
\begin{equation}
c_{d,i}(\mathbf{v}) =
\sigma\!\left(\mathrm{SH}_{d,i}(\mathbf{v})\right) \in [0,1],
\label{eq:dissolve_scalar}
\end{equation}
where $\mathrm{SH}_{d,i}$ denotes a scalar spherical-harmonic function attached to the $i$-th dissolve Gaussian.

We rasterize dissolve Gaussians using the same Gaussian rasterizer. Let $\hat{\alpha}_{d,i}(\mathbf{u})$ denote the effective per-pixel opacity of the $i$-th dissolve Gaussian at pixel $\mathbf{u}$, sorted in front-to-back order. The dissolve value is composited as
\begin{equation}
D(\mathbf{u}) =
\sum_i T_{d,i}(\mathbf{u})\,
\hat{\alpha}_{d,i}(\mathbf{u})\,
c_{d,i}(\mathbf{v}),
\label{eq:dissolve_map}
\end{equation}
with
\begin{equation}
T_{d,i}(\mathbf{u}) =
\prod_{j<i}\left(1-\hat{\alpha}_{d,j}(\mathbf{u})\right).
\label{eq:dissolve_transmittance}
\end{equation}
Although written as $D(\mathbf{u})$, the dissolve value is view-dependent through $c_{d,i}(\mathbf{v})$. The effective mesh opacity used in Eq.~\eqref{eq:continuous_fusion} is
\begin{equation}
\alpha_m^{\mathrm{eff}}(\mathbf{u}) = 1-D(\mathbf{u}).
\label{eq:effective_mesh_alpha}
\end{equation}
Thus, $D(\mathbf{u})=0$ preserves the mesh contribution, while $D(\mathbf{u})=1$ fully dissolves it. The dissolve branch does not emit radiance. It only attenuates mesh evidence in the final fusion, allowing residual RGB Gaussians to refill the revealed content.

\subsection{Optimization Curriculum and Loss-Driven Dual Densification}
\label{sec:optimization_curriculum}

We progressively introduce the three components instead of optimizing the full hybrid representation from the beginning. 
1) We first optimize the residual Gaussian branch alone using the standard 3DGS objective, which provides a stable initialization for the subsequent hybrid optimization. 2) We then introduce the projection-textured mesh and NDF and enable continuous transmittance fusion. During hybrid optimization, we freeze the mesh texture and employ loss-driven dual densification: First, we sample high-error pixels from the current fused rendering. These pixels are back-projected into 3D, with mesh depth serving as a coarse geometric anchor, to add candidates to both the dissolve branch and the residual RGB branch. The former increases the capacity to dissolve inconsistent mesh content, while the latter increases the capacity to refill the revealed regions. All inserted candidates remain fully trainable and are not constrained to the mesh surface. 3) Finally, after Gaussian densification terminates, we continue optimizing the Gaussian representation for a short stabilization period before unfreezing the mesh texture and jointly fine-tuning all components.
 
The main Gaussian branch follows the same global densification window as standard 3DGS. For plug-in variants, each Gaussian backbone retains its update strategy. Dissolve Gaussians are initialized according to multi-view mesh-texel errors. Additional details of the initialization and dual densification are provided in the supplementary material.

\maybeinput{exp1}{Original experiment setup / dataset visual fragment.}
\maybeinput{exp12}{Original decomposition figure: GT / Mesh / Dissolve / Residual / Ours.}

\newcommand{\directionalDissolveCaption}{\textbf{Directional dissolve for view-dependent appearance.} A view-invariant dissolve field can suppress some static geometry errors, but it cannot provide a directional handoff for view-specific evidence. The scalar-SH dissolve field attenuates mesh evidence only from conflict-prone viewing directions, allowing residual RGB Gaussians to recover the corresponding view-dependent details.}
\begingroup
\let\origcaption\caption
\renewcommand{\caption}[2][]{\origcaption{\directionalDissolveCaption}}
\maybeinput{exp4}{Original view-dependent dissolve / w-o SH visual comparison.}
\endgroup

\begin{figure*}[t]
\centering
\IfFileExists{mesh/saintanne/rc_texture_cropped.jpg}{%
\def\w{0.121}
\setlength{\tabcolsep}{0.35pt}
\renewcommand{\arraystretch}{0.62}
\begingroup
\setlength{\fboxsep}{0pt}
\setlength{\fboxrule}{0.80pt}
\definecolor{lightcyanborder}{RGB}{120,210,235}

\newcommand{\textureWithBoxByCoord}[5]{%
  \begingroup
  \setlength{\meshcellw}{\w\linewidth}%
  \begin{tikzpicture}
    \node[anchor=south west, inner sep=0] (img) at (0,0)
      {\includegraphics[width=\meshcellw]{##1}};
    \begin{scope}[x={(img.south east)}, y={(img.north west)}]
      \draw[lightcyanborder, line width=0.75pt] (##2,##3) rectangle (##4,##5);
    \end{scope}
  \end{tikzpicture}%
  \endgroup
}

\newcommand{\textureCropByCoord}[5]{%
  \begingroup
  \setlength{\meshcellw}{\w\linewidth}%
  \resizebox{\meshcellw}{!}{%
    \begin{tikzpicture}
      \node[anchor=south west, inner sep=0, opacity=0] (img) at (0,0)
        {\includegraphics[width=\meshcellw]{##1}};
      \begin{scope}[x={(img.south east)}, y={(img.north west)}]
        \pgfresetboundingbox
        \path[use as bounding box] (##2,##3) rectangle (##4,##5);
        \clip (##2,##3) rectangle (##4,##5);
        \node[anchor=south west, inner sep=0] at (0,0)
          {\includegraphics[width=\meshcellw]{##1}};
      \end{scope}
    \end{tikzpicture}%
  }%
  \endgroup
}

\newcommand{\bluecrop}[1]{\fcolorbox{lightcyanborder}{white}{##1}}

\setlength{\tabcolsep}{0.2pt}
\newlength{\groupgap}
\setlength{\groupgap}{1pt}

\begin{tabular}{@{}cccc@{\hspace{\groupgap}}cccc@{}}
\multicolumn{4}{c}{\small\textbf{SaintAnne}} &
\multicolumn{4}{c}{\small\textbf{ML-Hypersim-001}} \\[0pt]

\includegraphics[width=\w\linewidth]{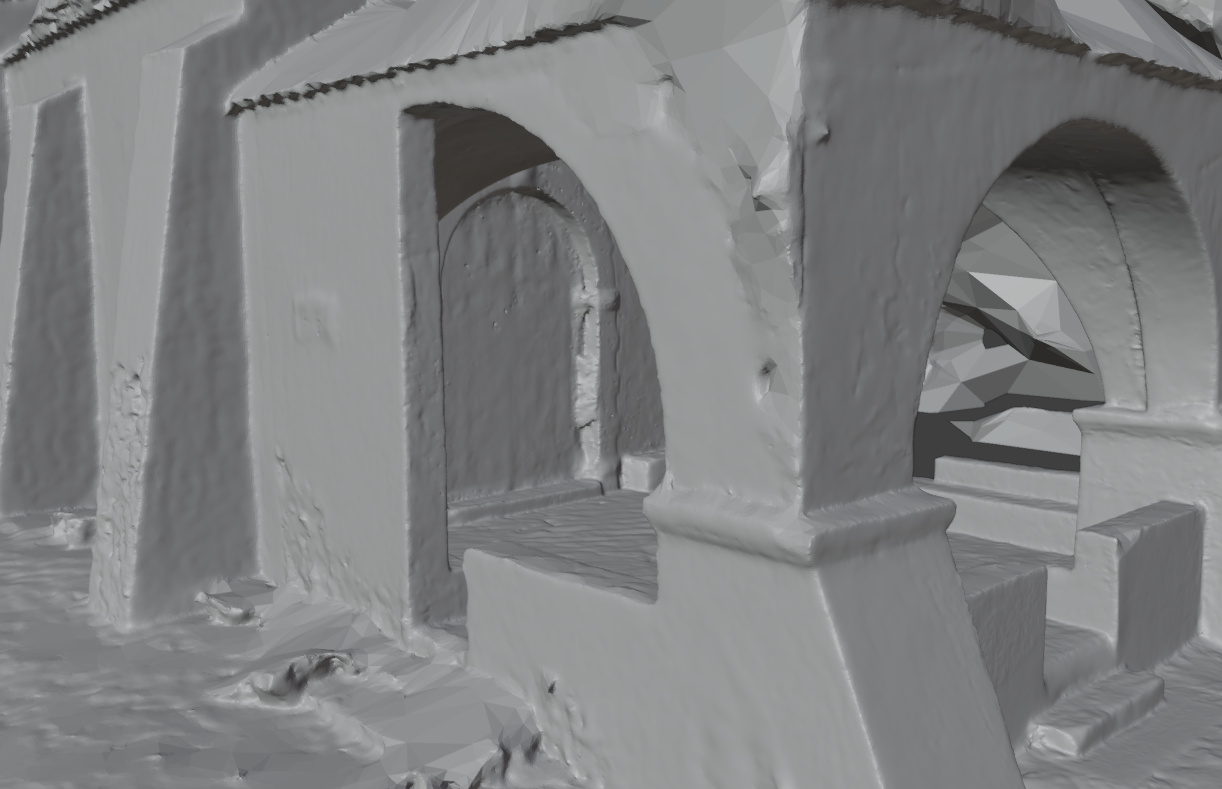} &
\includegraphics[width=\w\linewidth]{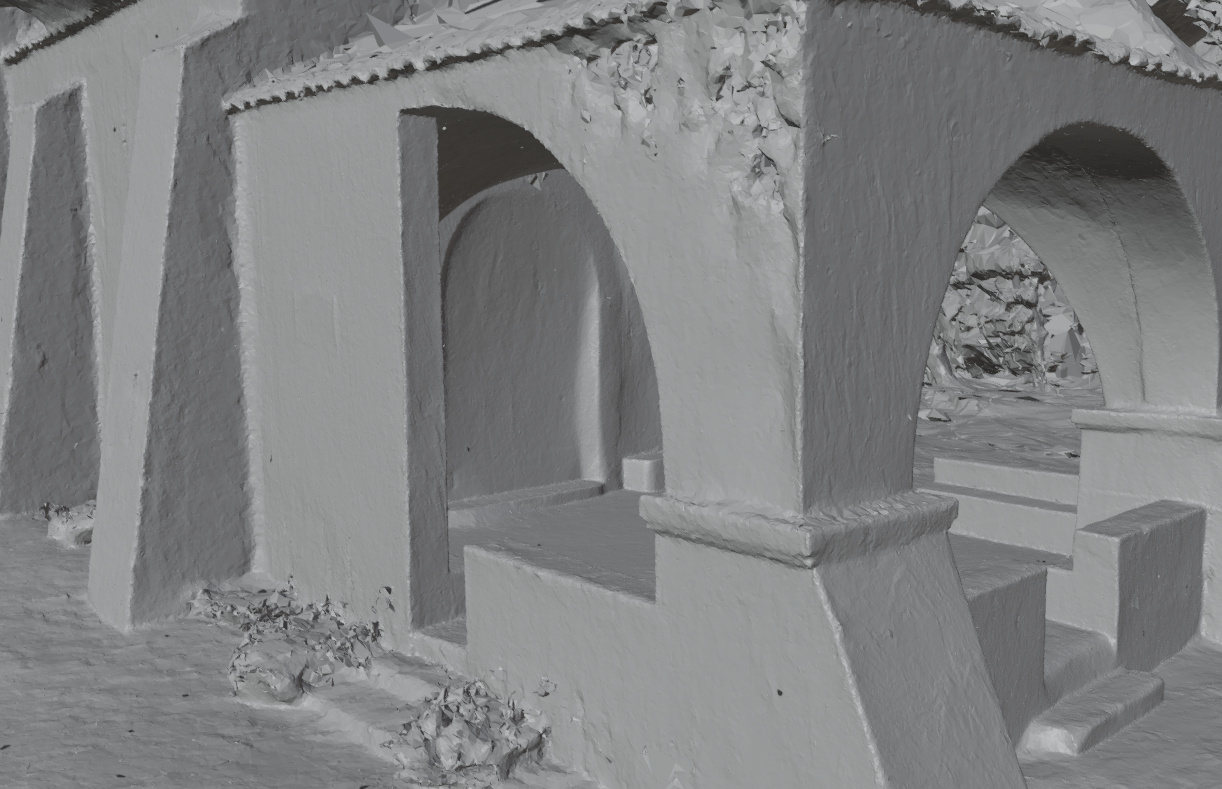} &
\includegraphics[width=\w\linewidth]{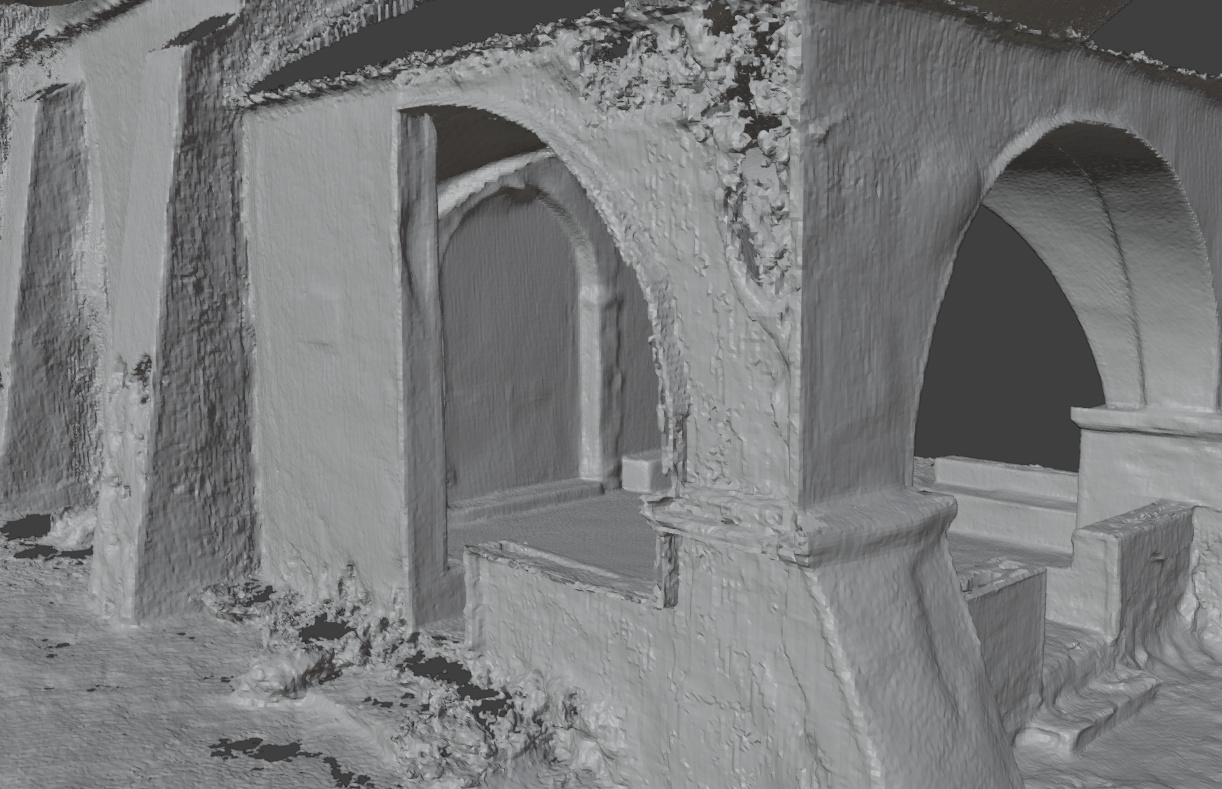} &
\includegraphics[width=\w\linewidth]{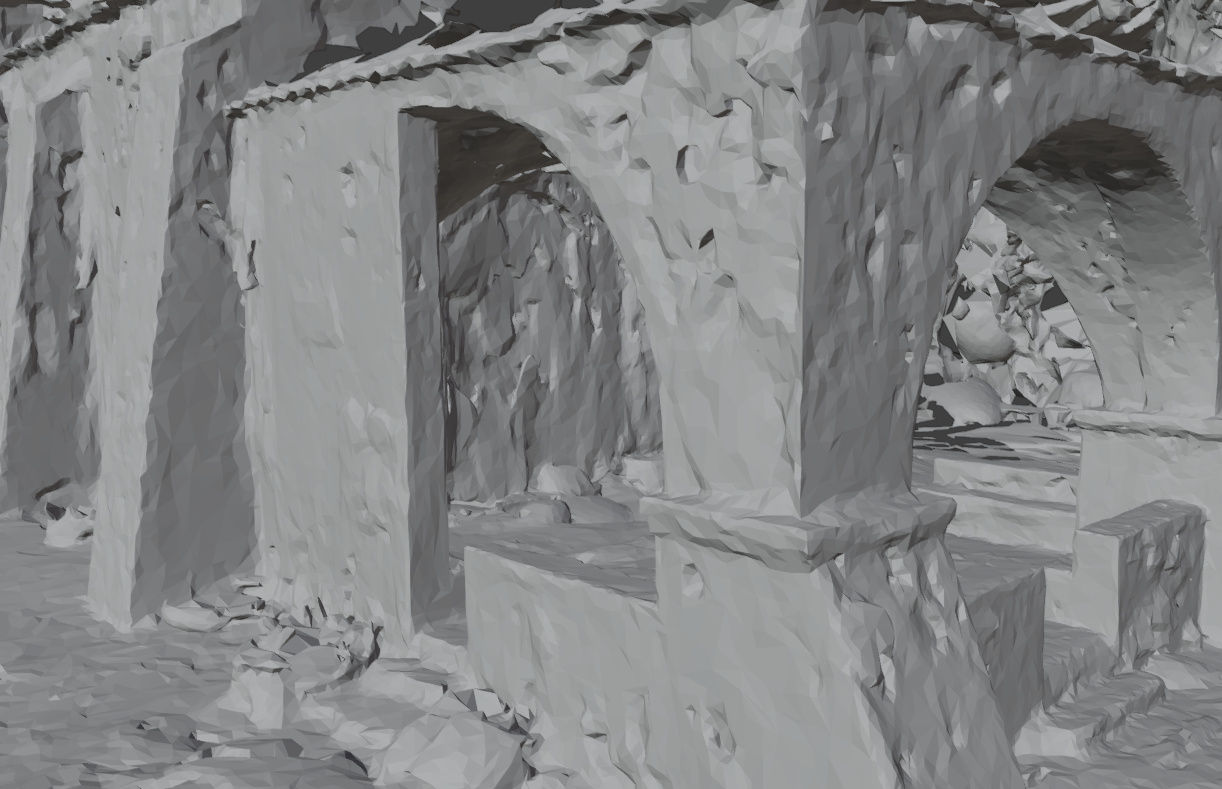} &
\includegraphics[width=\w\linewidth]{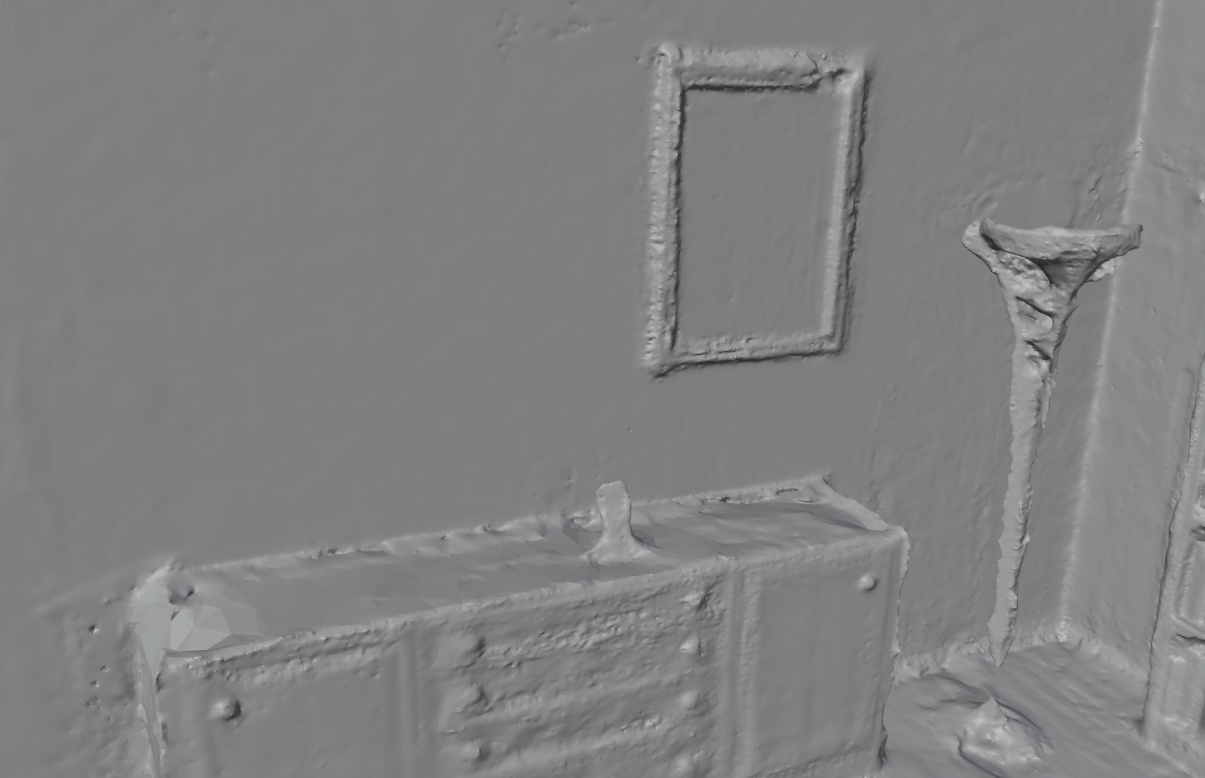} &
\includegraphics[width=\w\linewidth]{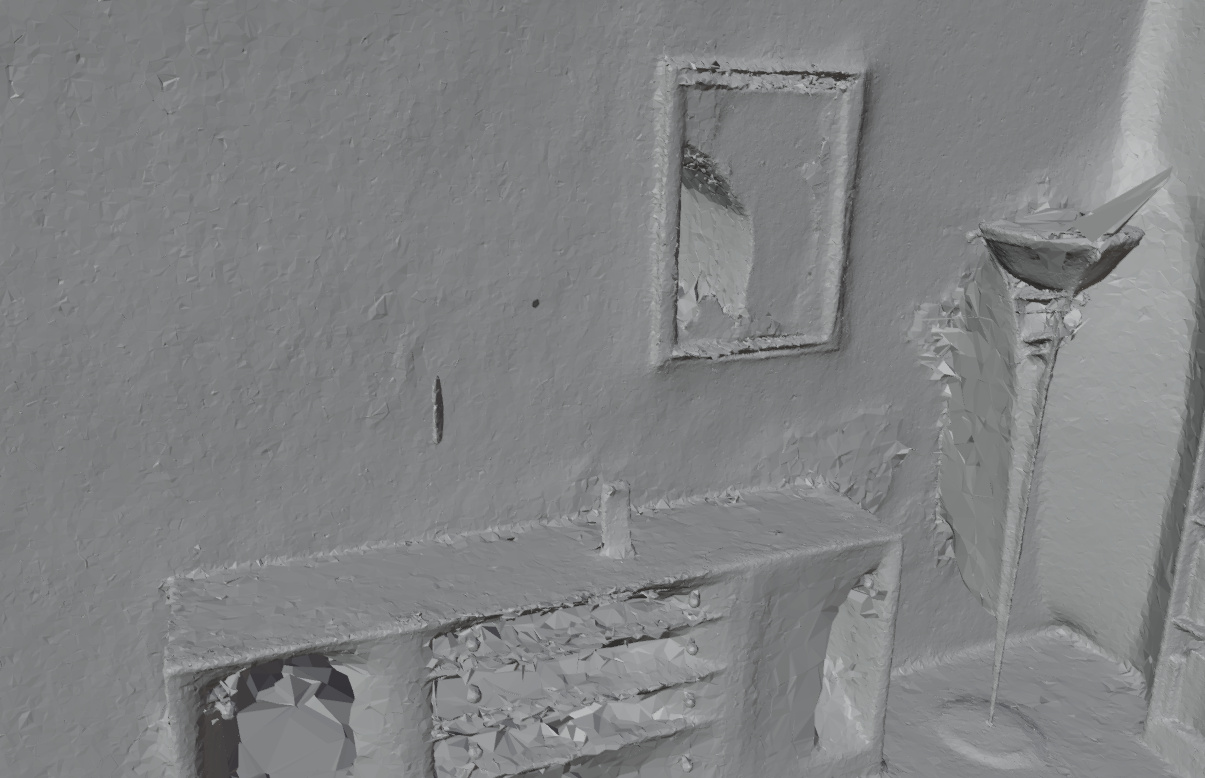} &
\includegraphics[width=\w\linewidth]{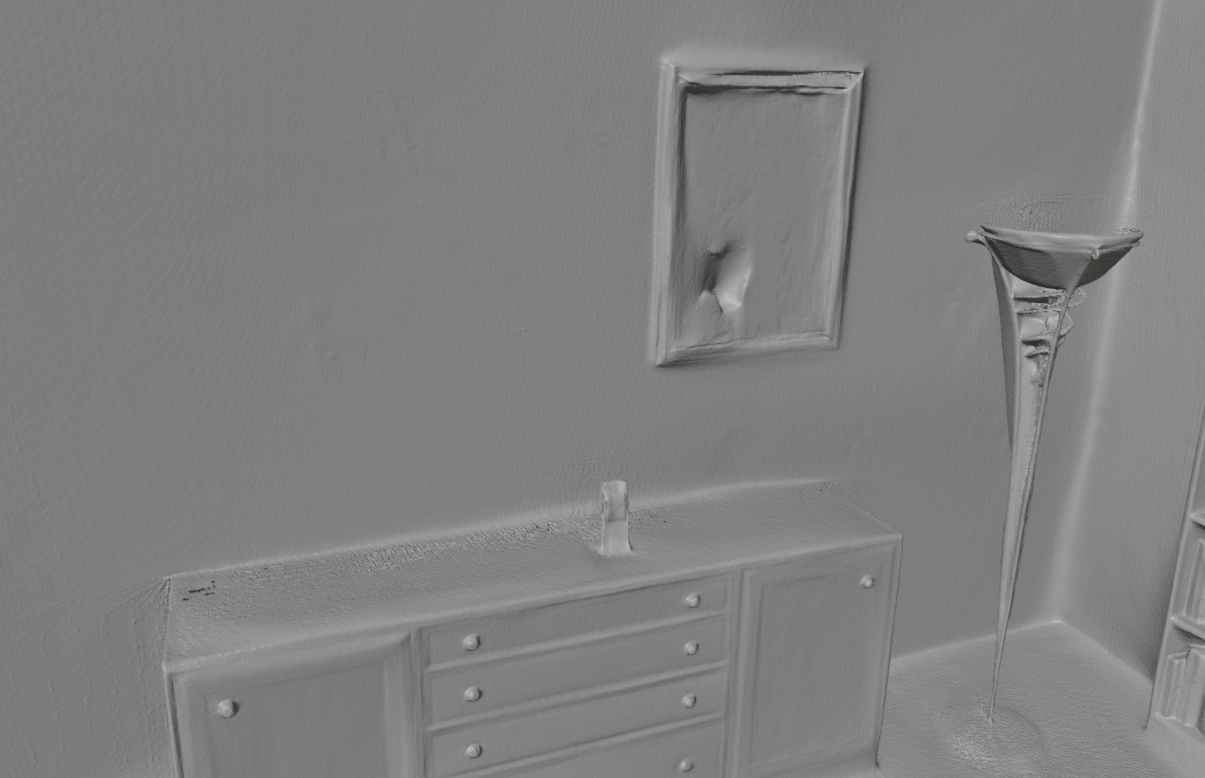} &
\includegraphics[width=\w\linewidth]{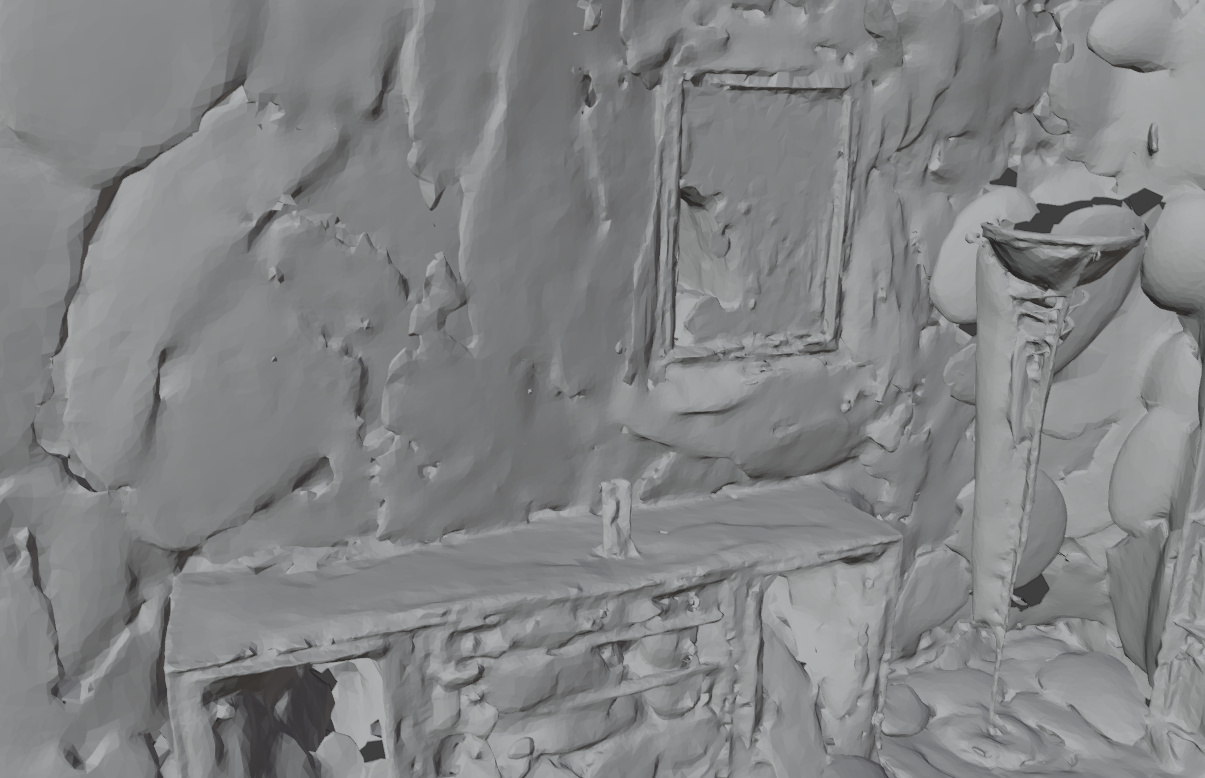} \\[0pt]

\bluecrop{\textureCropByCoord{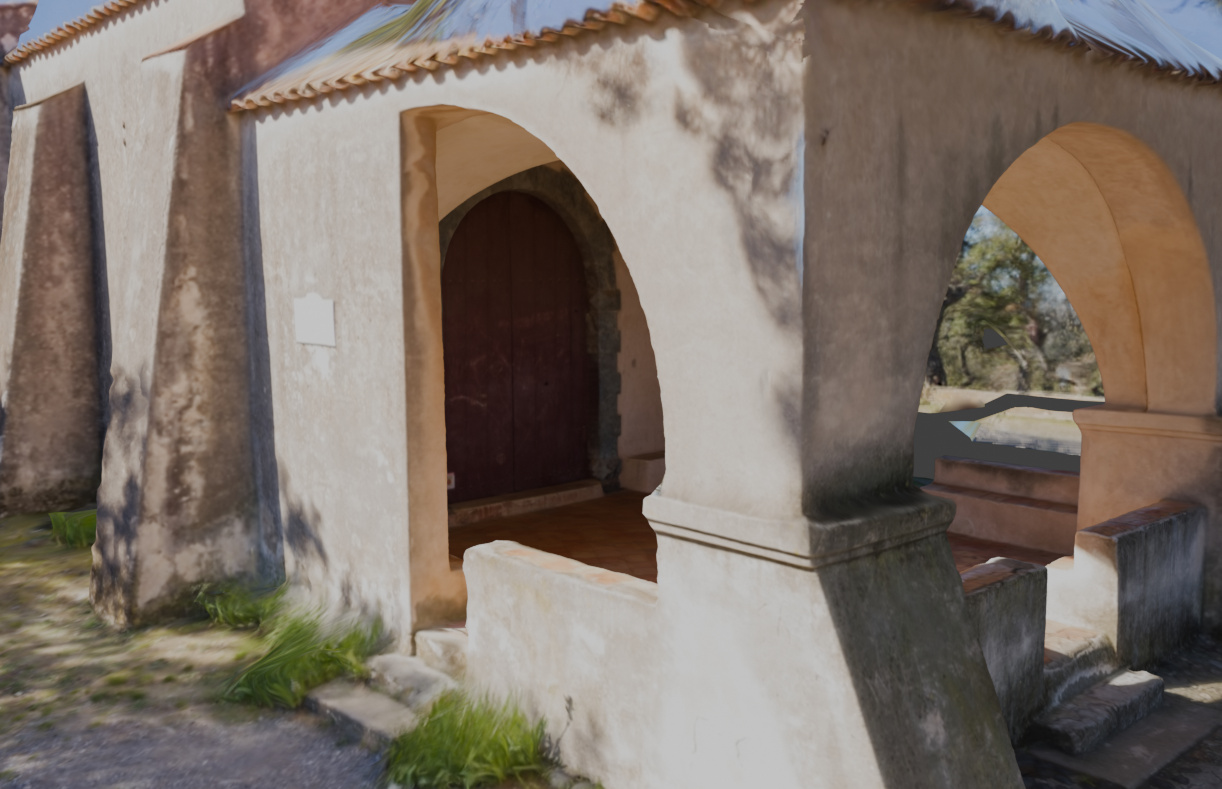}{0.60}{0.10}{0.83}{0.32}} &
\bluecrop{\textureCropByCoord{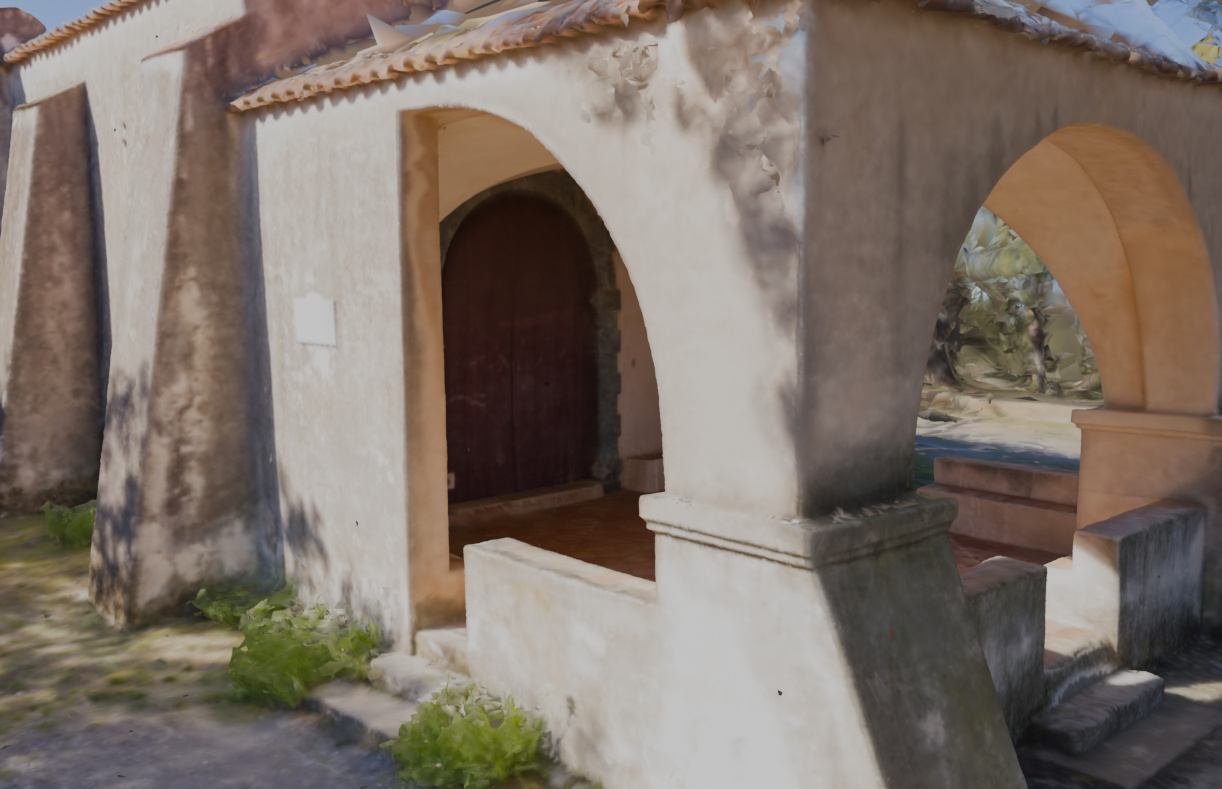}{0.60}{0.10}{0.83}{0.32}} &
\bluecrop{\textureCropByCoord{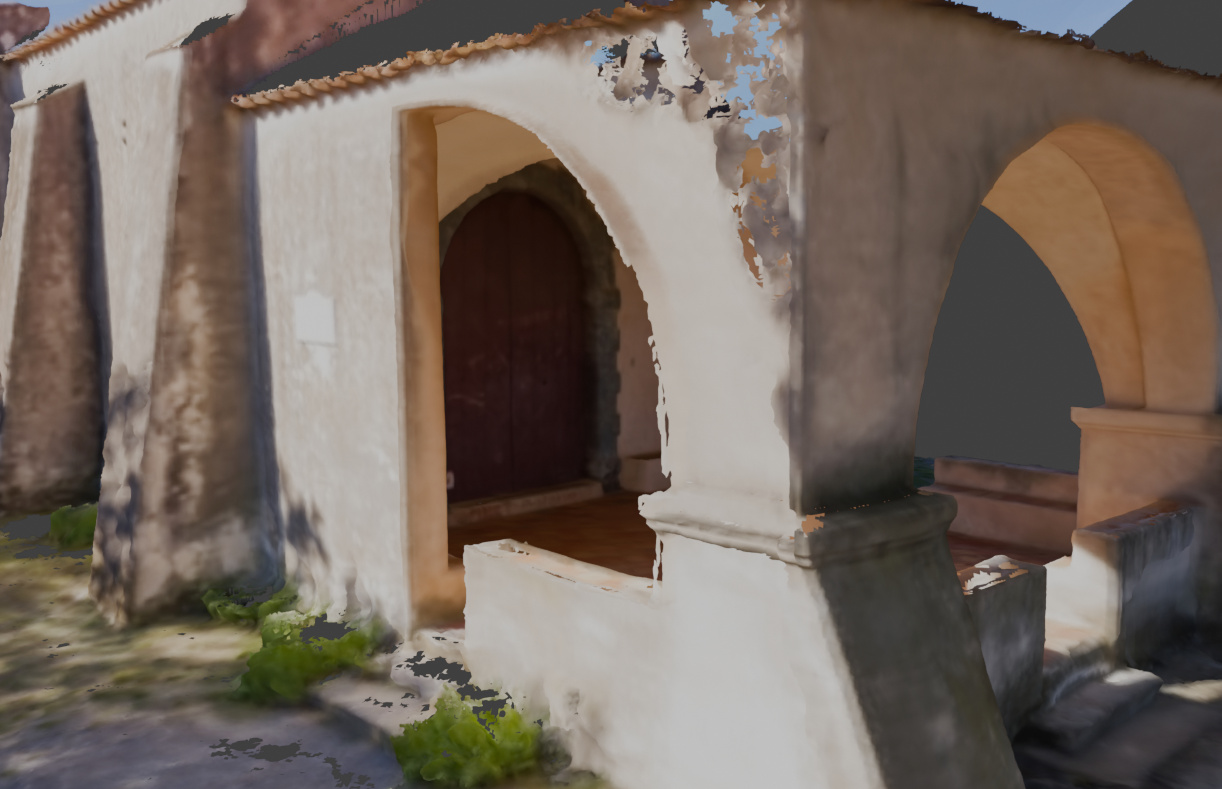}{0.60}{0.10}{0.83}{0.32}} &
\bluecrop{\textureCropByCoord{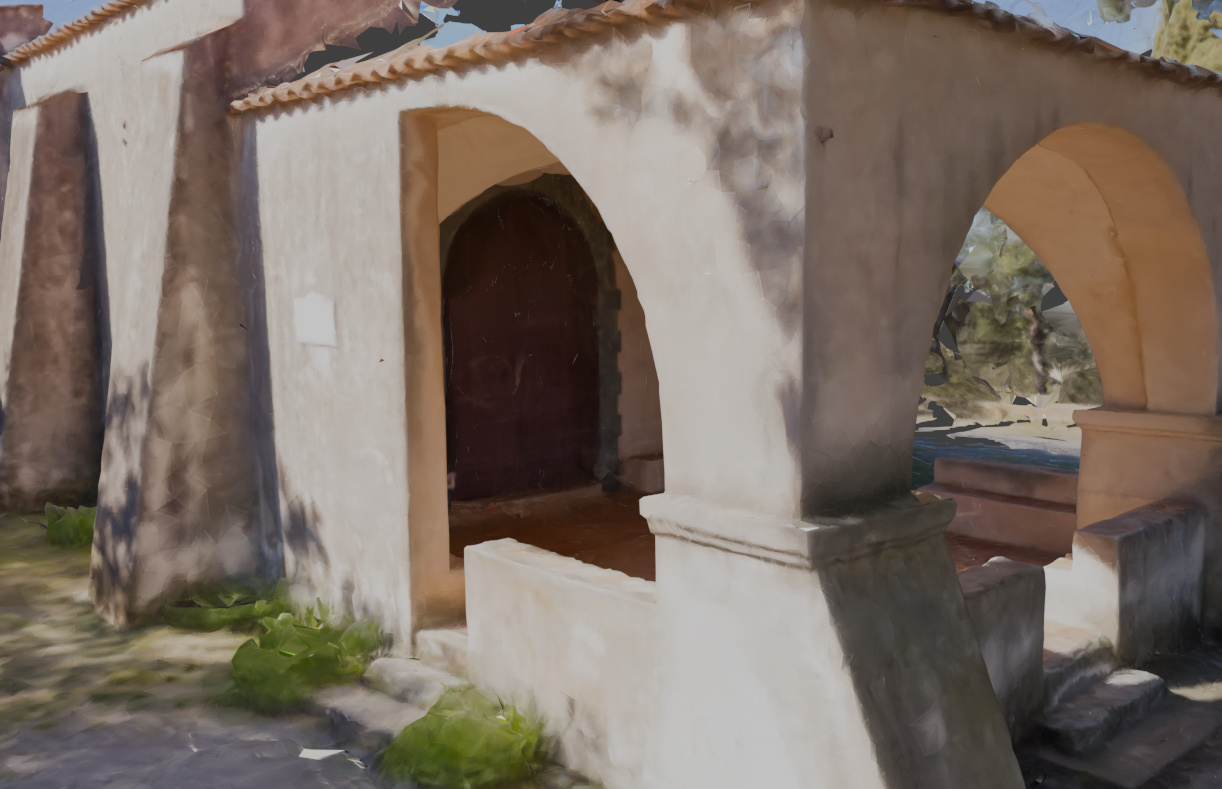}{0.60}{0.10}{0.83}{0.32}} &
\bluecrop{\textureCropByCoord{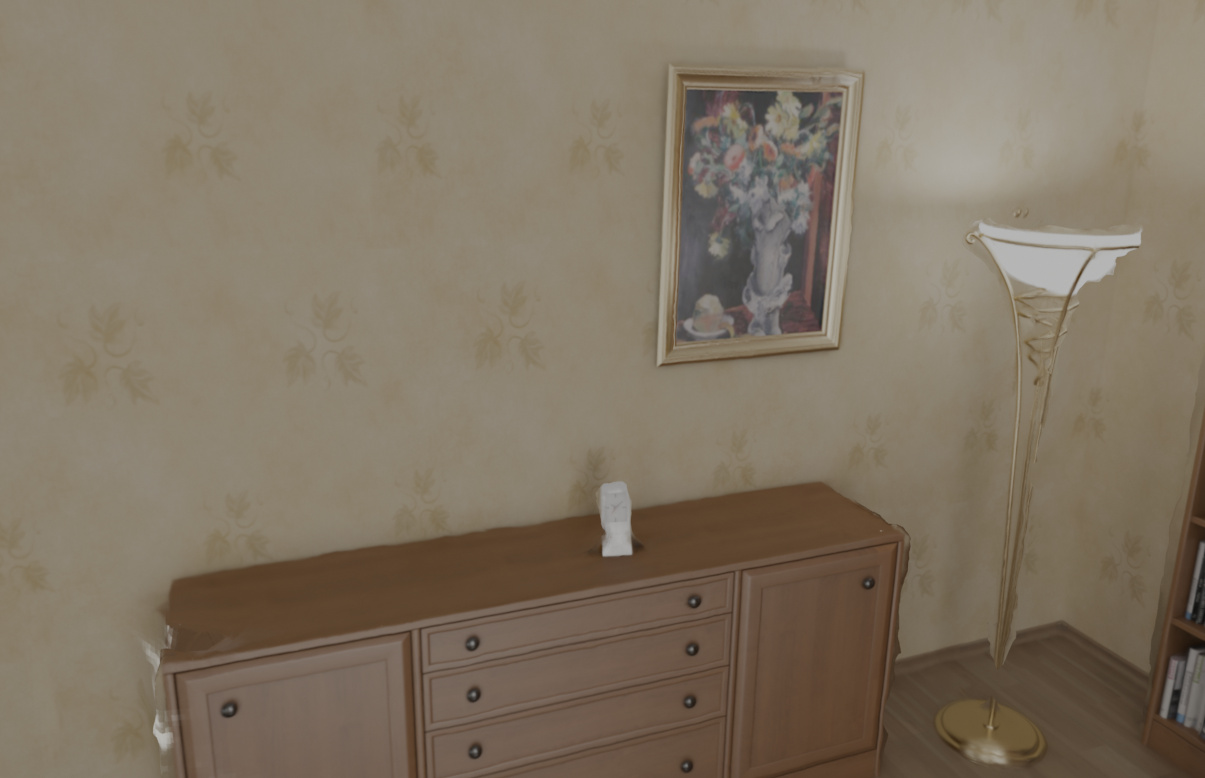}{0.46}{0.68}{0.70}{0.91}} &
\bluecrop{\textureCropByCoord{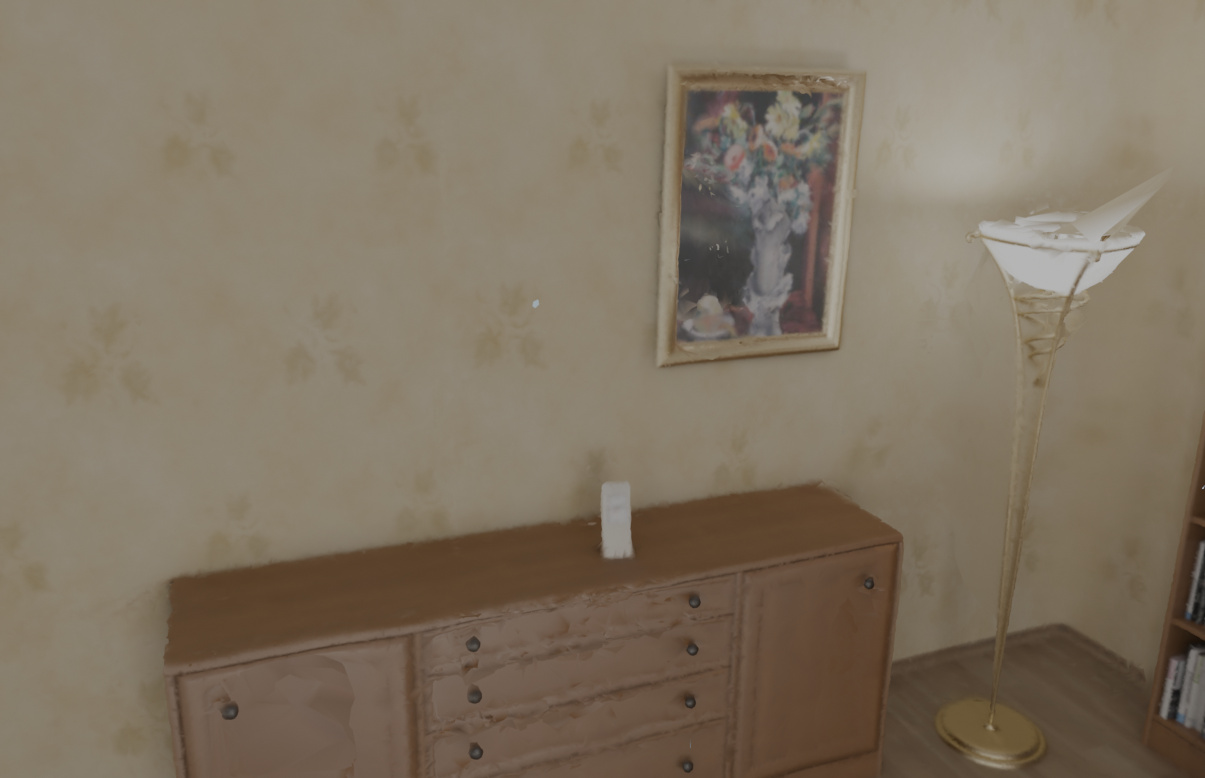}{0.46}{0.68}{0.70}{0.91}} &
\bluecrop{\textureCropByCoord{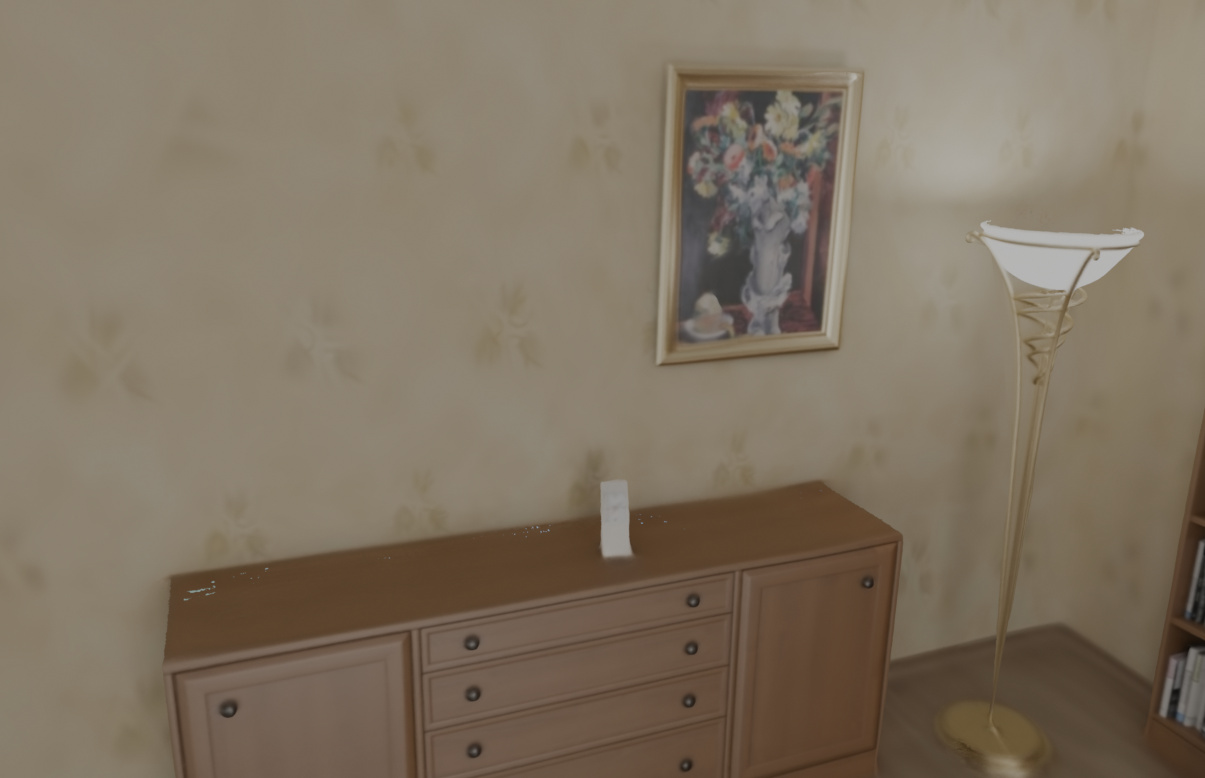}{0.46}{0.68}{0.70}{0.91}} &
\bluecrop{\textureCropByCoord{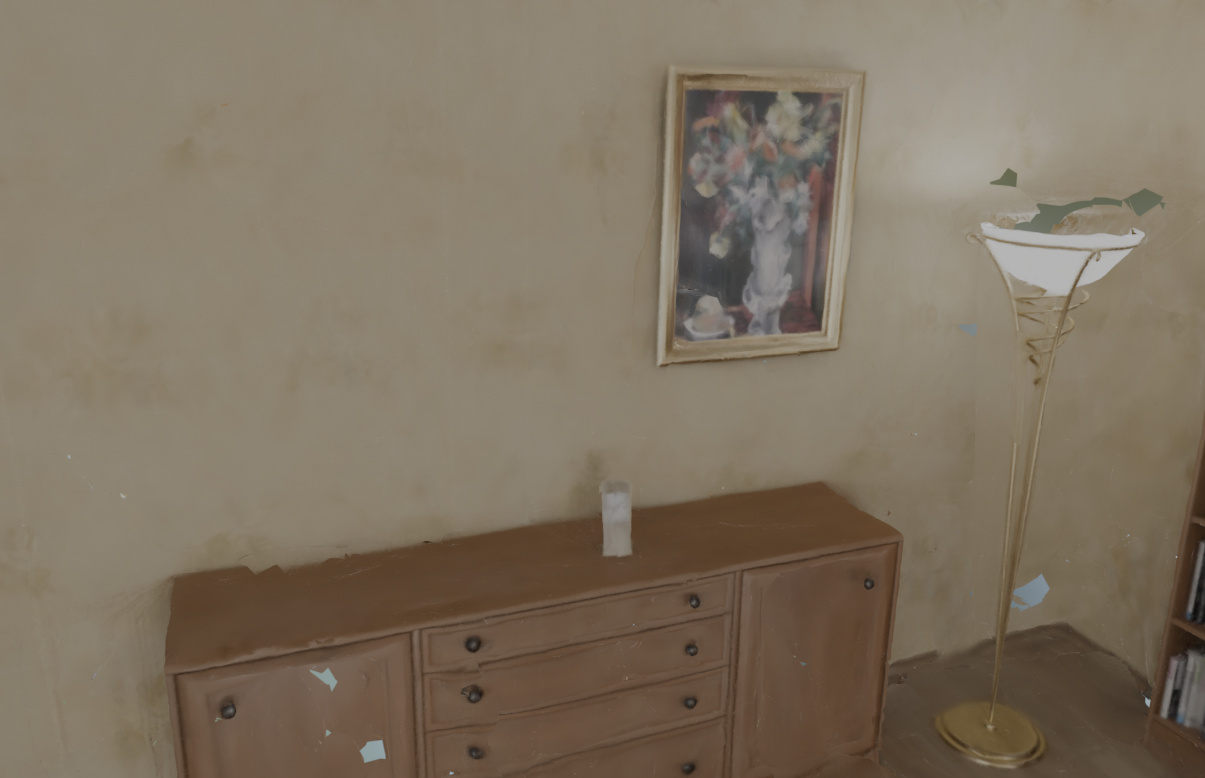}{0.46}{0.68}{0.70}{0.91}} \\[0pt]

\small RC & \small MILo & \small PGSR & \small SuGaR &
\small RC & \small MILo & \small PGSR & \small SuGaR 
\end{tabular}
\endgroup
}{\fbox{\parbox[c][0.14\textheight][c]{0.92\linewidth}{\centering Missing mesh prior images.}}}
\caption{\textbf{Textured mesh priors and direct image projection.} We compare RC, MILo, PGSR, and SuGaR mesh priors on SaintAnne and ML-Hypersim-001. The top row visualizes the reconstructed mesh geometry, and the bottom row shows enlarged textured renderings. Direct projection preserves sharper local appearance than appearance baked from reconstructed representations.}
\label{fig:mesh_prior_sanity}
\end{figure*}

\section{Experiments}
\label{sec:experiments}
\subsection{Setup}
\label{sec:setup}

\begin{figure}[t]
\centering
\setlength{\tabcolsep}{0pt}
\renewcommand{\arraystretch}{1.0}

\def\depthw{0.241\linewidth}
\def\rowgap{0pt}

\newcommand{\depthImgBox}[5]{%
\begin{tikzpicture}[
  baseline=(depthimage.south),
  inner sep=0pt,
  outer sep=0pt
]
  \node[
    anchor=south west,
    inner sep=0pt,
    outer sep=0pt
  ] (depthimage) at (0,0)
  {\includegraphics[width=\depthw]{#1}};

  \begin{scope}[
    x={(depthimage.south east)},
    y={(depthimage.north west)}
  ]
    \draw[
      red!70!black,
      line width=0.65pt
    ] (#2,#3) rectangle (#4,#5);
  \end{scope}
\end{tikzpicture}%
}

\begin{tabular}{
@{}c@{\hspace{1.2pt}}
c@{\hspace{1.2pt}}
c@{\hspace{1.2pt}}
c@{}
}
\small\textbf{DA-V2} &
\small\textbf{Mesh} &
\small\textbf{Ours} &
\small\textbf{Hard-Z} \\[-1pt]

\depthImgBox{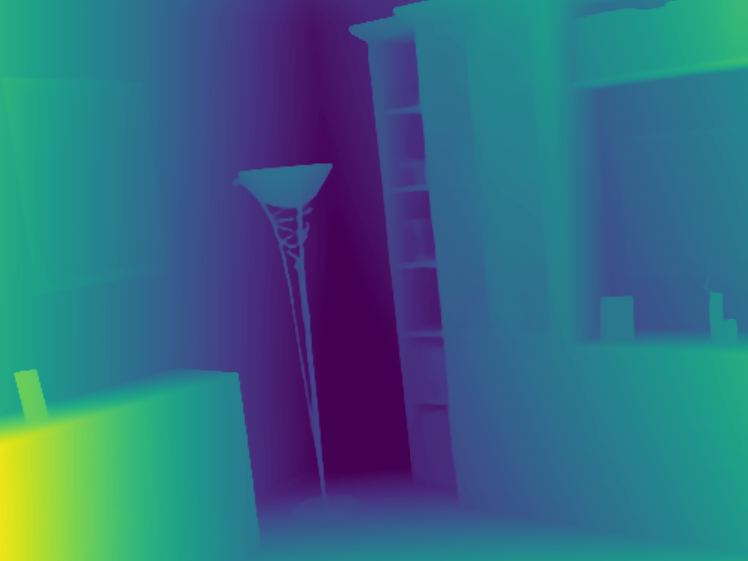}
  {0.20}{0.10}{0.55}{0.82} &
\depthImgBox{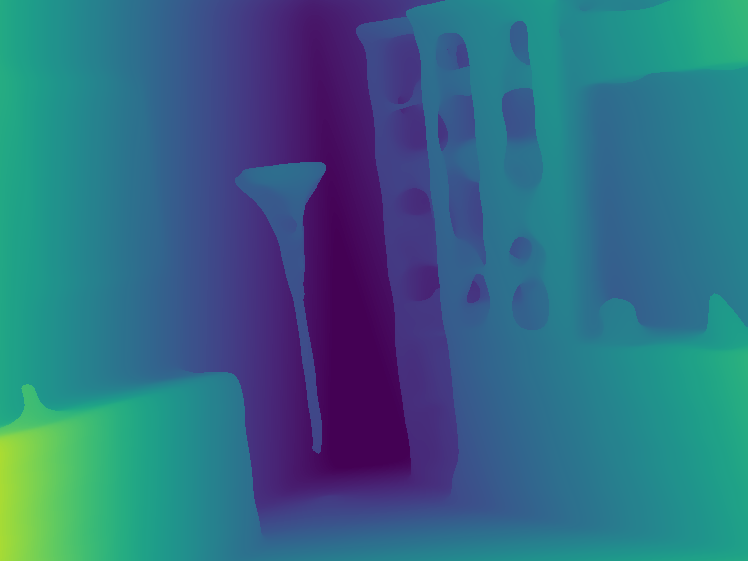}
  {0.20}{0.10}{0.55}{0.82} &
\depthImgBox{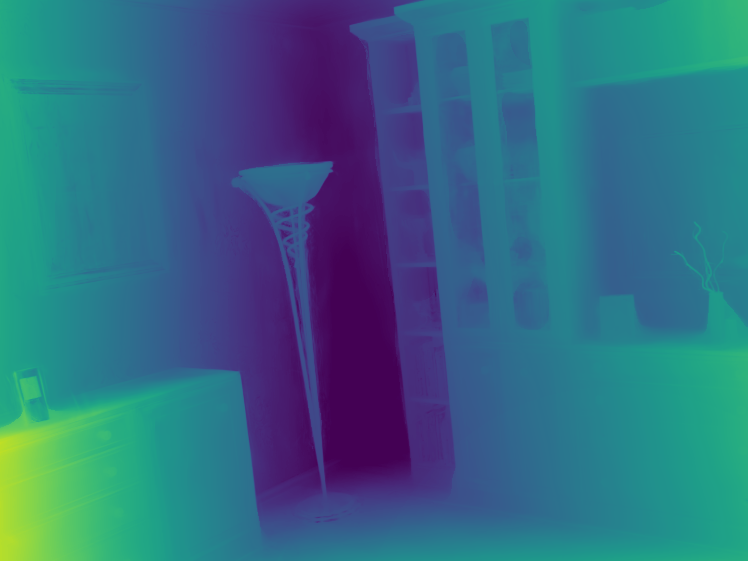}
  {0.20}{0.10}{0.55}{0.82} &
\depthImgBox{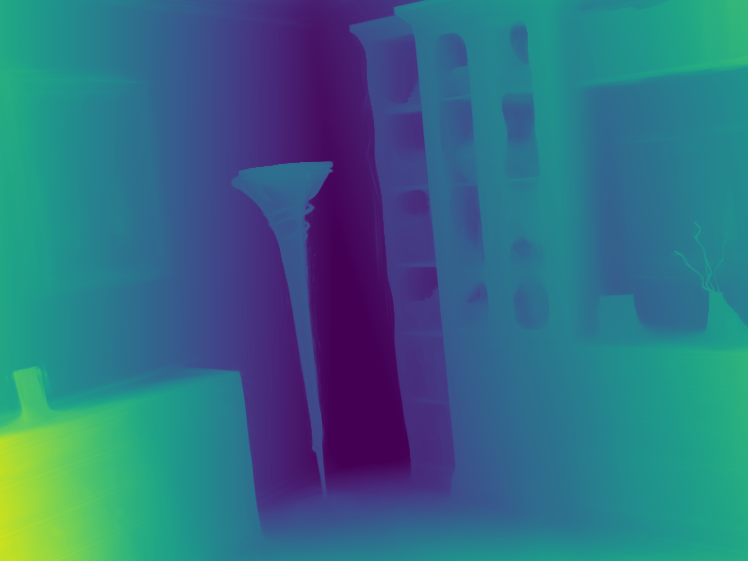}
  {0.20}{0.10}{0.55}{0.82}
\\[\rowgap]

\addlinespace[-1pt]

\depthImgBox{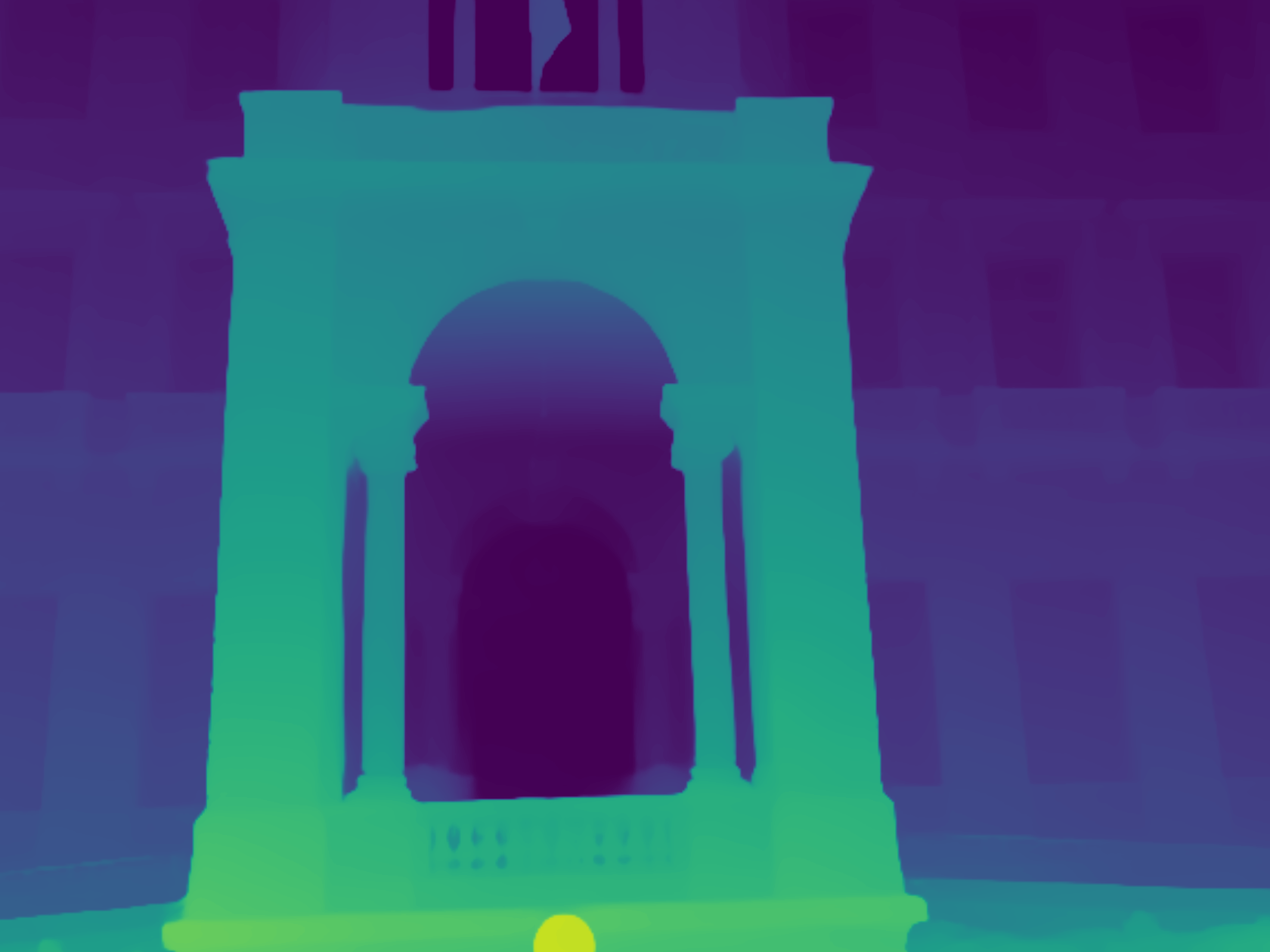}
  {0.10}{0.70}{0.78}{0.95} &
\depthImgBox{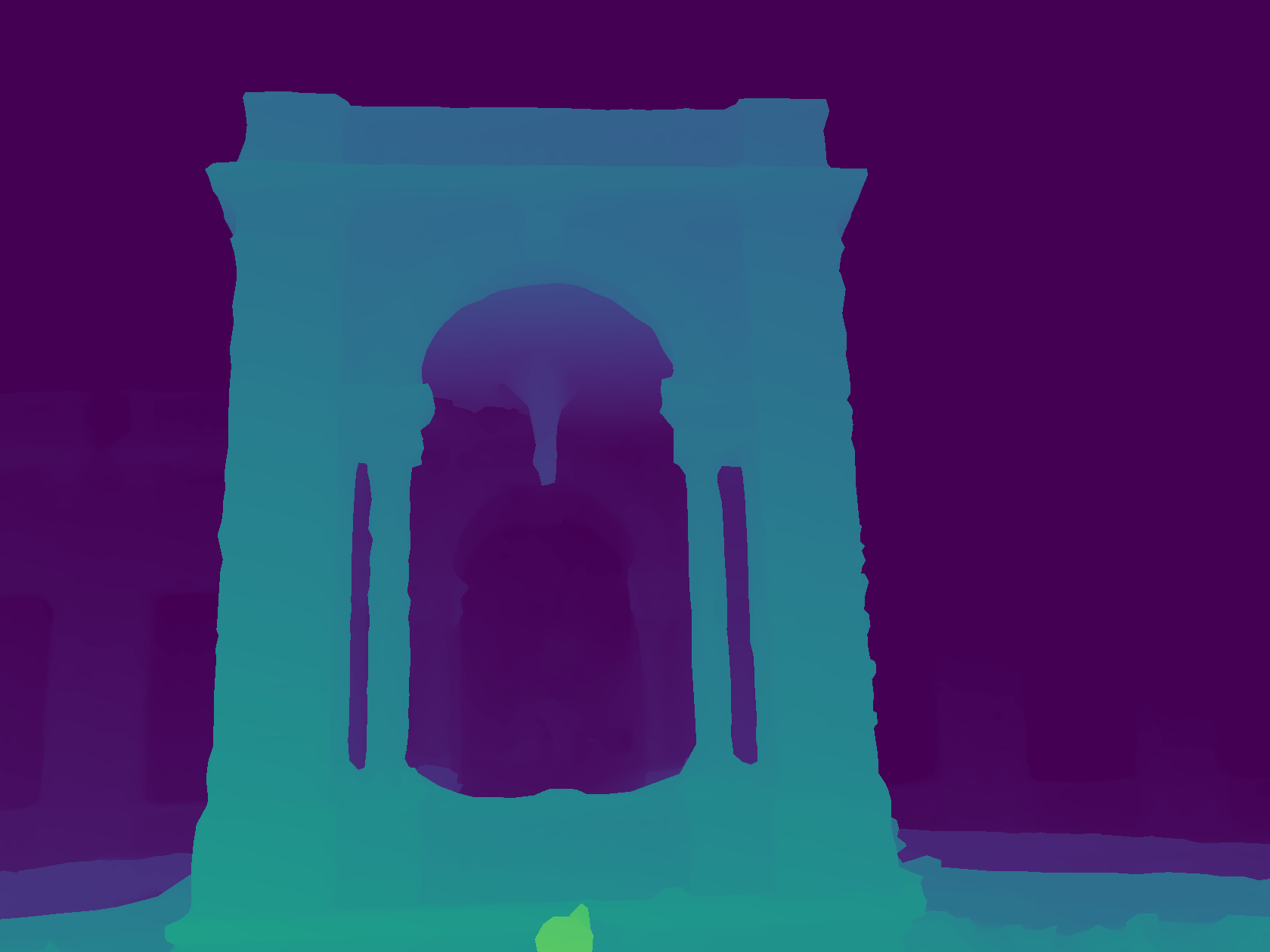}
  {0.10}{0.70}{0.78}{0.95} &
\depthImgBox{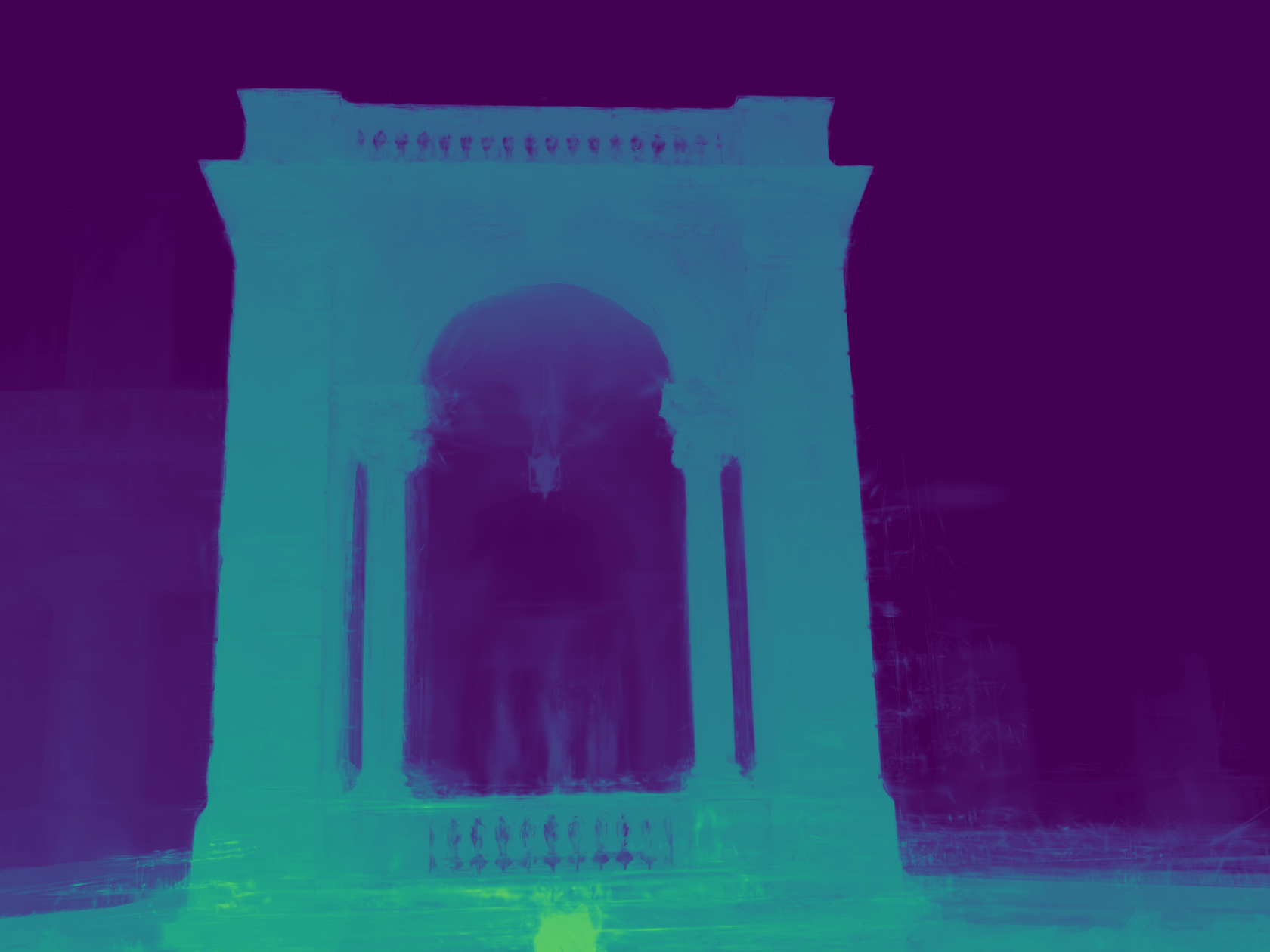}
  {0.10}{0.70}{0.78}{0.95} &
\depthImgBox{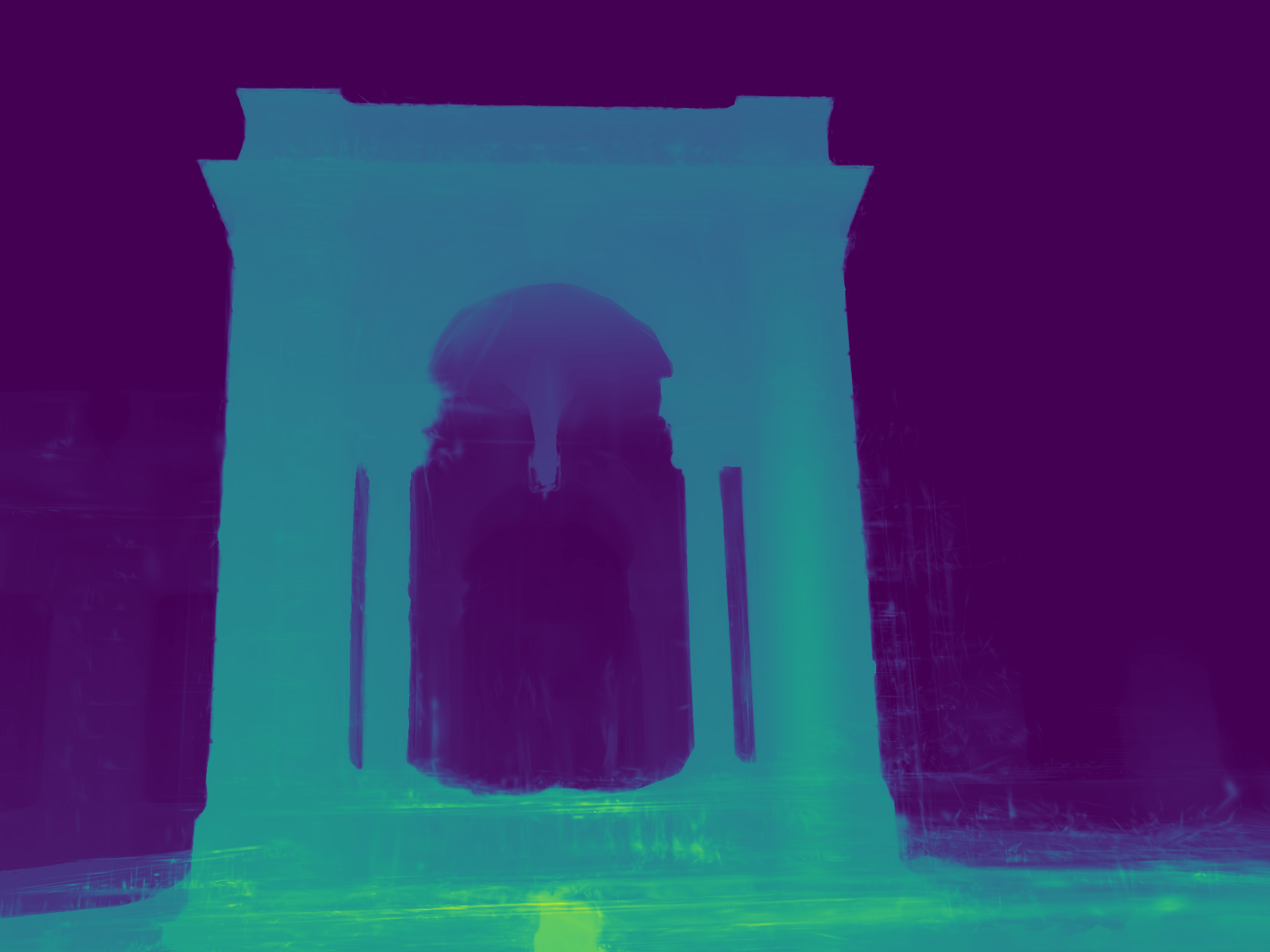}
  {0.10}{0.70}{0.78}{0.95}
\\[\rowgap]

\addlinespace[-1pt]

\depthImgBox{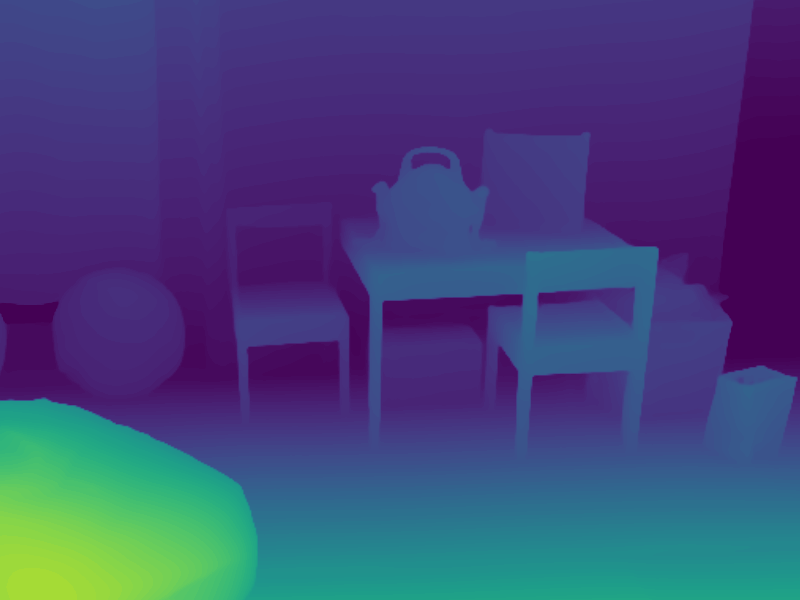}
  {0.38}{0.20}{0.80}{0.85} &
\depthImgBox{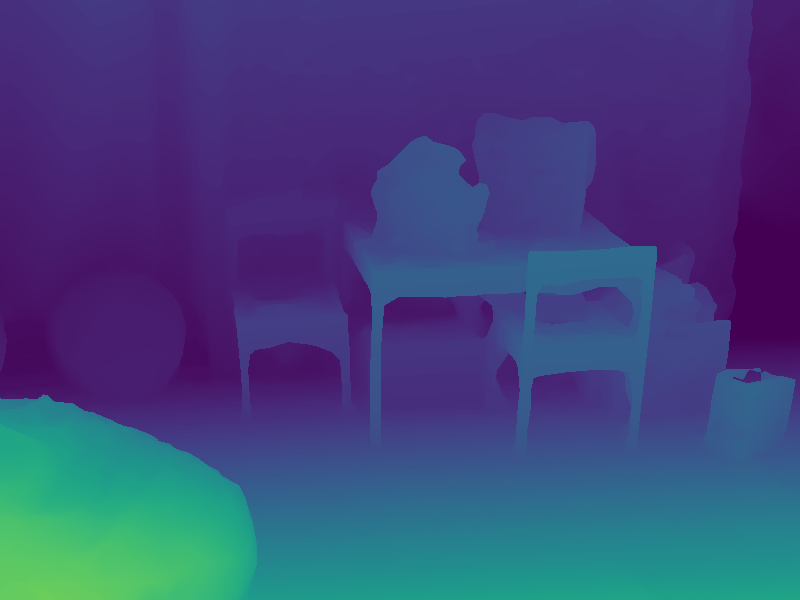}
  {0.38}{0.20}{0.80}{0.85} &
\depthImgBox{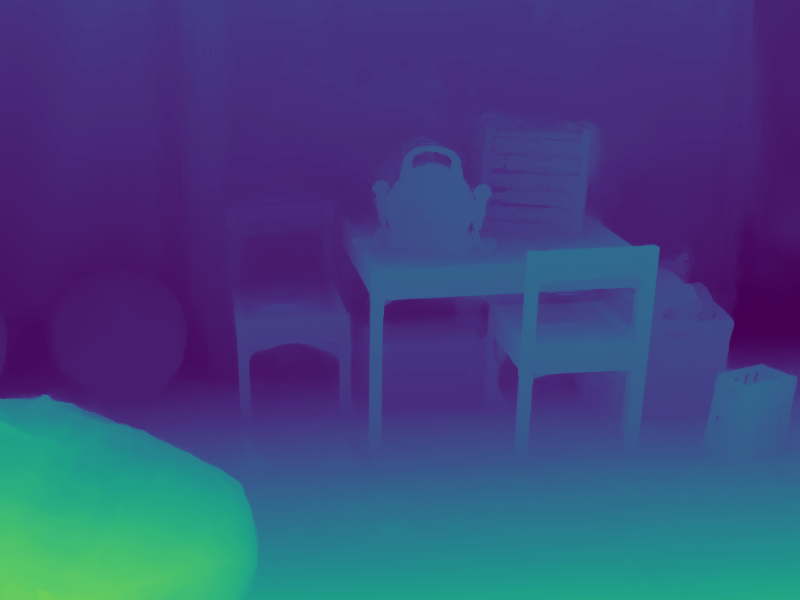}
  {0.38}{0.20}{0.80}{0.85} &
\depthImgBox{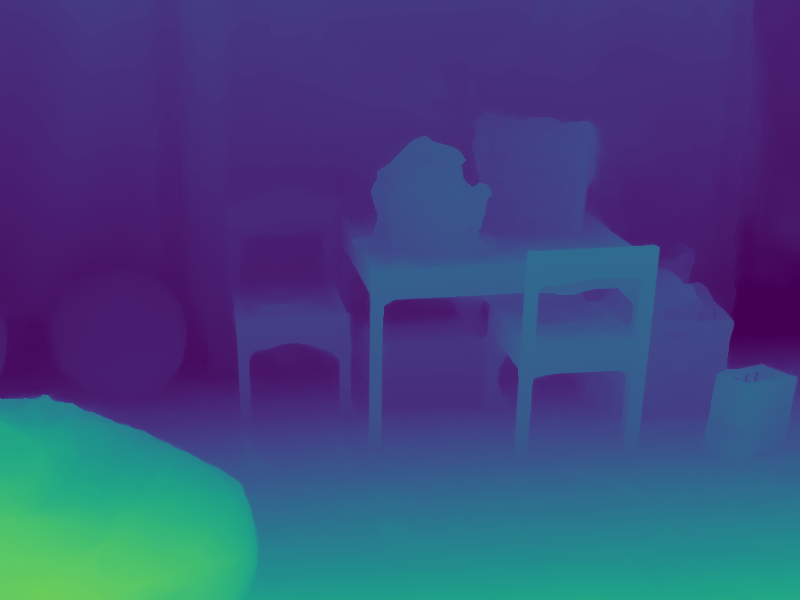}
  {0.38}{0.20}{0.80}{0.85}
\\[\rowgap]

\addlinespace[-1pt]

\depthImgBox{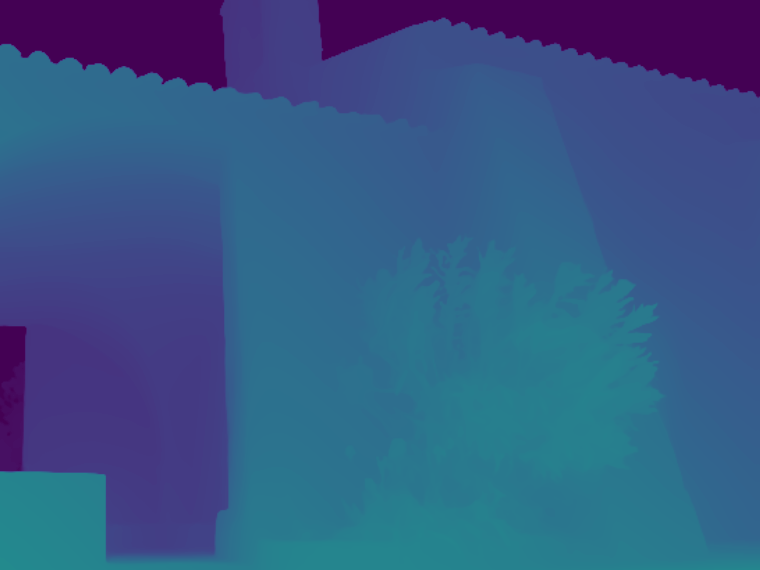}
  {0.10}{0.70}{0.78}{0.97} &
\depthImgBox{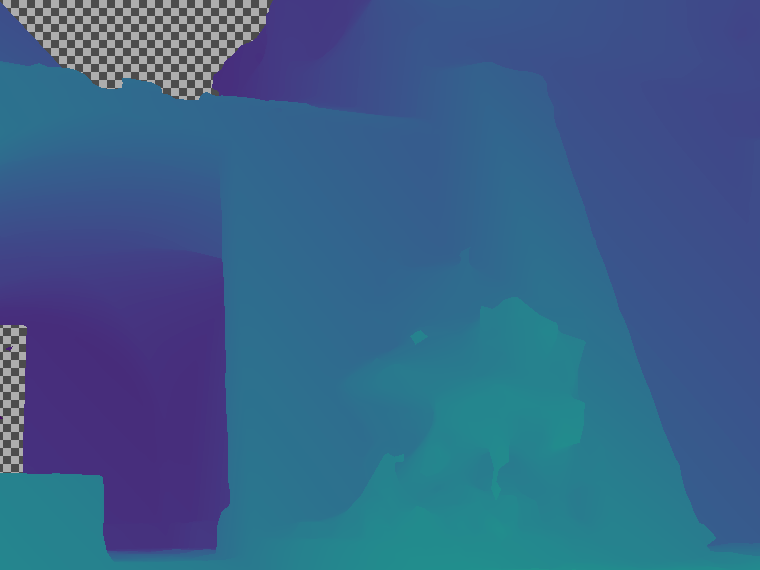}
  {0.10}{0.70}{0.78}{0.97} &
\depthImgBox{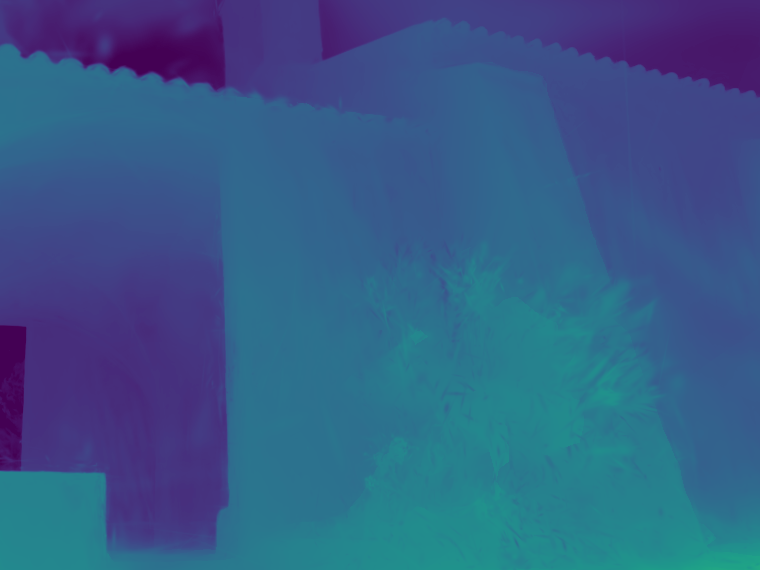}
  {0.10}{0.70}{0.78}{0.97} &
\depthImgBox{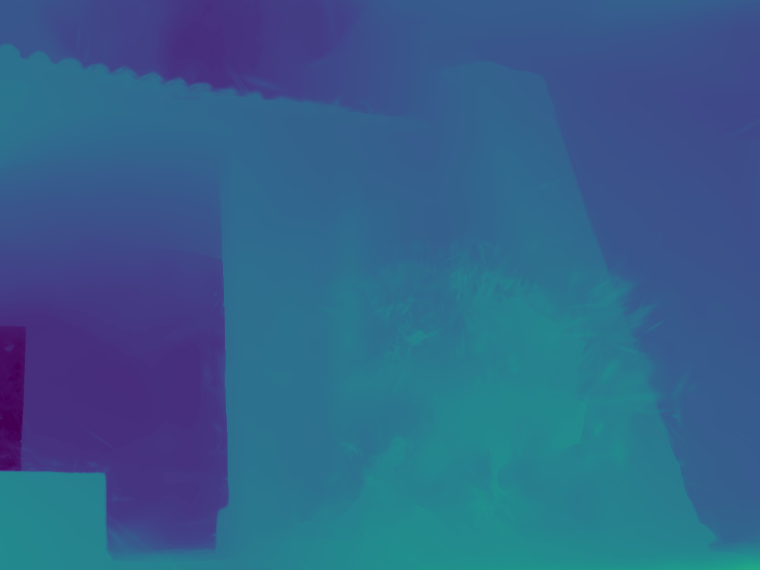}
  {0.10}{0.70}{0.78}{0.97}
\end{tabular}

\caption{\textbf{Rendered-depth comparison.}
Aligned Depth Anything V2 (DA-V2) reference~\cite{yang2024depthanythingv2}, mesh depth, \gaussolve{}, and Hard-Z. Red boxes mark representative conflict regions.}
\label{fig:depth_consistency}
\end{figure}

\paragraph{Datasets and metrics.}
We evaluate on ML-Hypersim-AI~\cite{roberts2021hypersim}, Deep Blending~\cite{hedman2018deepblending}. These datasets cover structured indoor scenes, semi-structured indoor/outdoor scenes, and foliage-heavy unstructured scenes. We report PSNR and SSIM~\cite{wang2004ssim} for distortion and structural fidelity, LPIPS~\cite{zhang2018lpips} for perceptual similarity, Gaussian reduction $\Delta GS=(N_{\mathrm{3DGS}}-N_{\mathrm{method}})/N_{\mathrm{3DGS}}$, and Aux.\% as the fraction of auxiliary Gaussians within total Gaussians. Stress tests on unstructured scenes are provided in the supplementary material.

\paragraph{Baselines.}
We compare against Gaussian-only, surface-oriented, and hybrid baselines. For Gaussian-only methods, we include 3DGS~\cite{kerbl2023gaussian}, 3DGS-MCMC~\cite{kheradmand20243d}, and Reduced 3DGS~\cite{papantonakis2024reduced3dgs}. For surface-oriented reconstruction, we compare with SuGaR~\cite{guedon2024sugar}, PGSR~\cite{chen2024pgsr}, and MeshSplatting~\cite{held2026meshsplatting}. We further include 3D-GES~\cite{ye2025whengaussianmeetssurfel} as a compact hybrid baseline.  For a controlled comparison with hard depth-based fusion, we instantiate a Hard-Z Proxy with the same textured mesh prior as \gaussolve{}. Since official implementations of representative Mesh--Gaussian hybrid scene-reconstruction methods such as MeshGS~\cite{choi2024meshgs} and Hybrid Mesh-Gaussian~\cite{huang2025hybridmeshgaussian} are unavailable, we report the published Deep Blending results in Tab.~\ref{tab:hybrid_comparison}. To evaluate its compatibility with different Gaussian backbones, we further plug \gaussolve{} into 3DGS-MCMC~\cite{kheradmand20243d} and Reduced 3DGS~\cite{papantonakis2024reduced3dgs}. Detailed implementations are provided in the supplementary material. All experiments are run on an NVIDIA RTX 5000 Ada GPU.

\subsection{Quantitative and Qualitative Comparisons}
\label{sec:main_results}

Table~\ref{tab:main_results} summarizes the results on structured and semi-structured scenes. Across all three Gaussian backbones, integrating \gaussolve{} consistently improves reconstruction quality, showing that the proposed fusion is not tied to a specific Gaussian optimization strategy. The 3DGS-MCMC variant reaches the highest PSNR on both datasets, while the Reduced 3DGS variant provides a stronger quality--compactness trade-off with substantially fewer Gaussians. The MCMC variants show slightly negative $\Delta$GS because 3DGS-MCMC removes the opacity reset and opacity-based pruning used by vanilla 3DGS. We retain its original update strategy as our goal here is to evaluate whether the fusion framework improves the backbone without changing its native optimization behavior. Compared with the Hard-Z Proxy under the same textured-mesh setting, \gaussolve{} consistently improves reconstruction quality, confirming the benefit of continuous subtract-and-refill fusion over hard visibility. Table~\ref{tab:hybrid_comparison} further compares against MeshGS and Hybrid Mesh-Gaussian on the two commonly reported Deep Blending scenes. \gaussolve{} achieves higher average PSNR and substantially lower LPIPS while reducing the Gaussian population by 59.5\%, compared with approximately 30\% for MeshGS and 36\% for Hybrid Mesh-Gaussian.

\newcommand{\tms}[1]{\mbox{#1\,ms}}

\begin{table}[t]
\centering
\scriptsize
\setlength{\tabcolsep}{2.1pt}
\renewcommand{\arraystretch}{1.08}

\caption{\textbf{Efficiency breakdown.}
Average rendering efficiency, storage, and memory on the Deep Blending dataset.}
\label{tab:efficiency-modules}

\begin{tabularx}{\linewidth}{@{}l|YYY|YYY@{}}
\toprule

& \multicolumn{3}{c|}{\textbf{3DGS}}
& \multicolumn{3}{c}{\textbf{Reduced 3DGS}} \\
\cmidrule(lr){2-4}
\cmidrule(lr){5-7}

Component
& Storage & FPS & Mem.
& Storage & FPS & Mem. \\
\midrule

Mesh Only
& 0.069G & 1003 & 0.98G
& 0.069G & 988 & 0.98G \\

Dissolve GS Only
& 0.016G & 1110 & 0.09G
& 0.016G & 1111 & 0.09G \\

Residual GS Only
& 0.251G & 265 & 0.77G
& 0.134G & 463 & 0.49G \\

\midrule

GS Baseline
& 0.655G & 99 & \textbf{1.64G}
& \textbf{0.071G} & \textbf{311} & \textbf{0.48G} \\

\midrule

Ours
& \textbf{0.336G} & 168 & 1.70G
& 0.219G & 236 & 1.42G \\

\rowcolor{gray!10}
Ours (Parallel)
& \textbf{0.336G} & \textbf{174} & 1.71G
& 0.219G & 241 & 1.43G \\

\bottomrule
\end{tabularx}
\end{table}

\maybeinput{table3}{Original ablation and efficiency tables.}

\begin{table}[t]
\centering
\scriptsize
\setlength{\tabcolsep}{2.1pt}
\renewcommand{\arraystretch}{1.08}

\caption{\textbf{Depth consistency with the aligned DA-V2 reference.}
We report the standard AbsRel, RMSE, and $\delta_1$ metrics used by Depth Anything V2~\cite{yang2024depthanythingv2}, together with Boundary F1 to measure depth-discontinuity alignment. Boundary F1 uses a shared Sobel threshold and a two-pixel matching tolerance.}
\label{tab:depth_reference_consistency}

\begin{tabularx}{\linewidth}{@{}l|YYYY@{}}
\toprule
Method & AbsRel$\downarrow$ & RMSE$\downarrow$
& $\delta_1\uparrow$ & Boundary F1$\uparrow$
\\
\midrule

Mesh
& \cellcolor{second}0.0787
& 0.0909
& 0.9725
& 0.5274
\\

Hard-Z
& 0.0868
& \cellcolor{second}0.0845
& \cellcolor{second}0.9736
& \cellcolor{second}0.6720
\\

\midrule
\rowcolor{gray!10}
\textbf{Ours}
& \cellcolor{best}\textbf{0.0680}
& \cellcolor{best}\textbf{0.0785}
& \cellcolor{best}\textbf{0.9751}
& \cellcolor{best}\textbf{0.6814}
\\

\bottomrule
\end{tabularx}
\end{table}

\subsection{Subtract-and-Refill Analysis}
\label{sec:erase_refill_analysis}

\gaussolve{} separates complementary roles between the textured mesh and the dissolve-and-refill branches. Fig.~\ref{decomposition} visualizes this learned division of labor: the mesh branch explains regular structures, the dissolve field concentrates on local mesh--image conflicts, and residual RGB Gaussians allocate capacity to the revealed regions and remaining fine details. This decomposition supports the subtract-and-refill interpretation: \gaussolve{} selectively attenuates inconsistent mesh content and refills the exposed regions, rather than simply adding another Gaussian layer on top of the mesh.

To further examine the subtract-and-refill behavior, Fig.~\ref{fig:depth_consistency} compares the aligned DA-V2 reference with the mesh depth and rendered depths from Hard-Z and \gaussolve{}. Hard-Z largely follows the mesh depth, whereas \gaussolve{} suppresses inconsistent mesh evidence and allows residual Gaussians to explain the revealed regions. Table~\ref{tab:depth_reference_consistency} further shows that \gaussolve{} achieves the best depth consistency and boundary accuracy across all four metrics.

Fig.~\ref{fig:ablation_sh} isolates the view-dependent part of the dissolve field. The View-invariant variant, without scalar SH, cannot follow direction-specific appearance residuals. With scalar SH, it can attenuate mesh content only from conflicting viewing directions, allowing residual RGB Gaussians to recover highlights and reflection-like details when needed.

\subsection{Ablations}
\label{sec:ablation_quantitative}

Table~\ref{tab:ablation} summarizes the ablations on ML-Hypersim-001.
The component and training-design ablations use the RC mesh. The full model
achieves 35.21 PSNR and 0.106 LPIPS, outperforming the residual-GS-only and
Hard-Z variants. Removing scalar SH, dual densification, or progressive
initialization also degrades reconstruction quality, validating the proposed
training design. For the geometry and appearance ablation, we use the PGSR
mesh. Performance remains stable from 10K to 1M faces, while at around 100K
faces, direct projection reduces LPIPS from 0.140 to 0.112 compared with baked
appearance while maintaining similar PSNR and SSIM. Furthermore, the consistent performance across various resolutions and appearance constructions indicates that Gaussolve does not depend on a narrowly tuned mesh configuration.

\begin{table}[t]
\centering
\scriptsize
\setlength{\tabcolsep}{2.8pt}
\renewcommand{\arraystretch}{1.08}

\caption{\textbf{Training overhead.}
Training time and peak GPU memory for Gaussolve with different Gaussian backbones.}
\label{tab:training-overhead}

\begin{tabularx}{\linewidth}{@{}l|YY|YY|YY|Y@{}}
\toprule
\multirow{2}{*}{Method}
& \multicolumn{2}{c|}{\textbf{3DGS}}
& \multicolumn{2}{c|}{\textbf{Reduced 3DGS}}
& \multicolumn{2}{c|}{\textbf{3DGS-MCMC}}
& \multicolumn{1}{c}{\textbf{Mesh}} \\
\cmidrule(lr){2-3}
\cmidrule(lr){4-5}
\cmidrule(lr){6-7}
\cmidrule(lr){8-8}

& Time & Mem.
& Time & Mem.
& Time & Mem.
& Mem. \\
\midrule

Baseline
& 32.5min & 3.67G
& 24.5min & 1.51G
& 36.0min & 4.11G
& -- \\

Ours
& 32.9min & 4.91G
& 34.1min & 4.78G
& 53.1min & 6.95G
& 3.95G \\

\bottomrule
\end{tabularx}
\end{table}

\subsection{Rendering and Training Efficiency}
\label{sec:ablation_quantitative}

Table~\ref{tab:efficiency-modules} reports the average efficiency. With the 3DGS backbone, Ours (Parallel) improves rendering speed from 99 to 174 FPS while reducing storage from 0.655G to 0.336G. Replacing the Gaussian backbone with Reduced 3DGS further increases our rendering speed to 241 FPS and reduces storage to 0.219G. Although the hybrid remains slower than standalone Reduced 3DGS due to the additional mesh and NDF rendering overhead, it achieves substantially better reconstruction quality. Introducing the mesh also incurs additional training cost (Table~\ref{tab:training-overhead}), but the overhead remains within a manageable range, as the mesh itself has a nearly constant memory footprint independent of the Gaussian population.

\vfill
\section{Conclusion}
\label{sec:conclusion}

We presented \gaussolve{}, a hybrid Mesh--Gaussian renderer that reformulates representation conflicts as multi-view inconsistencies. \gaussolve{} selectively dissolves inconsistent mesh content with a Neural Dissolve Field and lets residual Gaussians refill the revealed regions. Experiments on structured and semi-structured scenes demonstrate improved visual quality across different Gaussian backbones, together with substantial Gaussian reduction and real-time rendering.

\clearpage
\raggedbottom

{\small
\bibliographystyle{ieeenat_fullname}
\bibliography{main}
}

\clearpage
\flushbottom
\makeatletter
\def\@title{\thetitle}
\def\@author{%
  \vphantom{Yiyan Wang$^{1,*}$ Haohan Liu$^{2,*}$ Tetsuo Shibuya$^{1}$}Supplementary Material\\
  \vphantom{$^{1}$The University of Tokyo, Japan $^{2}$Fuzhou University, China $^{*}$Equal contribution}\\
  {\tt\small\vphantom{5770716834@utac.u-tokyo.ac.jp 852203208@fzu.edu.cn tshibuya@hgc.jp}}}
\twocolumn[\suppmaketitle]
\makeatother
\setcounter{section}{0}
\setcounter{figure}{0}
\setcounter{table}{0}
\setcounter{equation}{0}
\setcounter{algorithm}{0}
\renewcommand{\thesection}{S\arabic{section}}
\renewcommand{\thefigure}{S\arabic{figure}}
\renewcommand{\thetable}{S\arabic{table}}
\renewcommand{\theequation}{S\arabic{equation}}
\renewcommand{\thealgorithm}{S\arabic{algorithm}}
\renewcommand{\theHsection}{S\arabic{section}}
\renewcommand{\theHfigure}{S\arabic{figure}}
\renewcommand{\theHtable}{S\arabic{table}}
\renewcommand{\theHequation}{S\arabic{equation}}
\renewcommand{\theHalgorithm}{S\arabic{algorithm}}
\section{Additional Method Details}

\subsection{Fixed Mesh Depth in Continuous Fusion}
A possible extension of dissolution is to recompute a new depth map after the mesh has been partially dissolved, and then use this updated depth to separate front and back Gaussian contributions. We do not adopt this design. Once dissolution becomes soft, a unique post-dissolution depth is no longer naturally defined. Any such depth would require an additional heuristic, such as thresholding a soft visibility profile or defining an expected depth along the ray. This would introduce extra hyperparameters and make the front/back partition itself depend on the dissolve prediction.

More importantly, such a design creates an undesirable feedback loop. If dissolution changes the depth, then the updated depth changes which Gaussians are assigned to the front or back segment. Those changed Gaussian contributions in turn modify the gradients that supervise the dissolve field. As a result, the partition boundary itself becomes unstable near occlusion transitions.

For this reason, \gaussolve{} always uses the rasterized mesh depth $Z_m(\mathbf{u})$ as a fixed depth for the Gaussian-center front/back partition. This depth is used only to compute the front-segment statistics $(C_f,T_f)$ required by the decomposition in the main paper. The dissolve field does not alter $Z_m(\mathbf{u})$. Instead, it only modulates the effective mesh opacity
\begin{equation}
\alpha_m^{\mathrm{eff}}(\mathbf{u}) = 1-D(\mathbf{u}).
\end{equation}
Thus the mesh depth remains fixed for the front/back decomposition, while dissolution only changes the final blending weight.

\subsection{Parallel Rendering Implementation}
For each viewpoint, the mesh is rasterized once to obtain the reference depth map $Z_m(\mathbf{u})$, the visible mesh mask, and screen-space coordinates. After this step, the mesh, dissolve, and residual Gaussian branches become largely independent until the final compositing stage. The mesh branch computes the mesh color $C_m(\mathbf{u})$; the dissolve branch renders $D(\mathbf{u})$ and converts it into $\alpha_m^{\mathrm{eff}}(\mathbf{u})=1-D(\mathbf{u})$; and the residual Gaussian branch uses the fixed mesh depth to compute $(C_f,T_f)$ and $C_g$. The final image is
\begin{equation}
\begin{split}
I_{\mathrm{hybrid}}(\mathbf{u}) &= \alpha_m^{\mathrm{eff}}(\mathbf{u})(C_f(\mathbf{u})+T_f(\mathbf{u})C_m(\mathbf{u})) \\ 
&\quad + (1-\alpha_m^{\mathrm{eff}}(\mathbf{u}))C_g(\mathbf{u}).
\end{split}
\end{equation}
After the shared mesh rasterization and depth computation, the three branch computations are scheduled asynchronously on separate CUDA streams.

\subsection{Multi-view Mesh-Texel Initialization}

We initialize the NDF around mesh regions that consistently disagree with
the training images. Specifically, we use the mesh UV atlas as a common domain for aggregating errors across
views by rendering the pretrained textured mesh from every
training view and transferring the resulting RGB reconstruction errors to the
corresponding mesh texels using the rasterized UV coordinates.

For a texel $q$, let $\mathcal{V}_q$ denote the training views in which it is
visible, and let $e_v(q)$ denote its mean RGB error in view $v$. Its
multi-view error is

\begin{equation}
E(q)=\frac{1}{|\mathcal{V}_q|}
\sum_{v\in\mathcal{V}_q} e_v(q).
\end{equation}

We select texels with high multi-view error that are observed from
multiple training views. For a selected texel $q$, let $\mathbf{x}_1,\mathbf{x}_2,\mathbf{x}_3$
denote the vertices of its mesh triangle, and let
$\beta_1(q),\beta_2(q),\beta_3(q)$ be the corresponding barycentric
coordinates. Its 3D surface position is

\begin{equation}
\mathbf{x}(q)=
\beta_1(q)\mathbf{x}_1+
\beta_2(q)\mathbf{x}_2+
\beta_3(q)\mathbf{x}_3.
\end{equation}

To remove UV-seam duplicates and overly dense samples, each position is
assigned to a 3D voxel,

\begin{equation}
\mathbf{k}(q)=
\left\lfloor
\frac{\mathbf{x}(q)-\mathbf{x}_{\min}}{\delta}
\right\rfloor,
\end{equation}

where $\delta$ is the voxel size and $\mathbf{x}_{\min}$ is the minimum
corner of the selected point set. Candidates assigned to the same voxel are
merged, retaining the one with the strongest multi-view support. A scalar-SH dissolve Gaussian is initialized
at each retained position. These Gaussians are subsequently fully trainable
and are not constrained to remain on the mesh surface.

\begin{figure*}[p]
\centering

\providecommand{\setboxes}[8]{%
  \xdef\RxA{#1}\xdef\RyA{#2}\xdef\RxB{#3}\xdef\RyB{#4}%
  \xdef\BxA{#5}\xdef\ByA{#6}\xdef\BxB{#7}\xdef\ByB{#8}%
}

\newcommand{\bluecrop}[2]{%
  {\setlength{\fboxsep}{0pt}\setlength{\fboxrule}{1pt}%
  \fcolorbox{blue}{white}{\includegraphics[viewport={\BxA} {\ByA} {\BxB} {\ByB}, clip, width=#2]{#1}}}%
}
\newcommand{\redcrop}[2]{%
  {\setlength{\fboxsep}{0pt}\setlength{\fboxrule}{1pt}%
  \fcolorbox{red}{white}{\includegraphics[viewport={\RxA} {\RyA} {\RxB} {\RyB}, clip, width=#2]{#1}}}%
}

\newcommand{\twocrops}[3]{%
  \redcrop{#1}{#3}\\[2pt]\bluecrop{#2}{#3}%
}

\newcommand{\annotatedimg}[2]{%
  \resizebox{#2}{!}{%
    \begin{tikzpicture}[x=1bp,y=1bp]
      \node[anchor=south west,inner sep=0] (img) at (0,0) {\includegraphics{#1}};
      \useasboundingbox (0,0) rectangle (img.north east);
      \draw[red, ultra thick] (\RxA,\RyA) rectangle (\RxB,\RyB);
      \draw[blue, ultra thick] (\BxA,\ByA) rectangle (\BxB,\ByB);
    \end{tikzpicture}%
  }%
}

\setlength{\tabcolsep}{2pt}

\setboxes{80}{120}{305}{300}{200}{550}{400}{710}

\begin{tabular}{c cccc}
 & Mesh/GT & Texture $\alpha$ / blend & Vertex $\alpha$ / blend & Ours / blend \\

\multirow{3}{*}{\rotatebox{90}{Scene 1}} &
\includegraphics[width=0.232\textwidth]{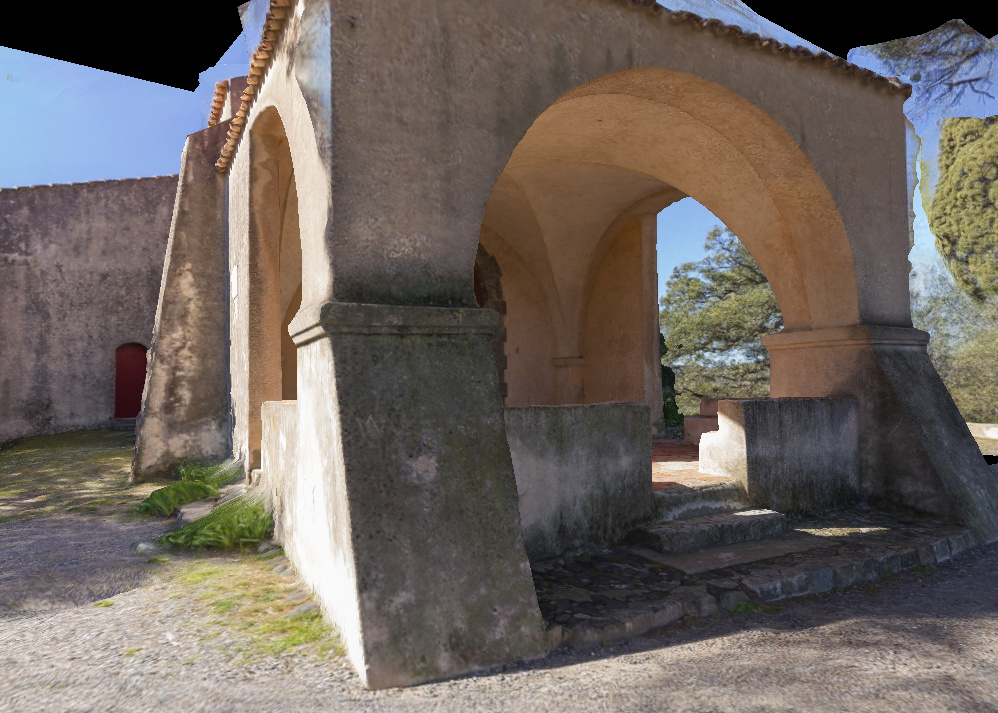} &
\includegraphics[width=0.232\textwidth]{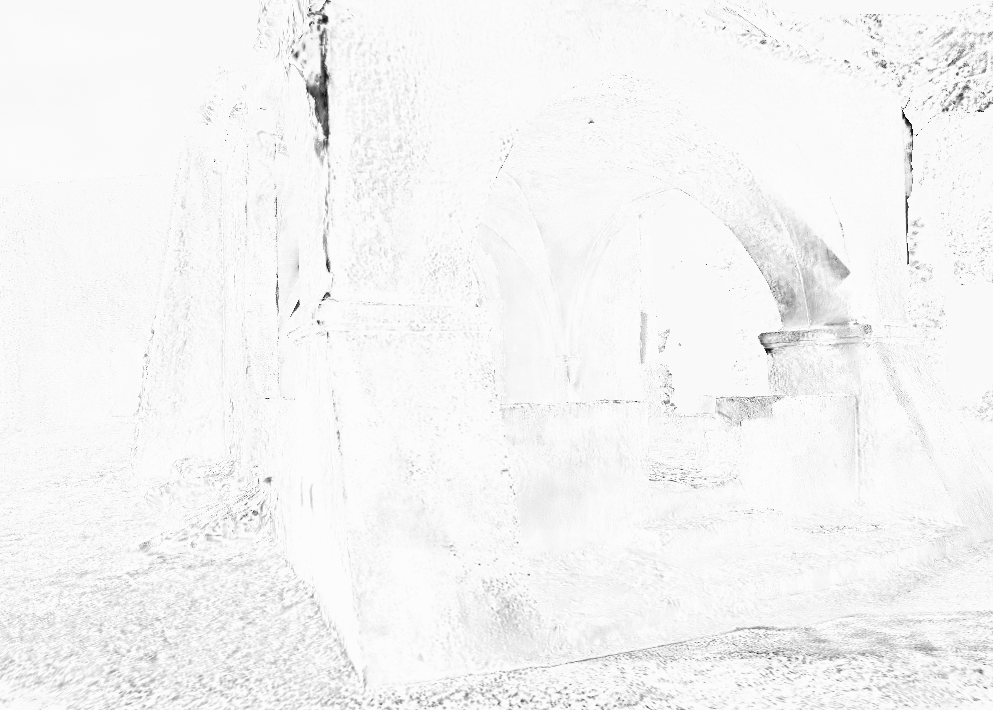} &
\includegraphics[width=0.232\textwidth]{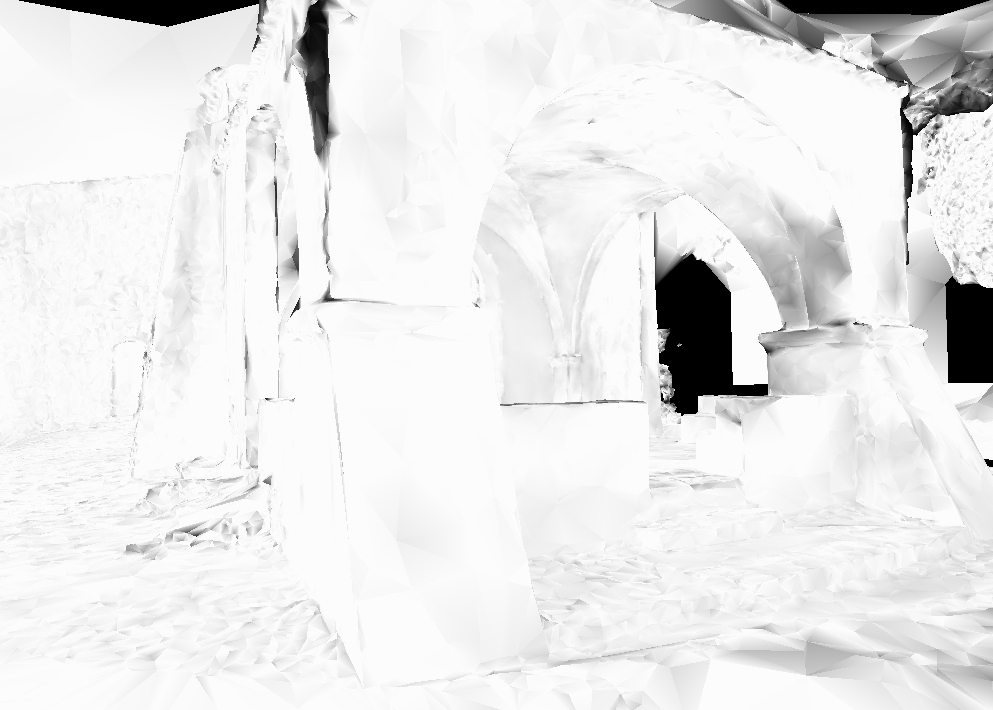} &
\includegraphics[width=0.232\textwidth]{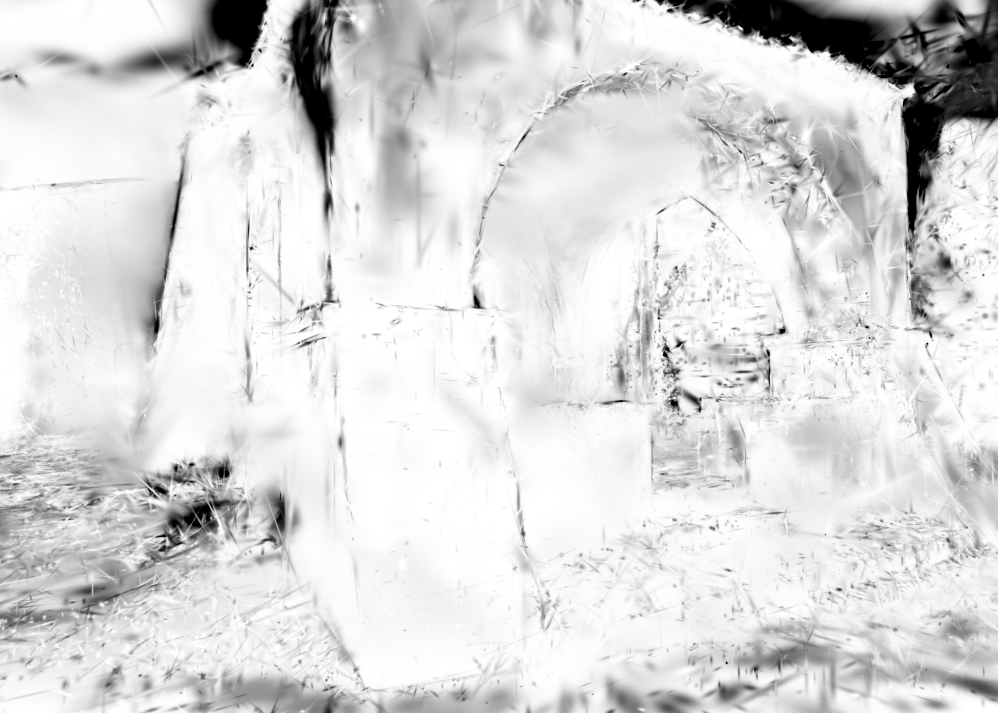} \\

&
\annotatedimg{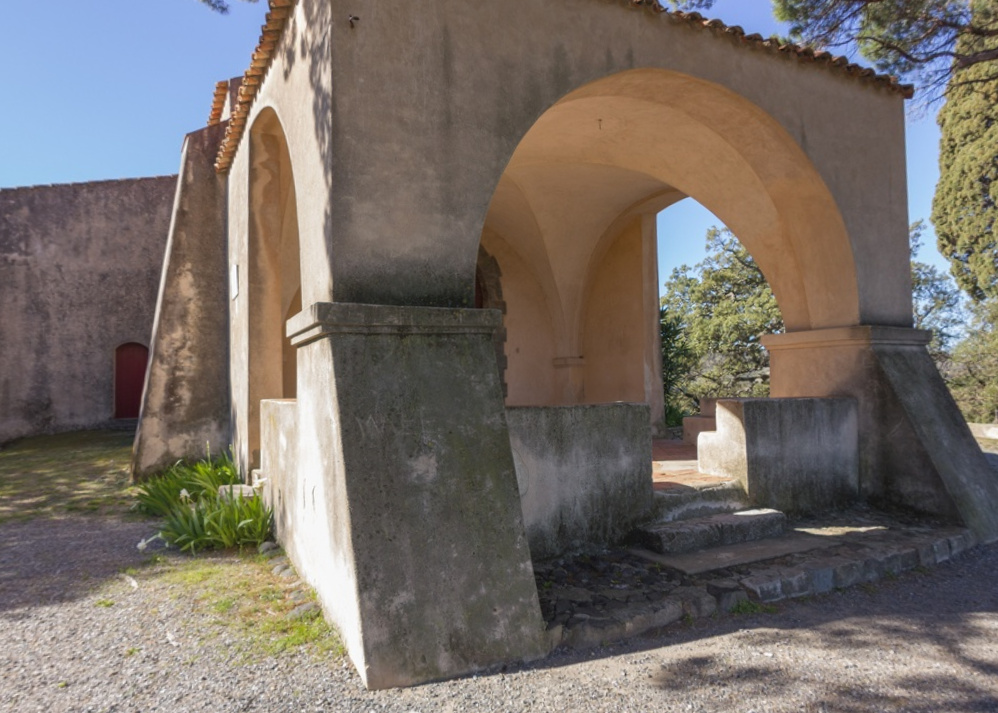}{0.232\textwidth} &
\annotatedimg{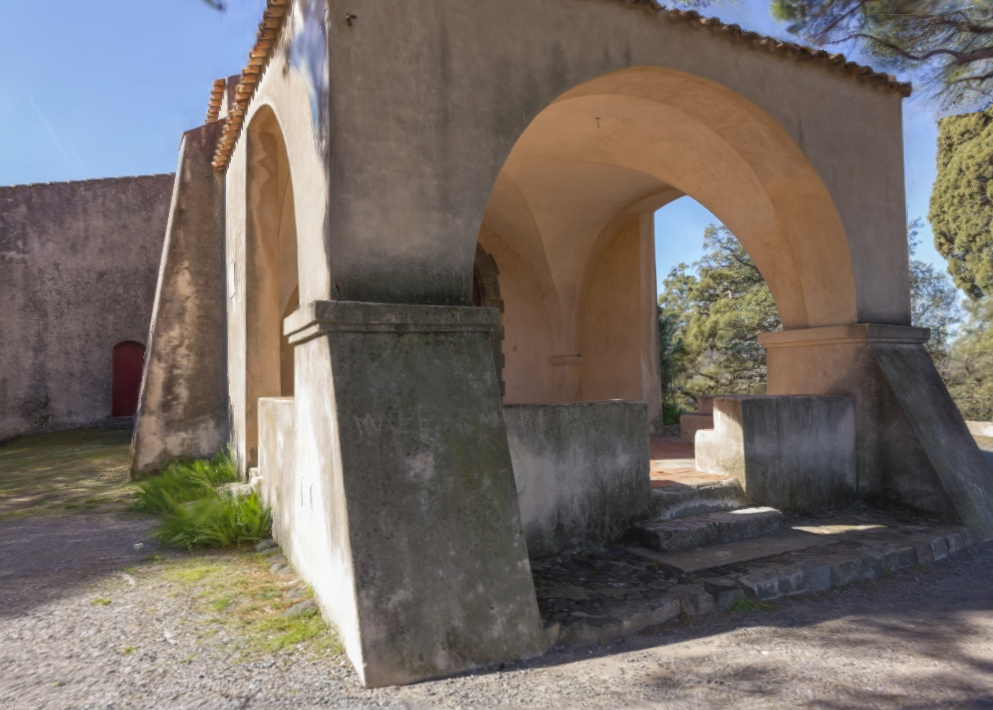}{0.232\textwidth} &
\annotatedimg{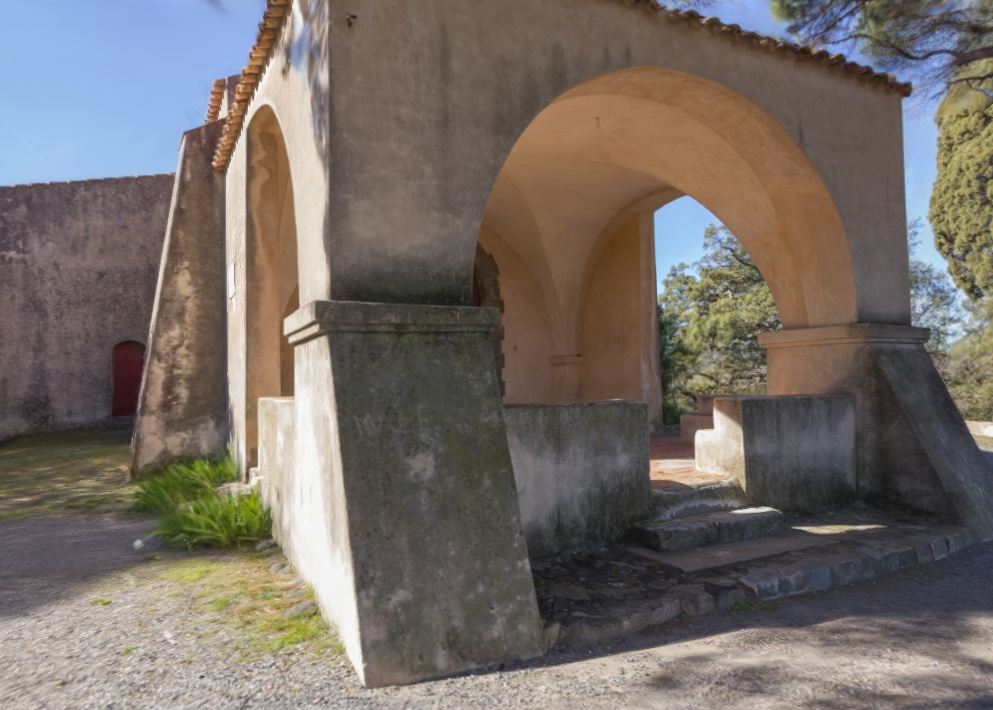}{0.232\textwidth} &
\annotatedimg{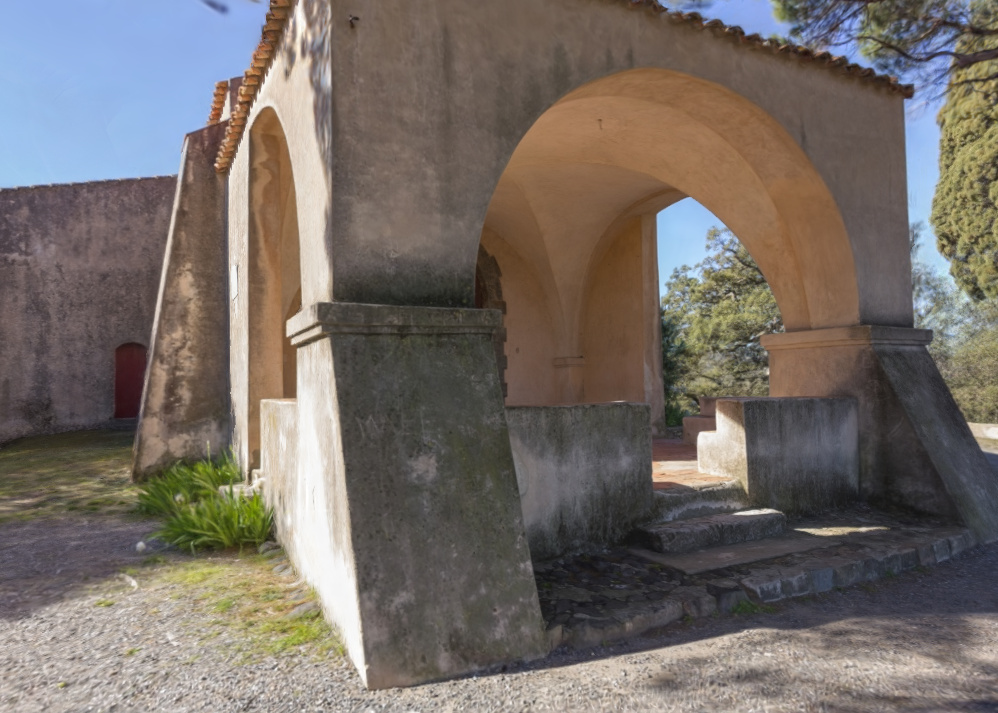}{0.232\textwidth} \\

&
\begin{minipage}{0.232\textwidth}\centering
\redcrop{images/exp5/scene1/gt_.jpg}{0.6\linewidth}\\[2pt]%
\bluecrop{images/exp5/scene1/gt_.jpg}{0.6\linewidth}
\end{minipage}
&
\begin{minipage}{0.232\textwidth}\centering
\redcrop{images/exp5/scene1/tex_blend_.jpg}{0.6\linewidth}\\[2pt]%
\bluecrop{images/exp5/scene1/tex_blend_.jpg}{0.6\linewidth}
\end{minipage}
&
\begin{minipage}{0.232\textwidth}\centering
\redcrop{images/exp5/scene1/vex_blend_.jpg}{0.6\linewidth}\\[2pt]%
\bluecrop{images/exp5/scene1/vex_blend_.jpg}{0.6\linewidth}
\end{minipage}
&
\begin{minipage}{0.232\textwidth}\centering
\redcrop{images/exp5/scene1/gs_blend_.jpg}{0.6\linewidth}\\[2pt]%
\bluecrop{images/exp5/scene1/gs_blend_.jpg}{0.6\linewidth}
\end{minipage}
\\
\end{tabular}

\vspace{1mm}

\setboxes{200}{350}{450}{530}{500}{350}{750}{530}

\begin{tabular}{c cccc}

\multirow{3}{*}{\rotatebox{90}{Scene 2}} &
\includegraphics[width=0.232\textwidth]{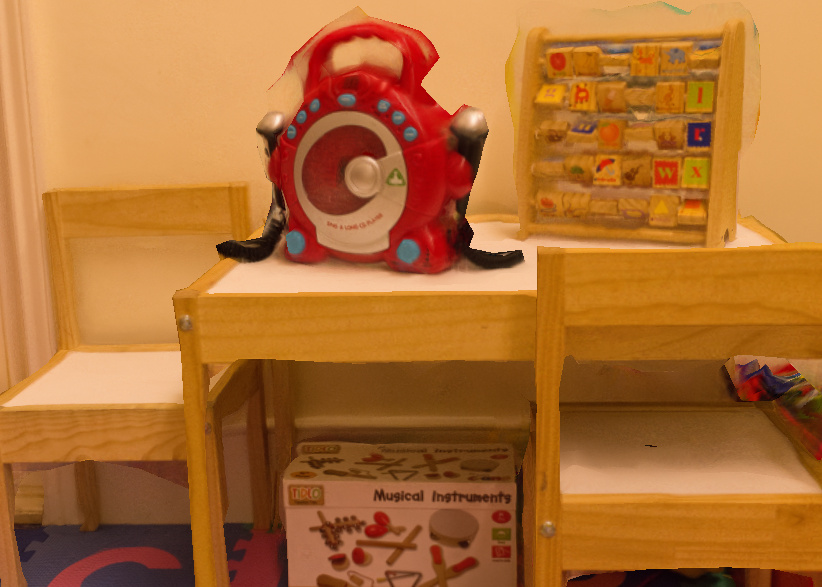} &
\includegraphics[width=0.232\textwidth]{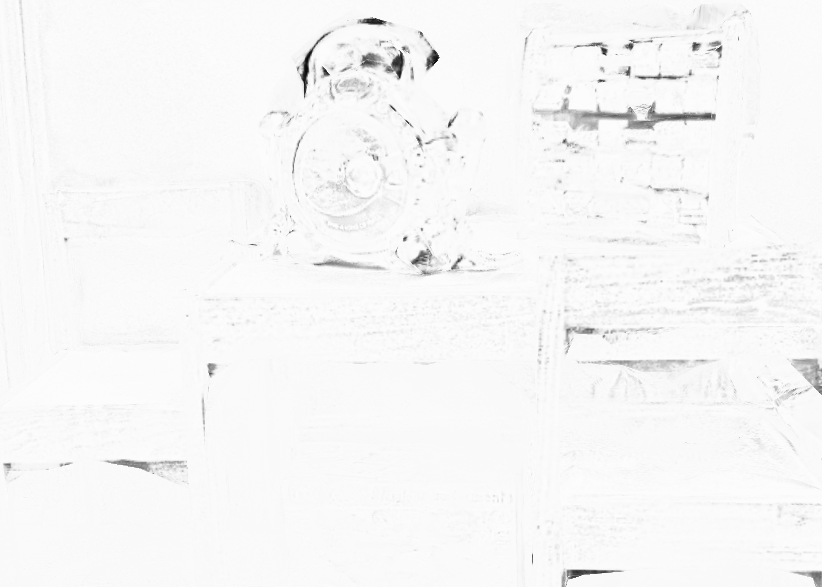} &
\includegraphics[width=0.232\textwidth]{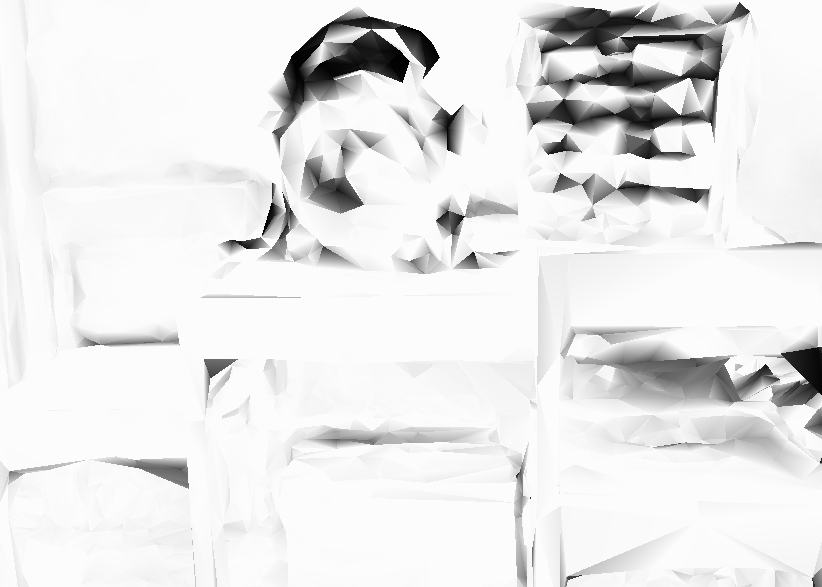} &
\includegraphics[width=0.232\textwidth]{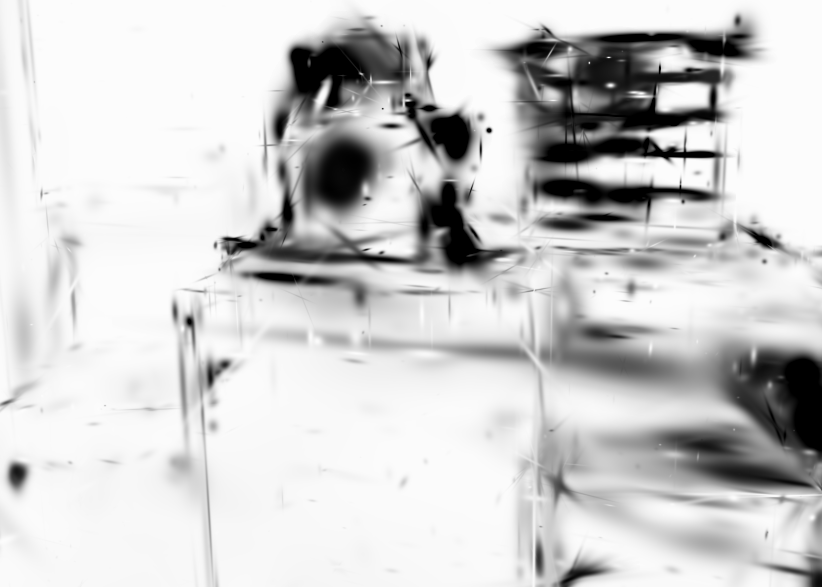} \\

&
\annotatedimg{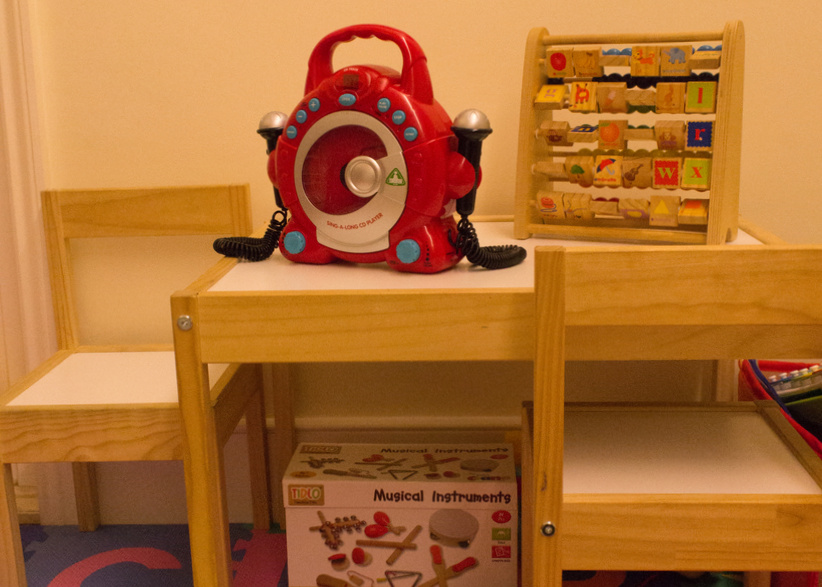}{0.232\textwidth} &
\annotatedimg{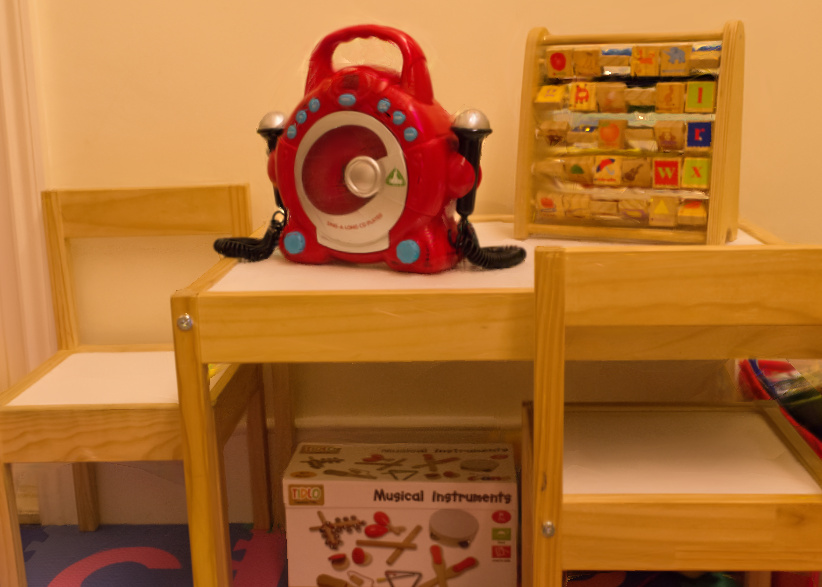}{0.232\textwidth} &
\annotatedimg{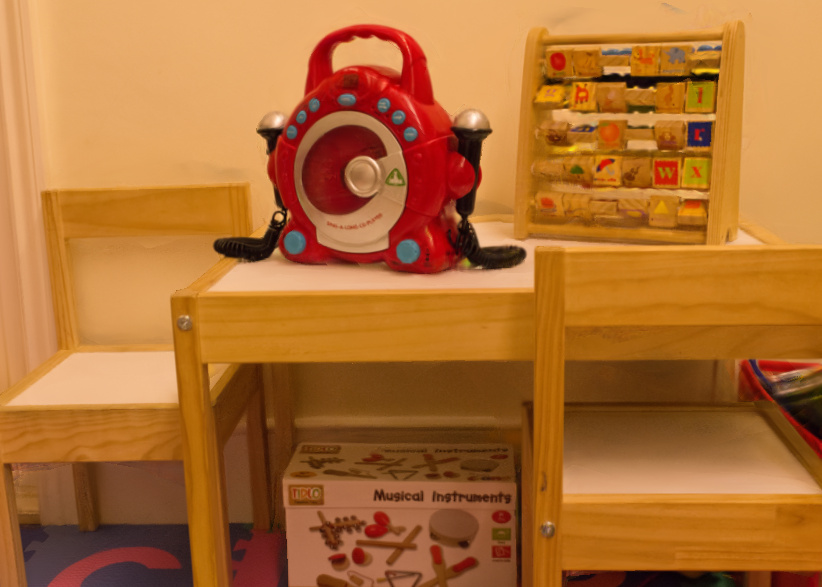}{0.232\textwidth} &
\annotatedimg{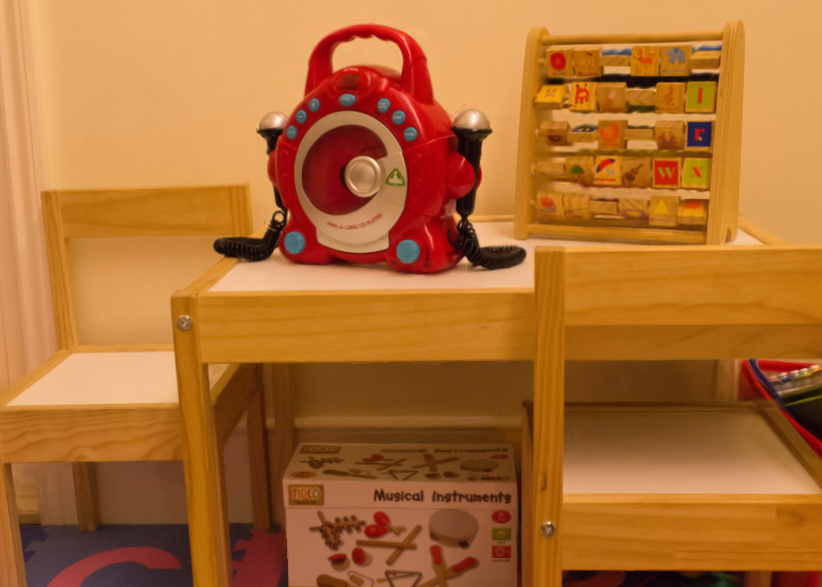}{0.232\textwidth} \\

&
\begin{minipage}{0.232\textwidth}\centering
\redcrop{images/exp5/scene2/gt_.jpg}{0.6\linewidth}\\[2pt]%
\bluecrop{images/exp5/scene2/gt_.jpg}{0.6\linewidth}
\end{minipage}
&
\begin{minipage}{0.232\textwidth}\centering
\redcrop{images/exp5/scene2/tex_blend_.jpg}{0.6\linewidth}\\[2pt]%
\bluecrop{images/exp5/scene2/tex_blend_.jpg}{0.6\linewidth}
\end{minipage}
&
\begin{minipage}{0.232\textwidth}\centering
\redcrop{images/exp5/scene2/vet_blend_.jpg}{0.6\linewidth}\\[2pt]%
\bluecrop{images/exp5/scene2/vet_blend_.jpg}{0.6\linewidth}
\end{minipage}
&
\begin{minipage}{0.232\textwidth}\centering
\redcrop{images/exp5/scene2/gs_blend_.jpg}{0.6\linewidth}\\[2pt]%
\bluecrop{images/exp5/scene2/gs_blend_.jpg}{0.6\linewidth}
\end{minipage}
\\

\end{tabular}

\caption{\textbf{Surface-alpha dissolve alternatives.}
We compare texture-map alpha, per-vertex alpha, and the proposed volumetric NDF. The first row of each scene shows the learned mesh/dissolve representation, the second row shows the blended rendering with red and blue boxes, and the third row shows the corresponding zoom-ins. Texture-map and vertex alpha can reduce some mesh--image conflicts but often produce diffuse or fragmented masks, while the volumetric NDF localizes unreliable mesh evidence and enables cleaner residual Gaussian refill.}
\label{fig:exp5_visual_zoom}
\end{figure*}

\begin{algorithm}[t]
\caption{Parallel Differentiable Hybrid Rendering}
\label{alg:rendering}
\begin{algorithmic}[1]
\STATE \textbf{Input:} viewpoint $\mathbf{v}$, mesh prior $\mathcal{M}$, residual Gaussians $\mathcal{G}_r$, dissolve Gaussians $\mathcal{G}_d$
\STATE Rasterize $\mathcal{M}$ to obtain mesh depth $Z_m(\mathbf{u})$, visibility mask, and screen-space coordinates.
\STATE Launch three asynchronous branches.

\begin{adjustwidth}{0.5em}{0pt}
· Mesh branch: render the textured mesh color $C_m(\mathbf{u})$.

· Dissolve branch: rasterize $\mathcal{G}_d$ to obtain $D(\mathbf{u})$ and $ \alpha_m^{\mathrm{eff}}(\mathbf{u})=1-D(\mathbf{u})$.

· Residual branch: render $\mathcal{G}_r$ with fixed mesh depth to obtain $C_f$, $T_f$, and $C_g$.
\end{adjustwidth}
\STATE Synchronize all streams and composite $I_{\mathrm{hybrid}} = \alpha_m^{\mathrm{eff}}(C_f+T_fC_m)+(1-\alpha_m^{\mathrm{eff}})C_g$.
\end{algorithmic}
\end{algorithm}

\begin{figure*}[t]
\centering

\newcommand{\vcenterbox}[1]{\raisebox{-0.5\height}{#1}}

\IfFileExists{consistency/hy/1_cropped.jpg}{%
\setlength{\tabcolsep}{1pt}
\renewcommand{\arraystretch}{0.70}

\begin{tabular}{@{}cccccc@{}}

\vcenterbox{\rotatebox{90}{ML-Hy-001}} &
\vcenterbox{\includegraphics[width=0.19\linewidth]{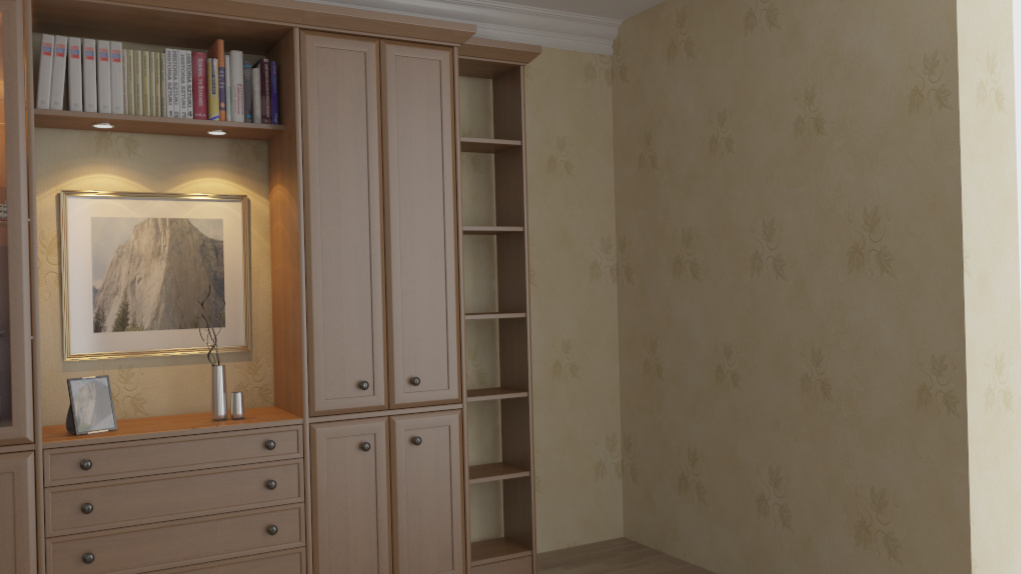}} &
\vcenterbox{\includegraphics[width=0.19\linewidth]{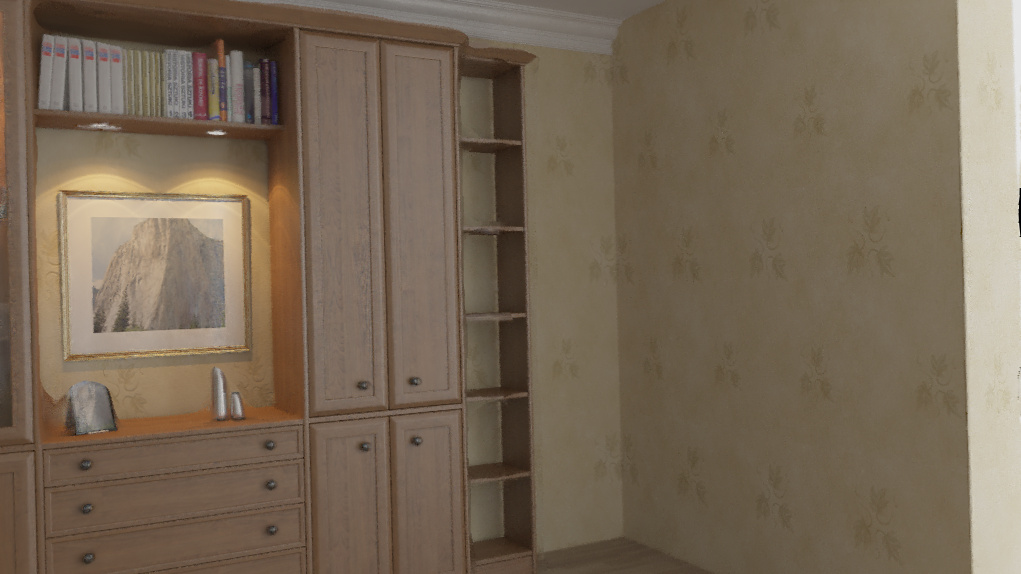}} &
\vcenterbox{\includegraphics[width=0.19\linewidth]{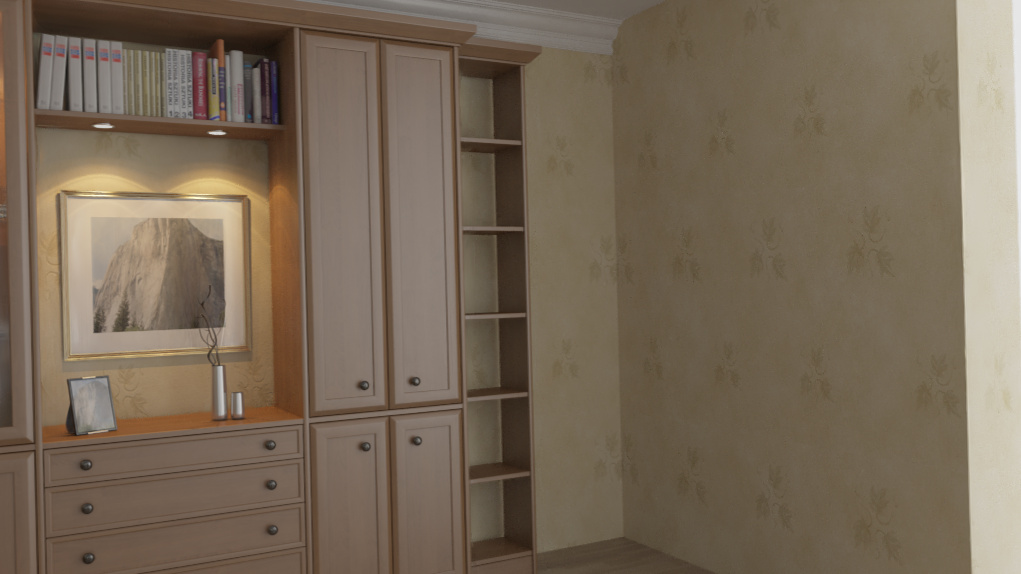}} &
\vcenterbox{\includegraphics[width=0.19\linewidth]{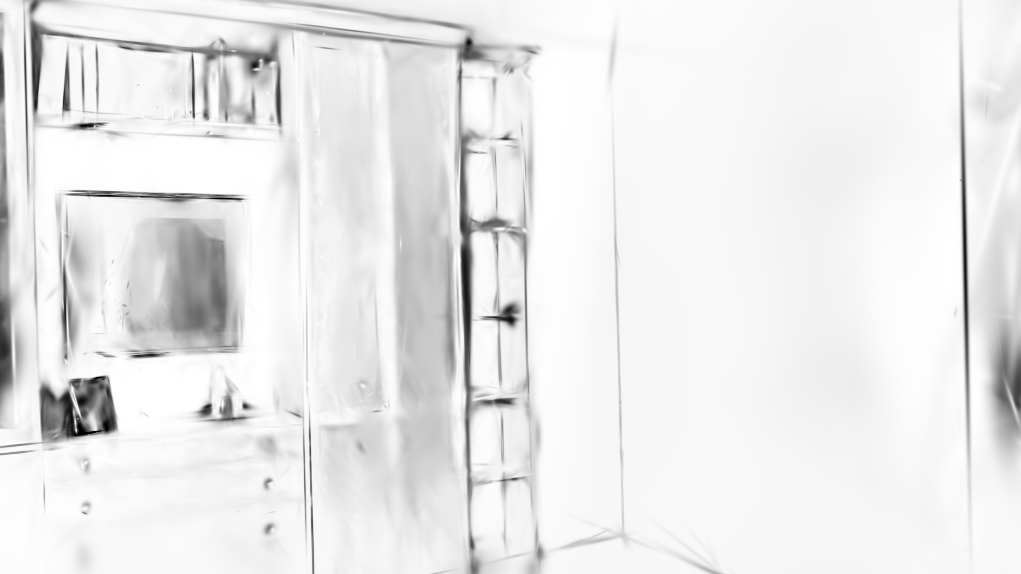}} &
\vcenterbox{\includegraphics[width=0.19\linewidth]{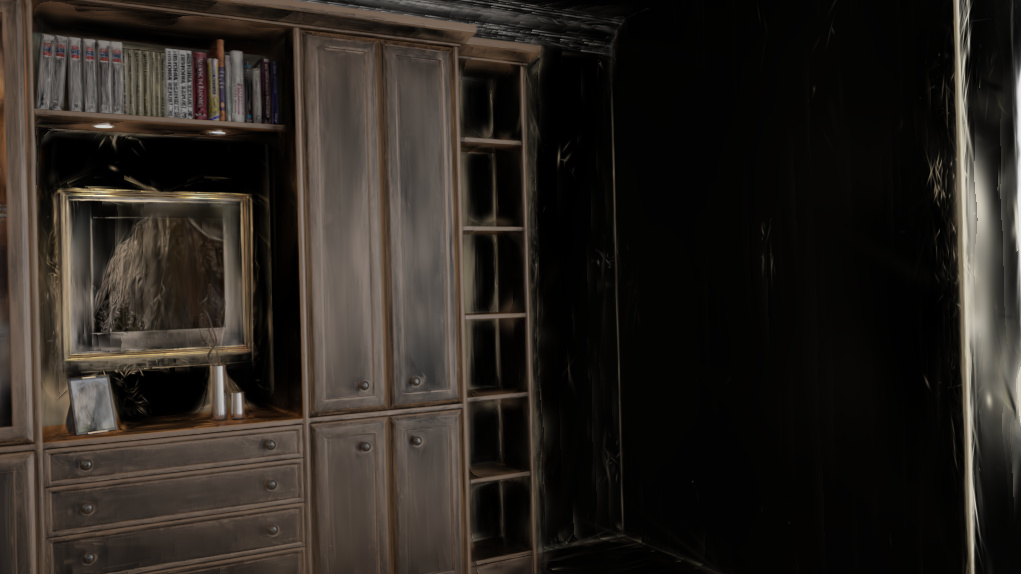}}
\\[28pt]

\vcenterbox{\rotatebox{90}{ML-Hy-035}} &
\vcenterbox{\includegraphics[width=0.19\linewidth]{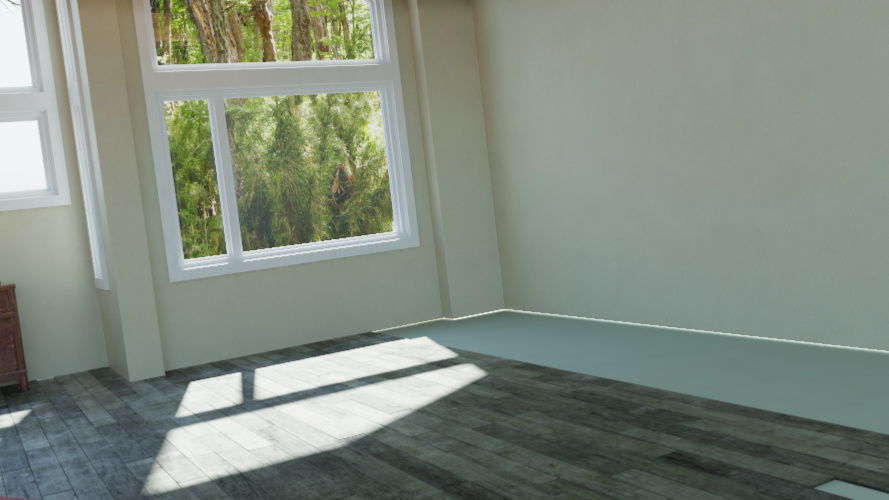}} &
\vcenterbox{\includegraphics[width=0.19\linewidth]{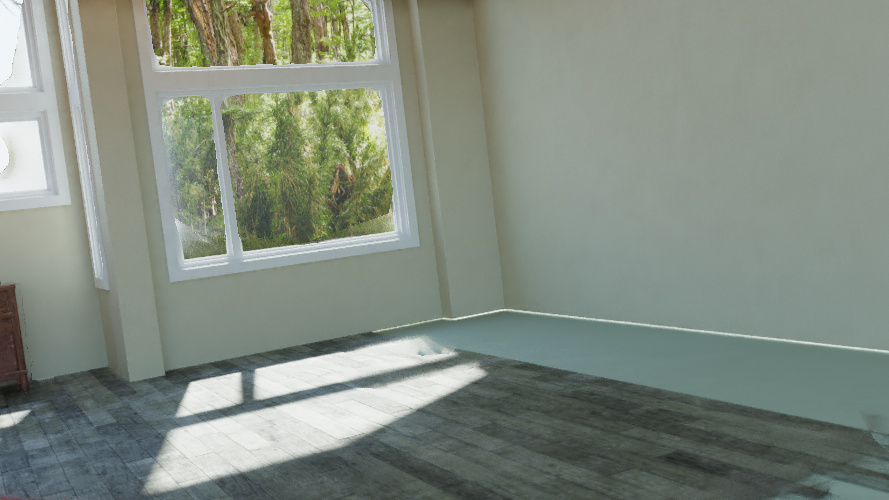}} &
\vcenterbox{\includegraphics[width=0.19\linewidth]{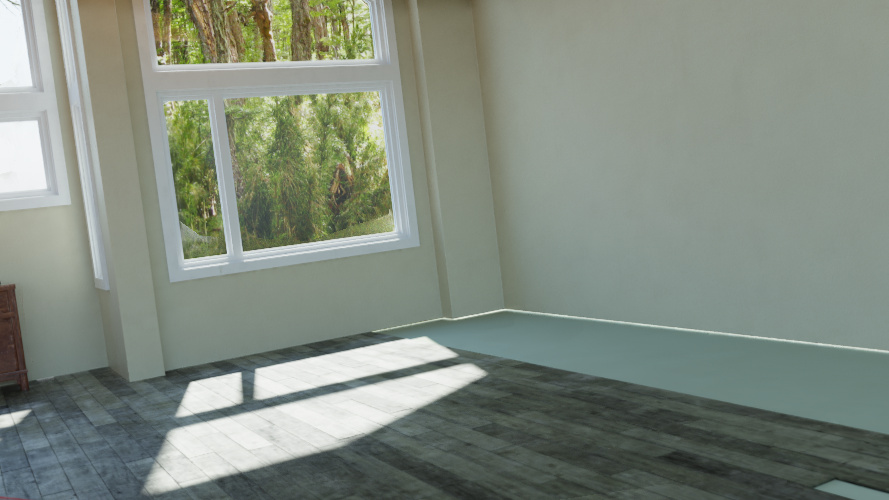}} &
\vcenterbox{\includegraphics[width=0.19\linewidth]{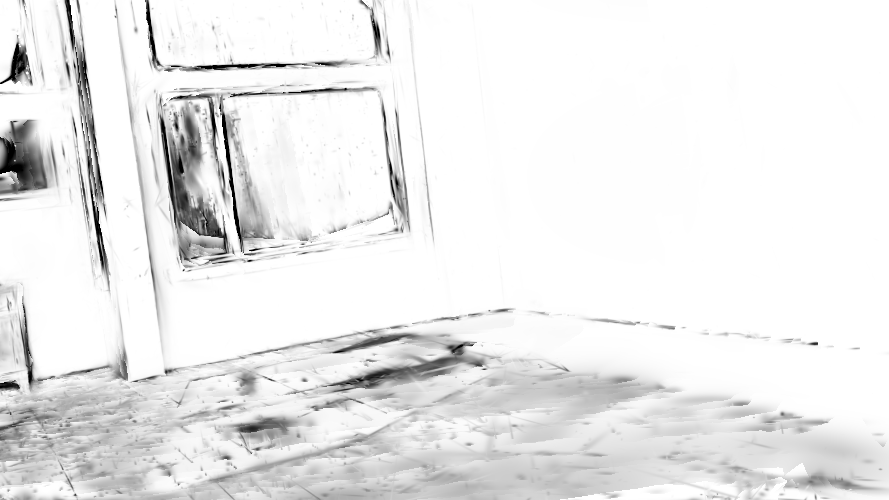}} &
\vcenterbox{\includegraphics[width=0.19\linewidth]{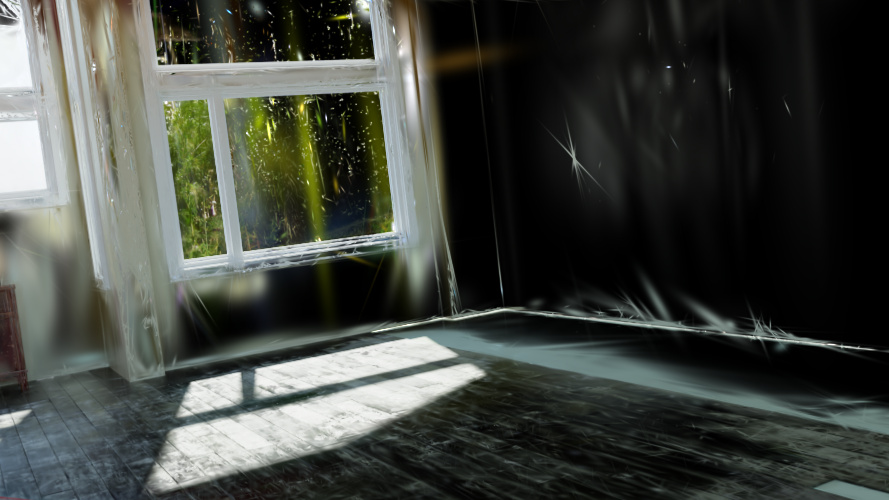}}
\\[28pt]

\vcenterbox{\rotatebox{90}{ML-Hy-039}} &
\vcenterbox{\includegraphics[width=0.19\linewidth]{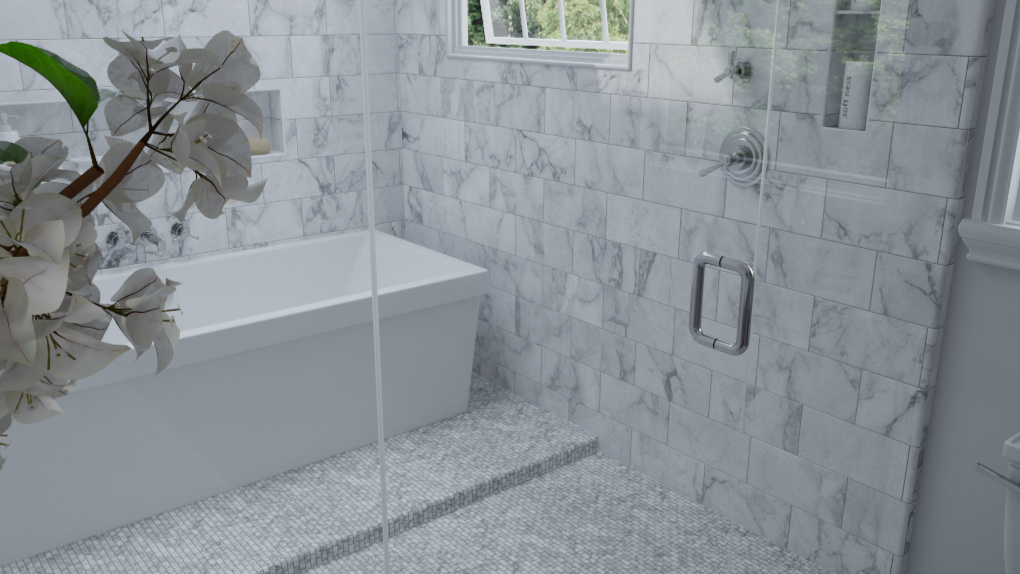}} &
\vcenterbox{\includegraphics[width=0.19\linewidth]{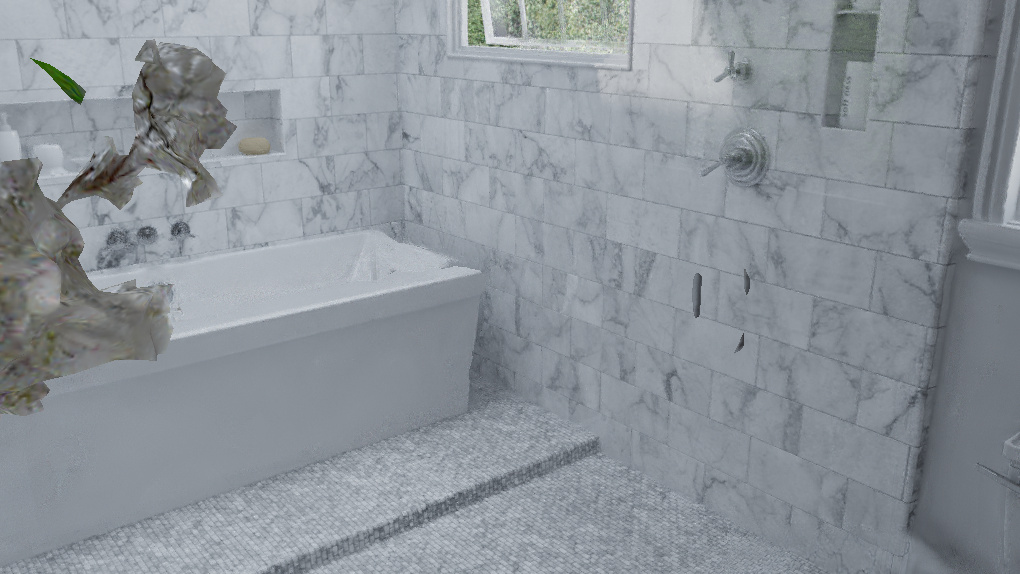}} &
\vcenterbox{\includegraphics[width=0.19\linewidth]{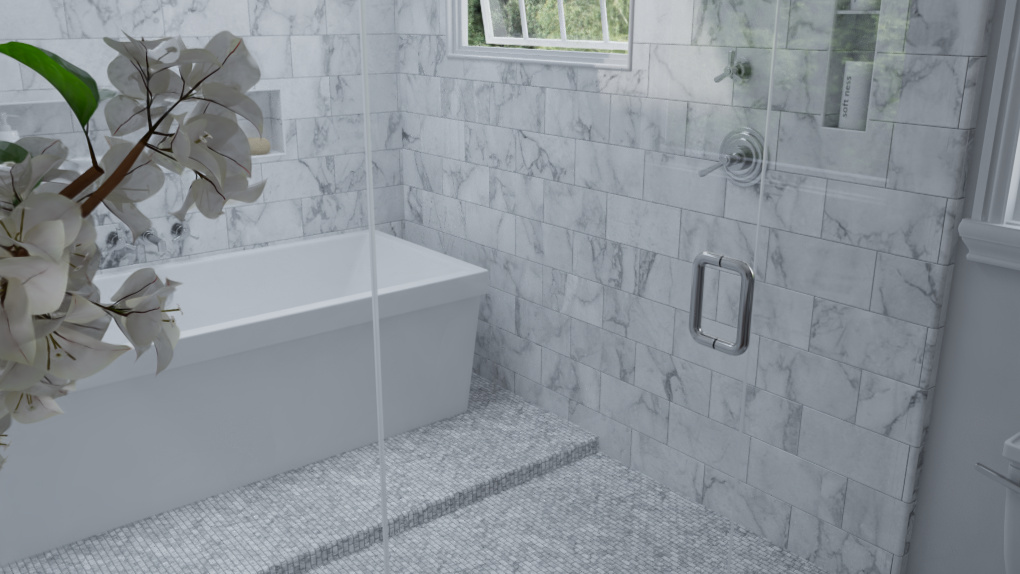}} &
\vcenterbox{\includegraphics[width=0.19\linewidth]{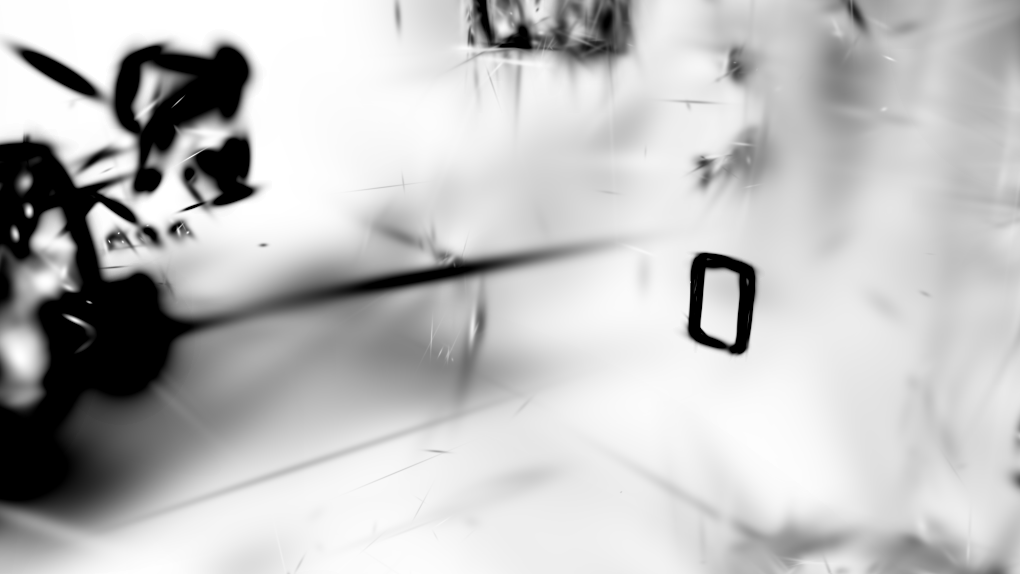}} &
\vcenterbox{\includegraphics[width=0.19\linewidth]{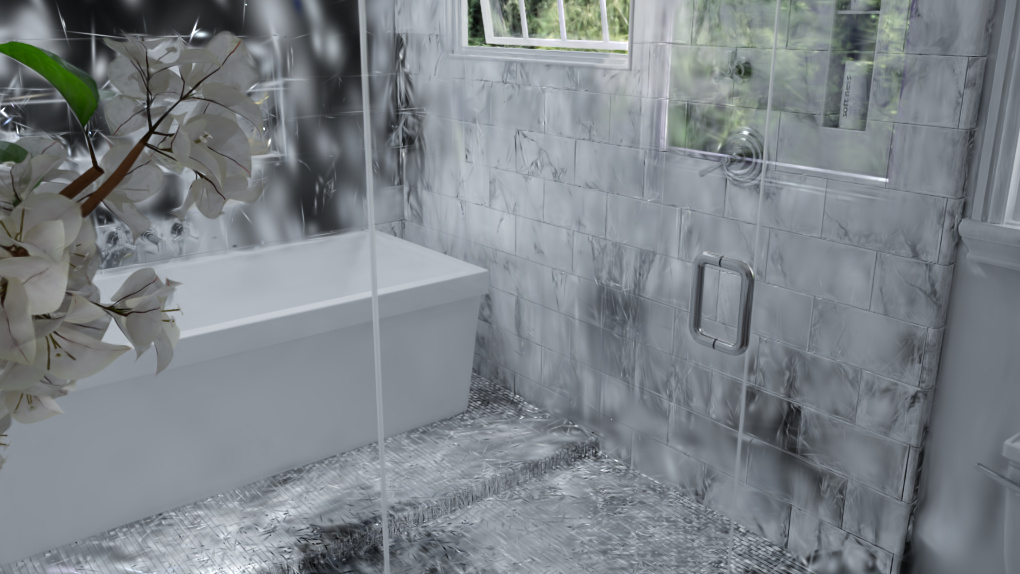}}
\\[28pt]

\vcenterbox{\rotatebox{90}{Bridge}} &
\vcenterbox{\includegraphics[width=0.19\linewidth]{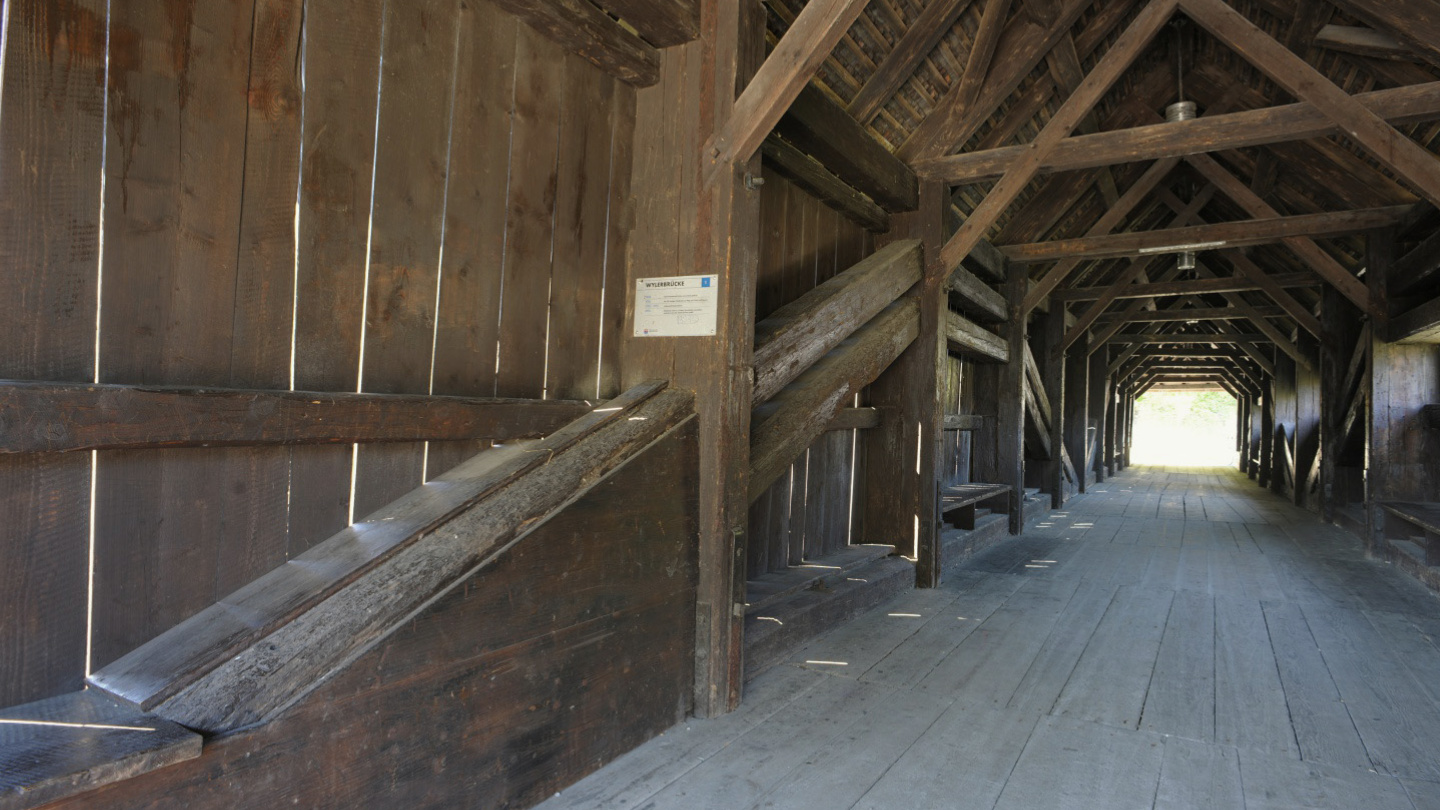}} &
\vcenterbox{\includegraphics[width=0.19\linewidth]{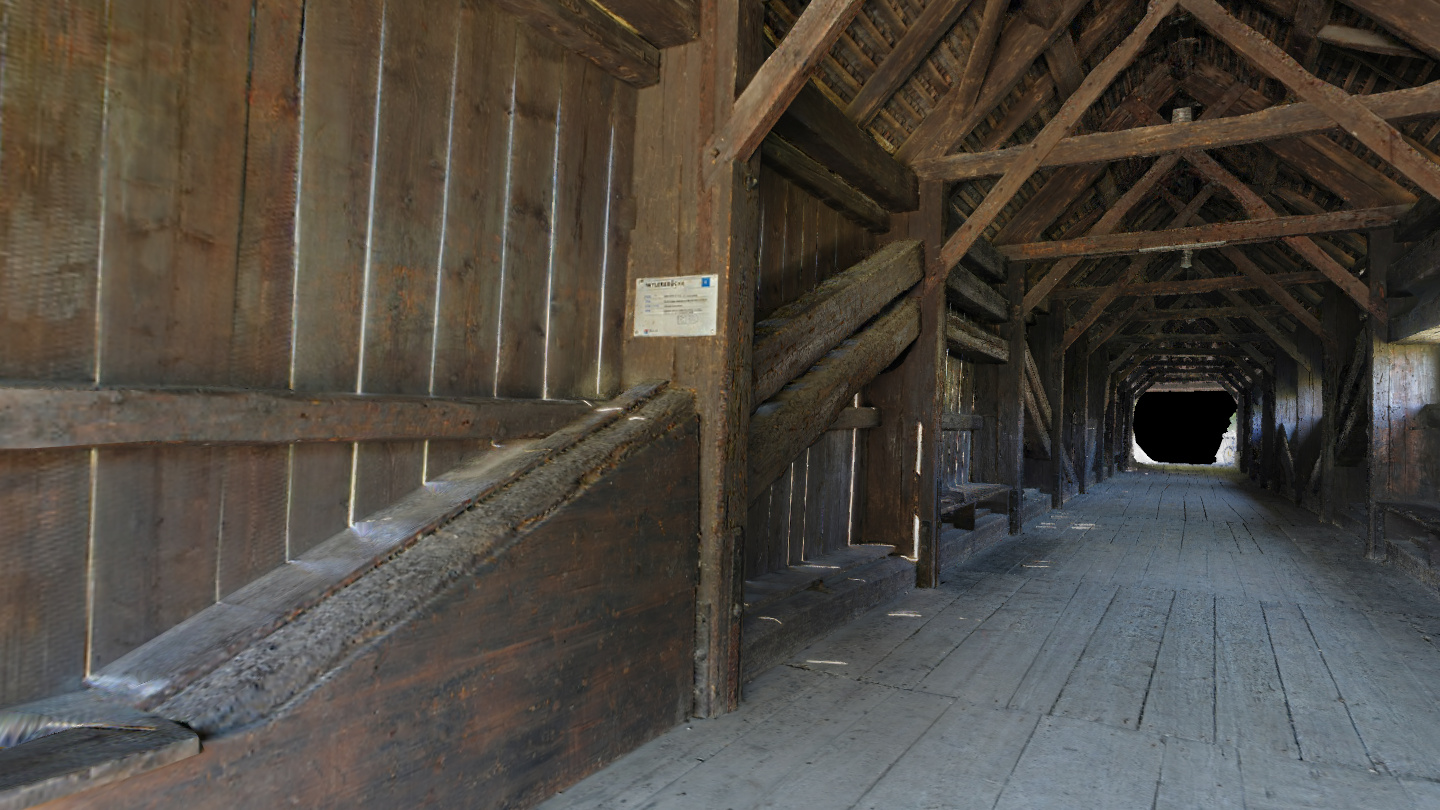}} &
\vcenterbox{\includegraphics[width=0.19\linewidth]{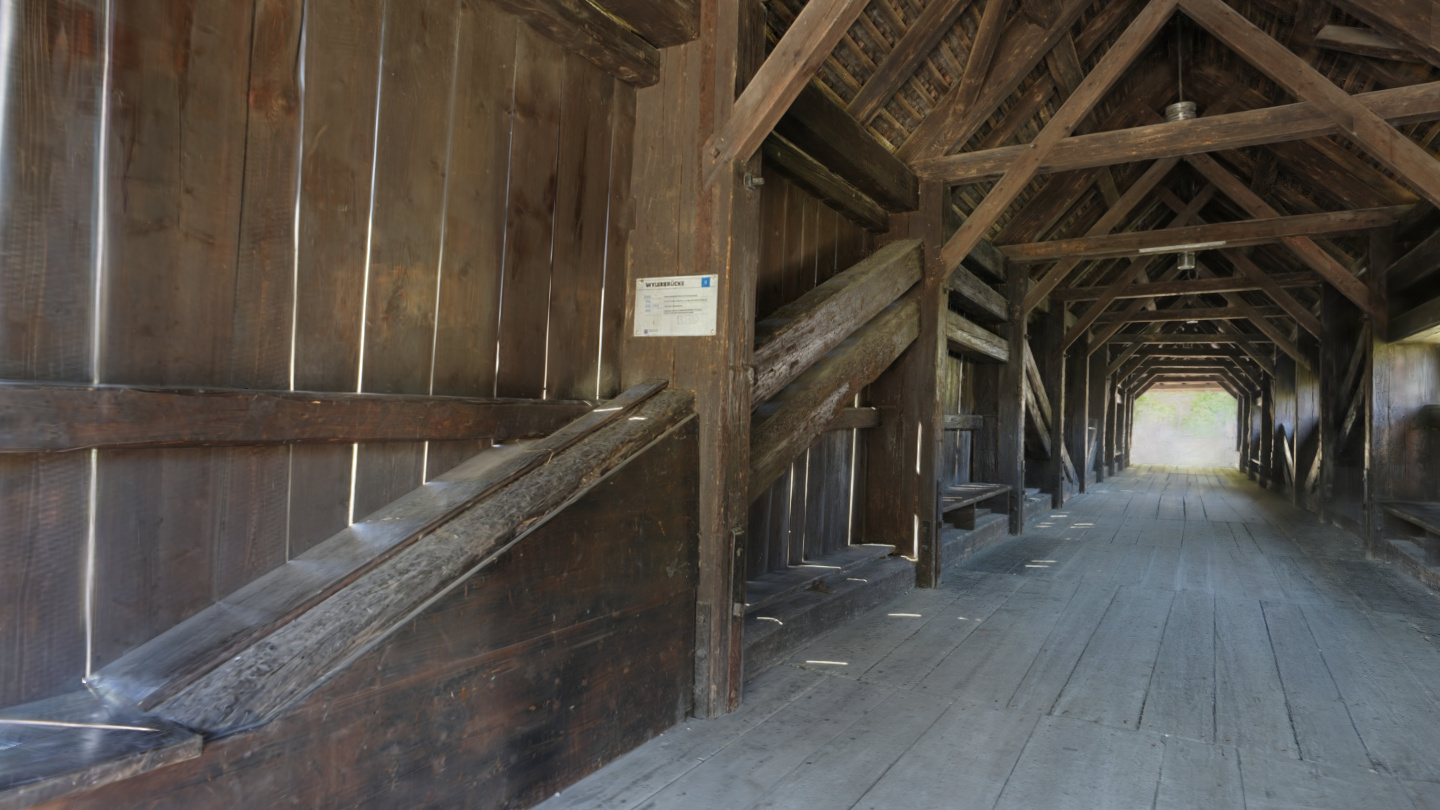}} &
\vcenterbox{\includegraphics[width=0.19\linewidth]{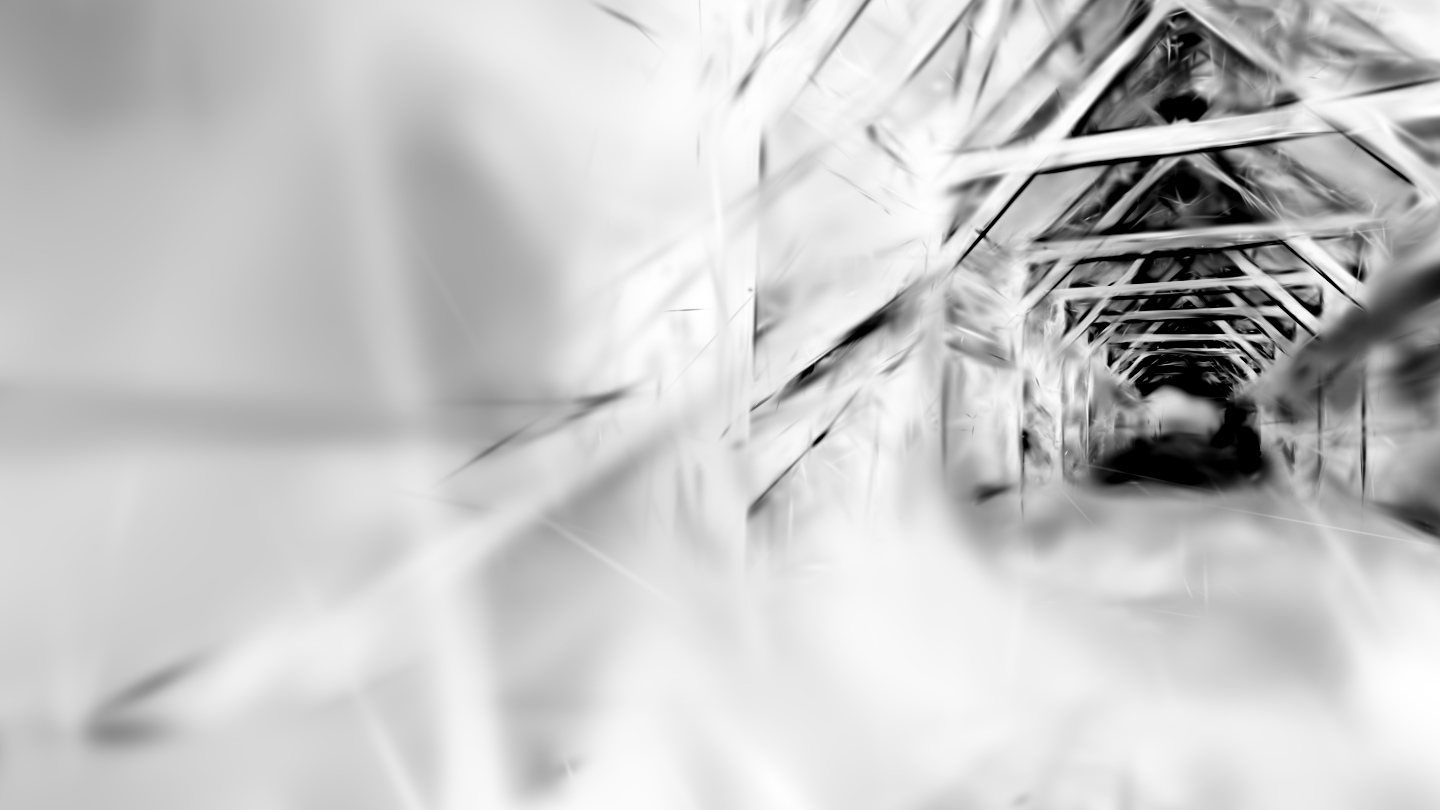}} &
\vcenterbox{\includegraphics[width=0.19\linewidth]{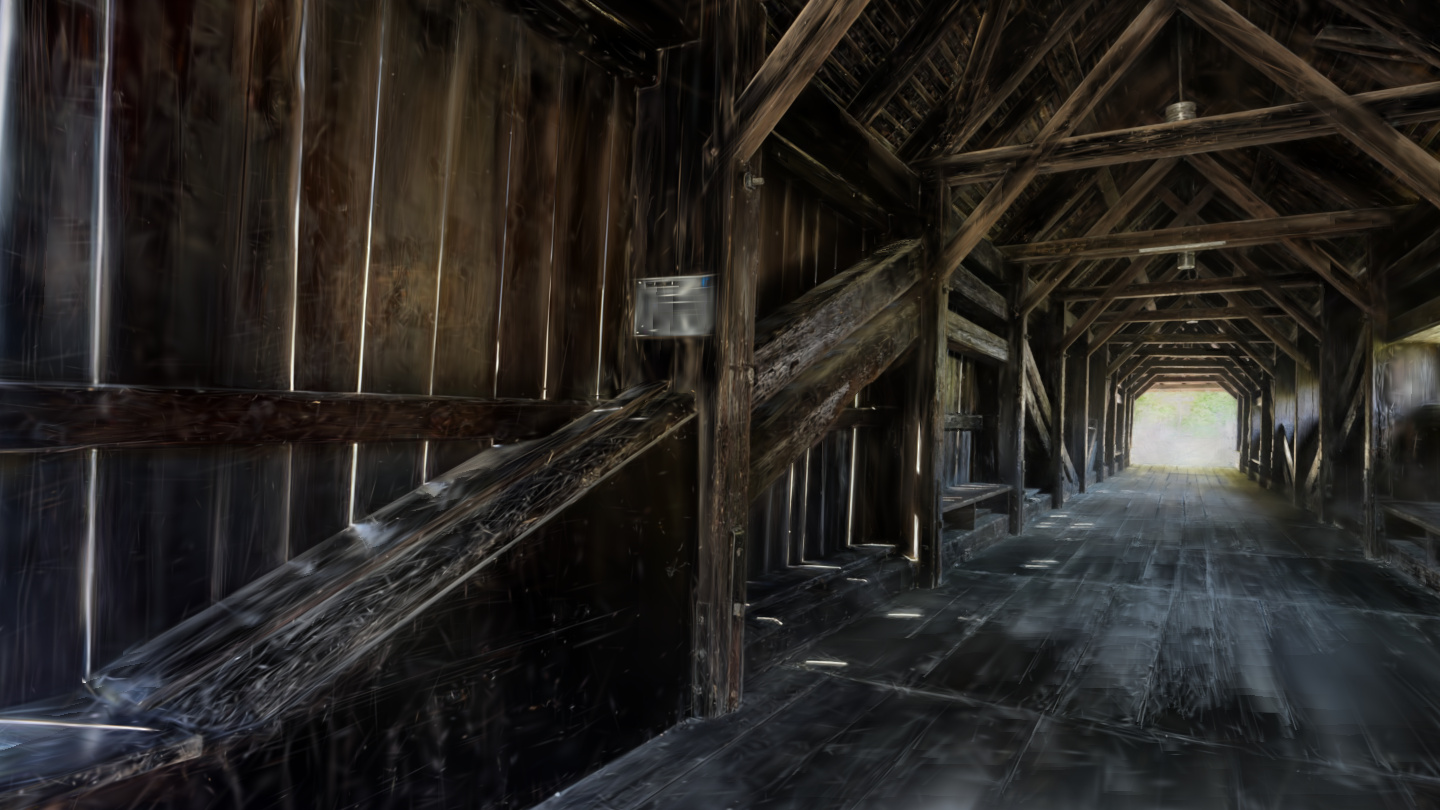}}
\\[28pt]

\vcenterbox{\rotatebox{90}{Museum-1}} &
\vcenterbox{\includegraphics[width=0.19\linewidth]{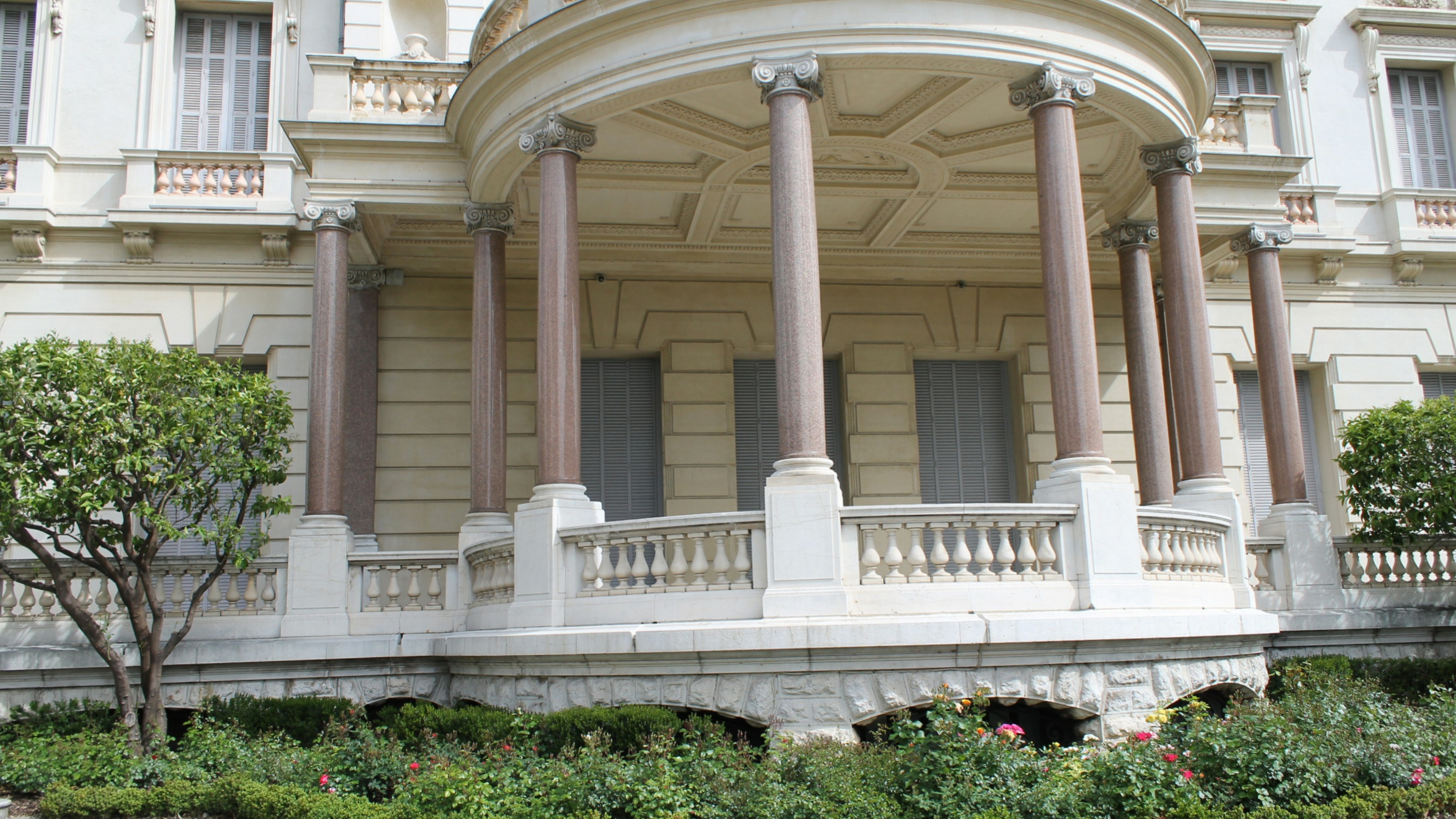}} &
\vcenterbox{\includegraphics[width=0.19\linewidth]{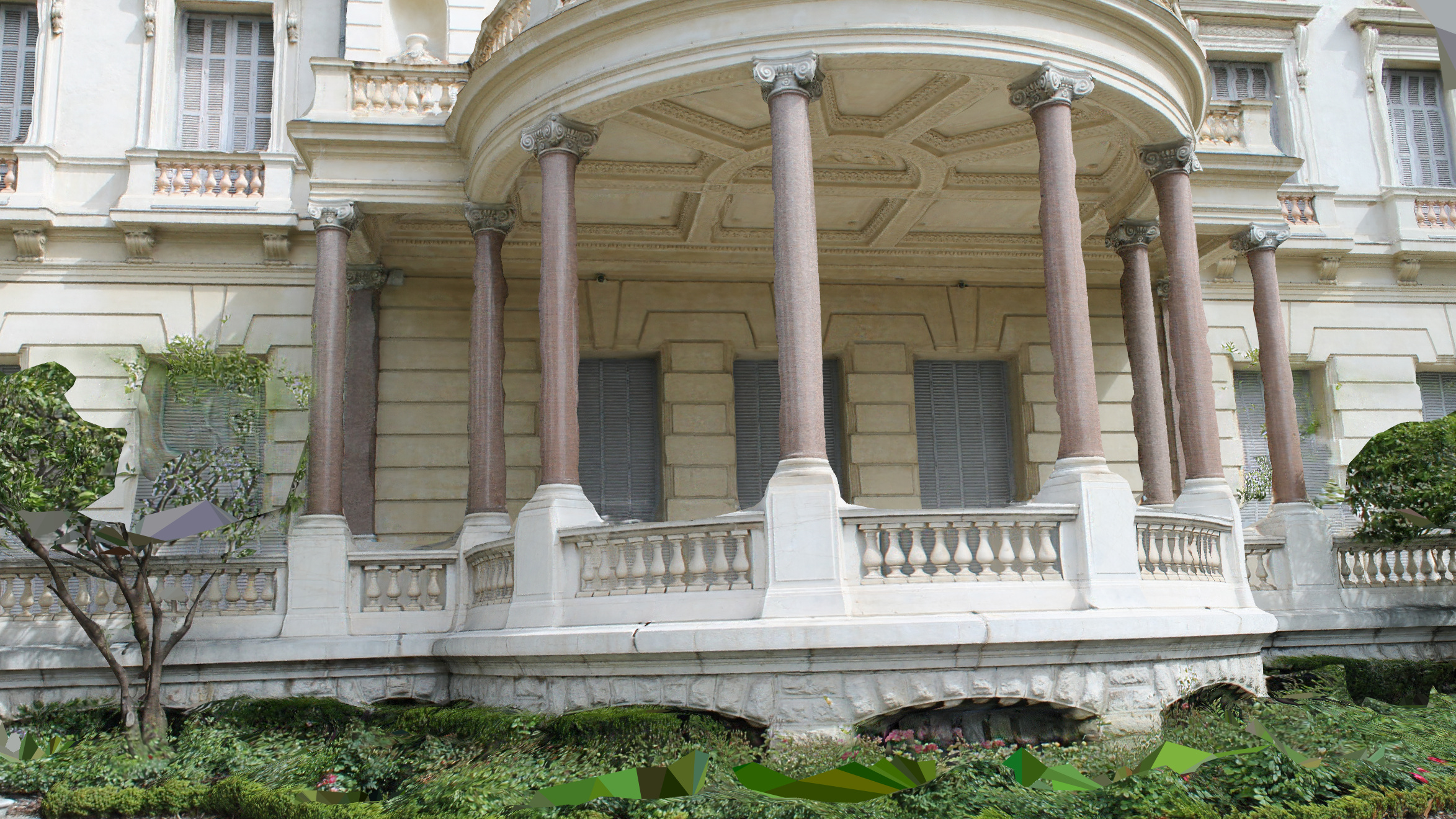}} &
\vcenterbox{\includegraphics[width=0.19\linewidth]{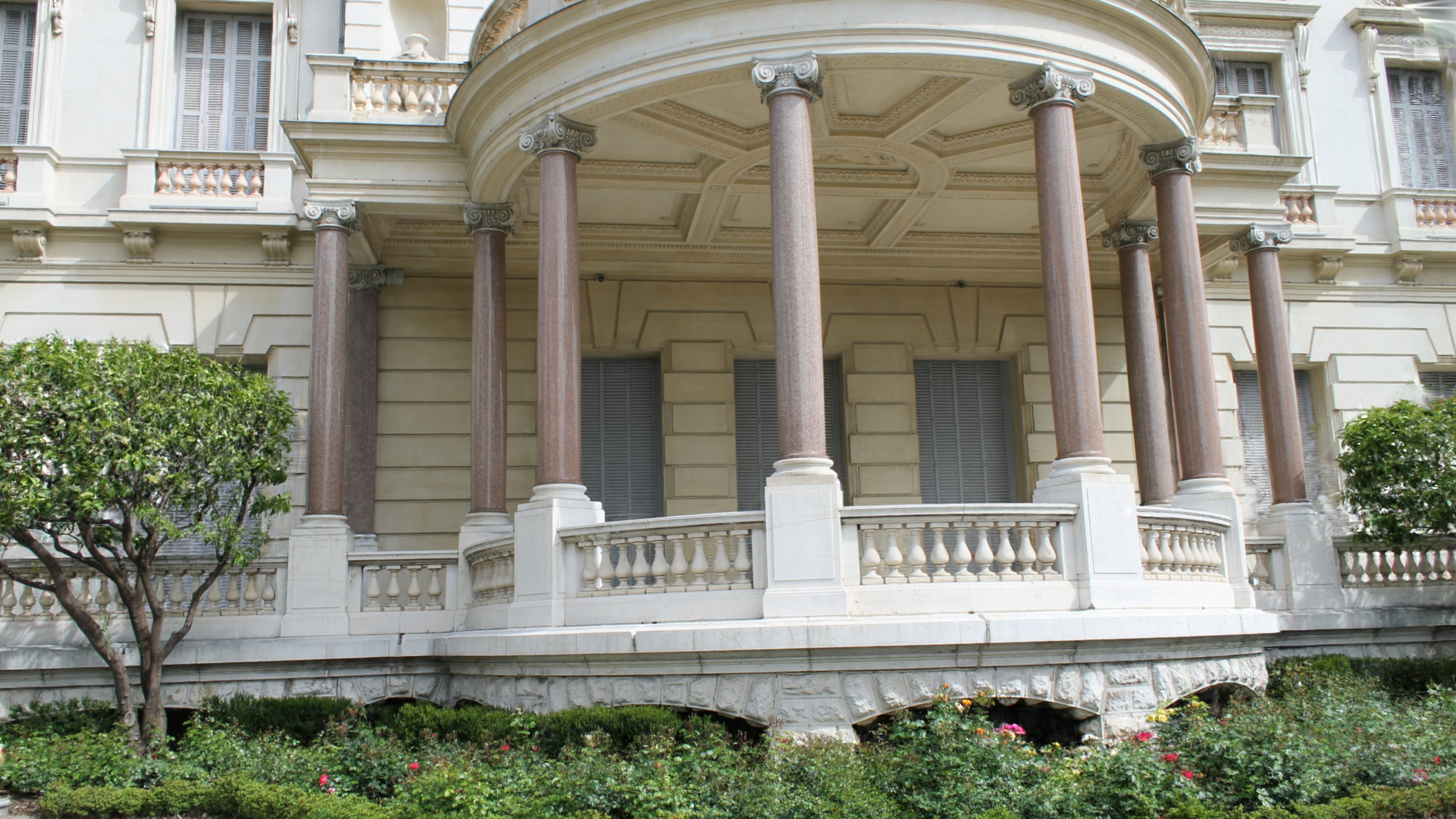}} &
\vcenterbox{\includegraphics[width=0.19\linewidth]{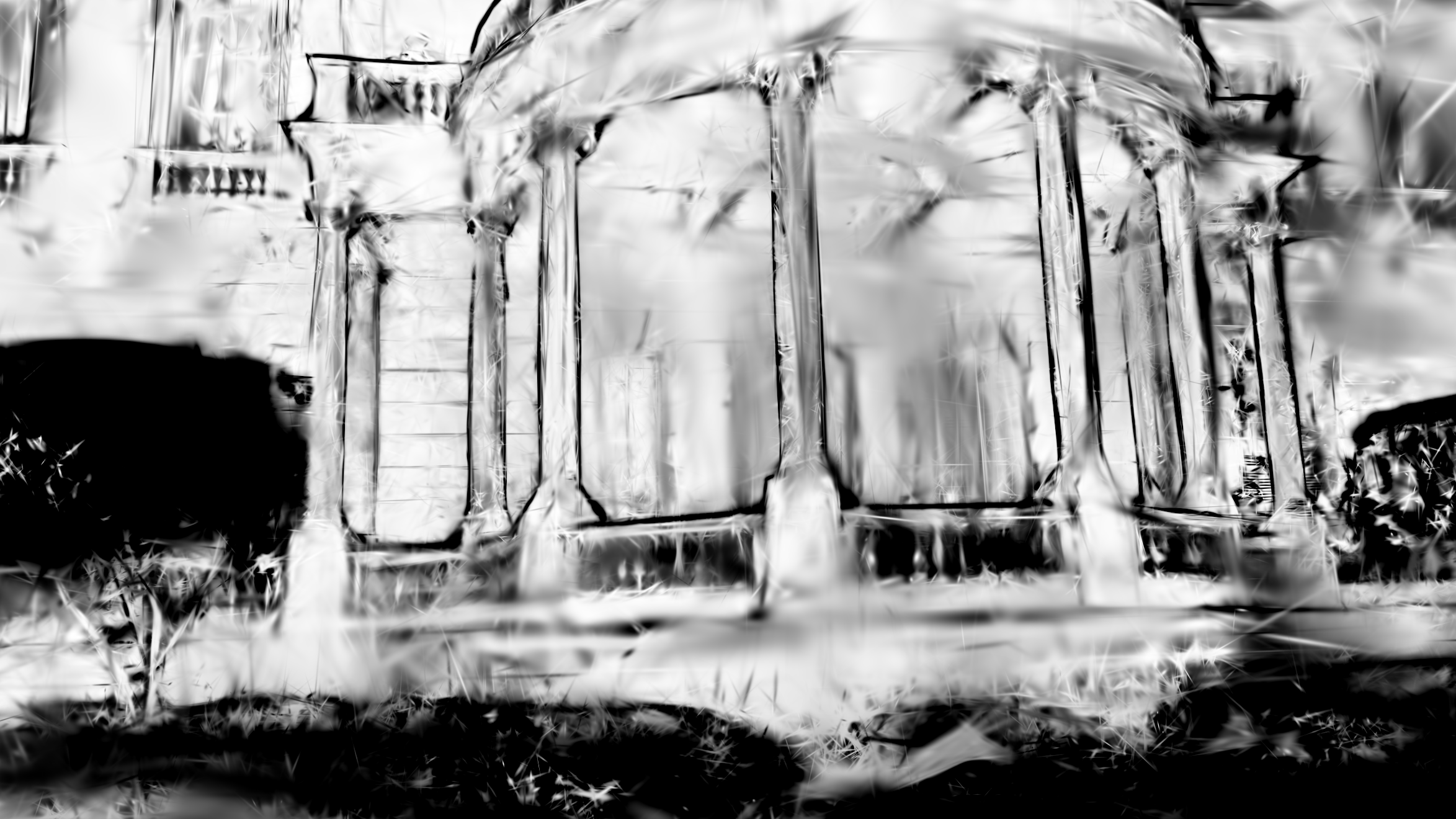}} &
\vcenterbox{\includegraphics[width=0.19\linewidth]{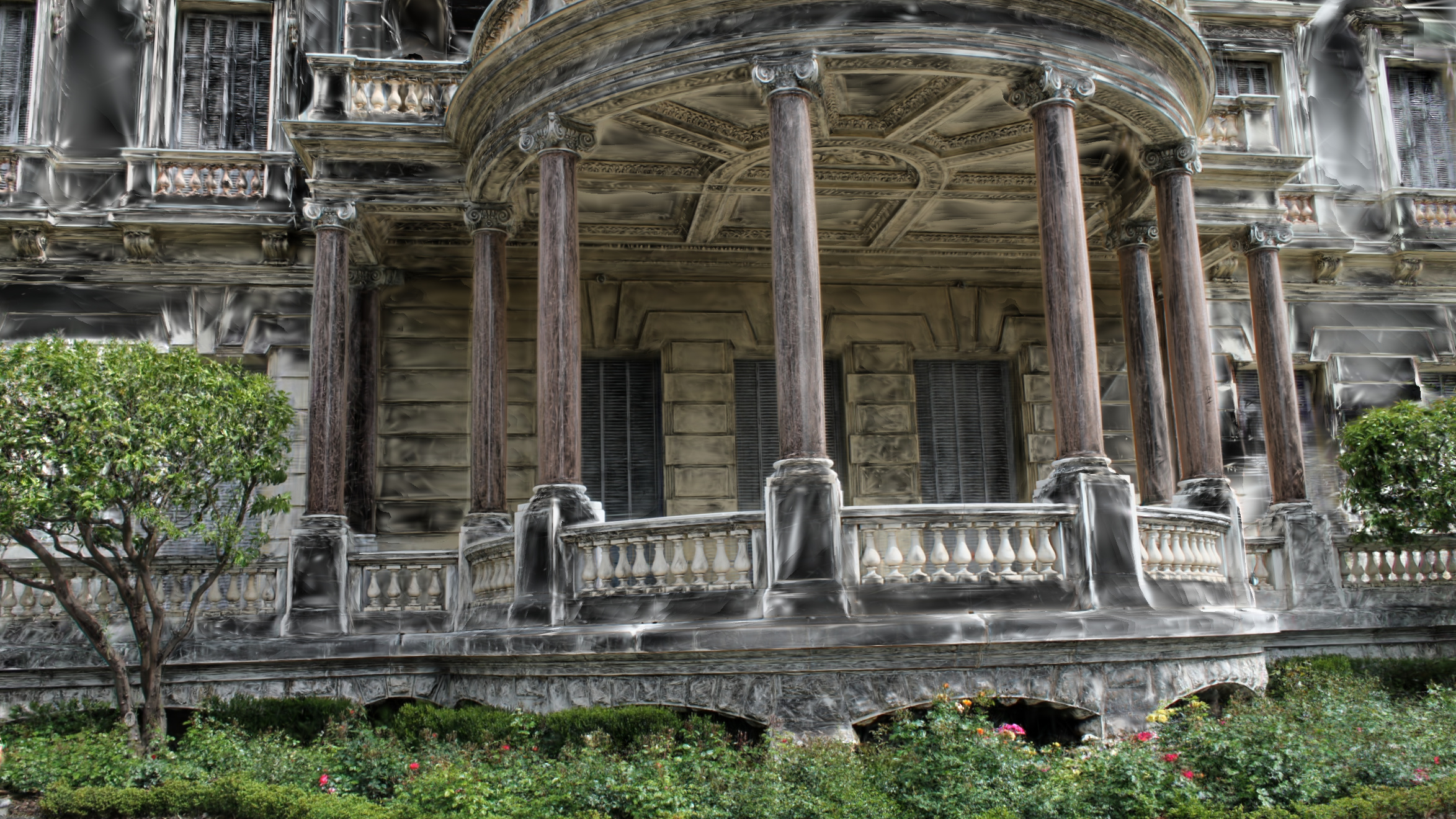}}
\\[28pt]

\vcenterbox{\rotatebox{90}{Museum-2}} &
\vcenterbox{\includegraphics[width=0.19\linewidth]{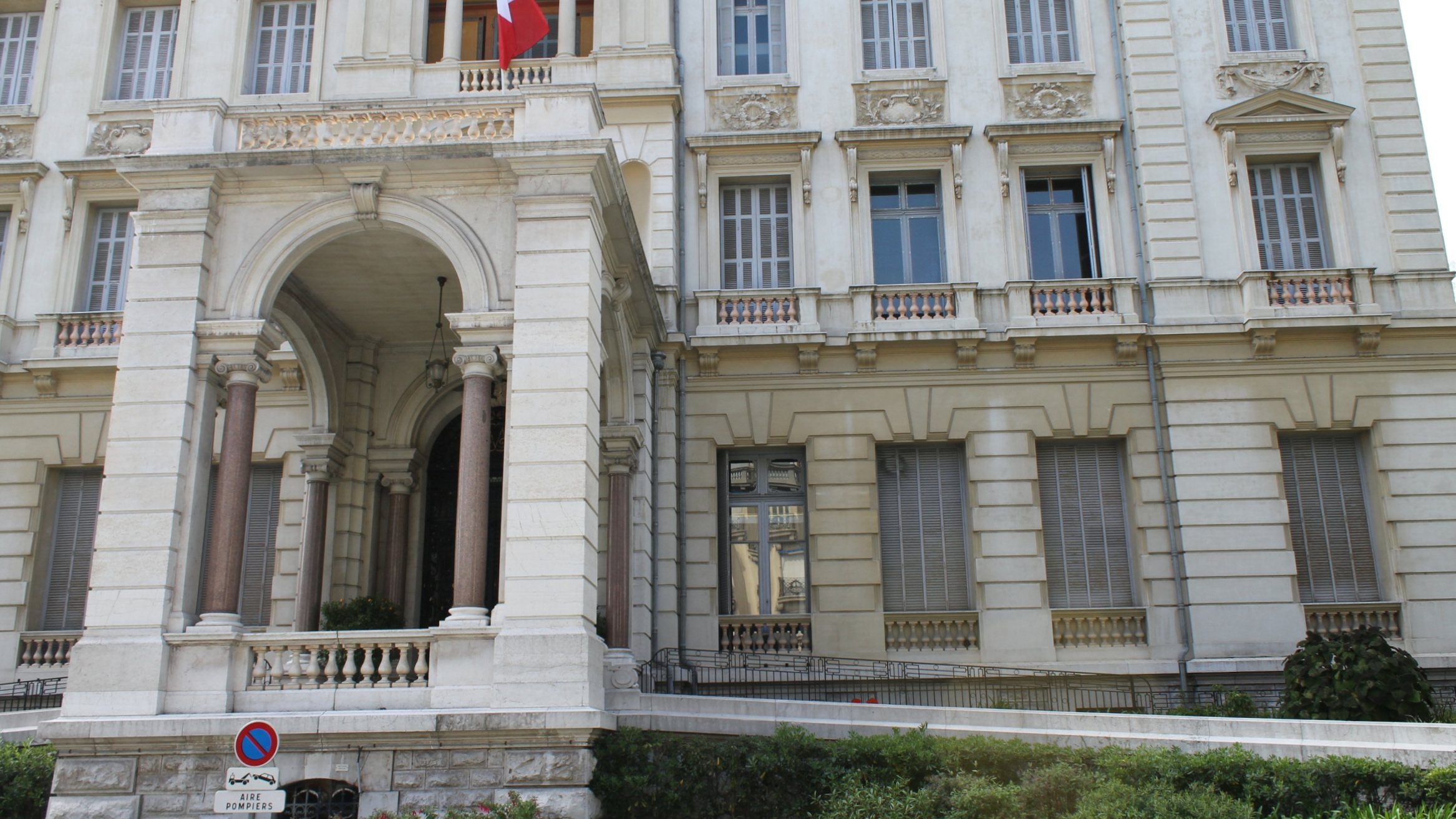}} &
\vcenterbox{\includegraphics[width=0.19\linewidth]{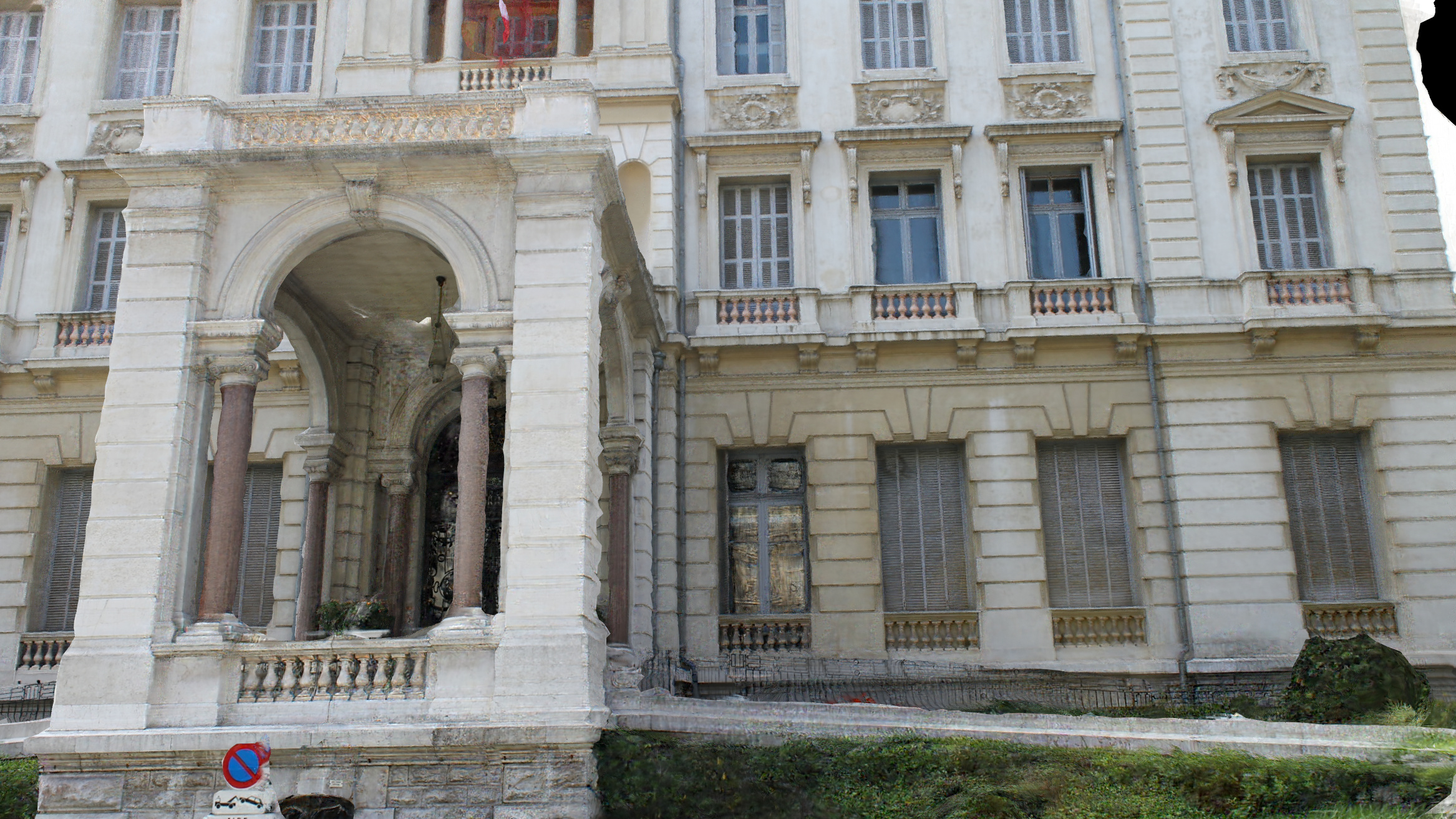}} &
\vcenterbox{\includegraphics[width=0.19\linewidth]{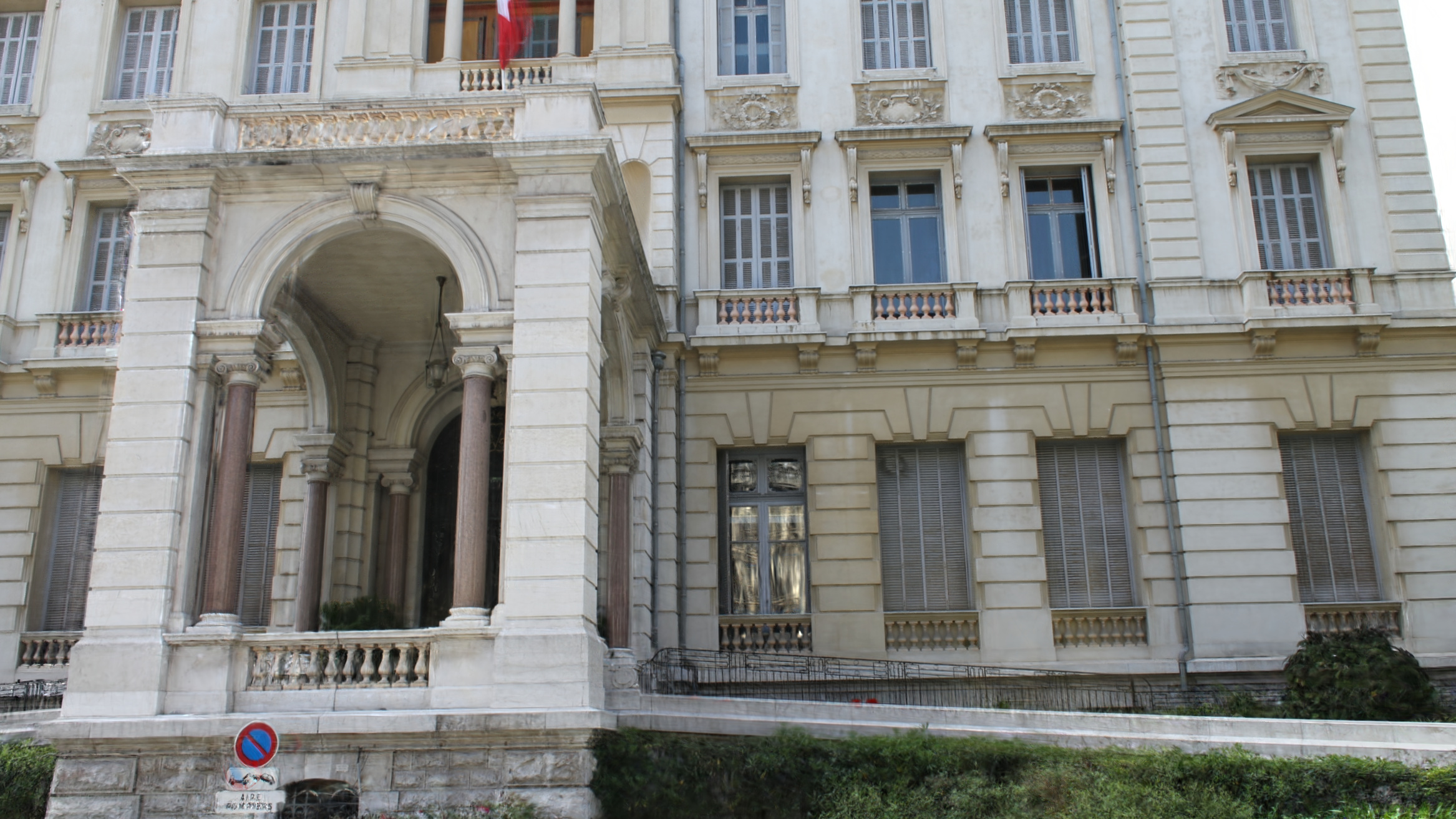}} &
\vcenterbox{\includegraphics[width=0.19\linewidth]{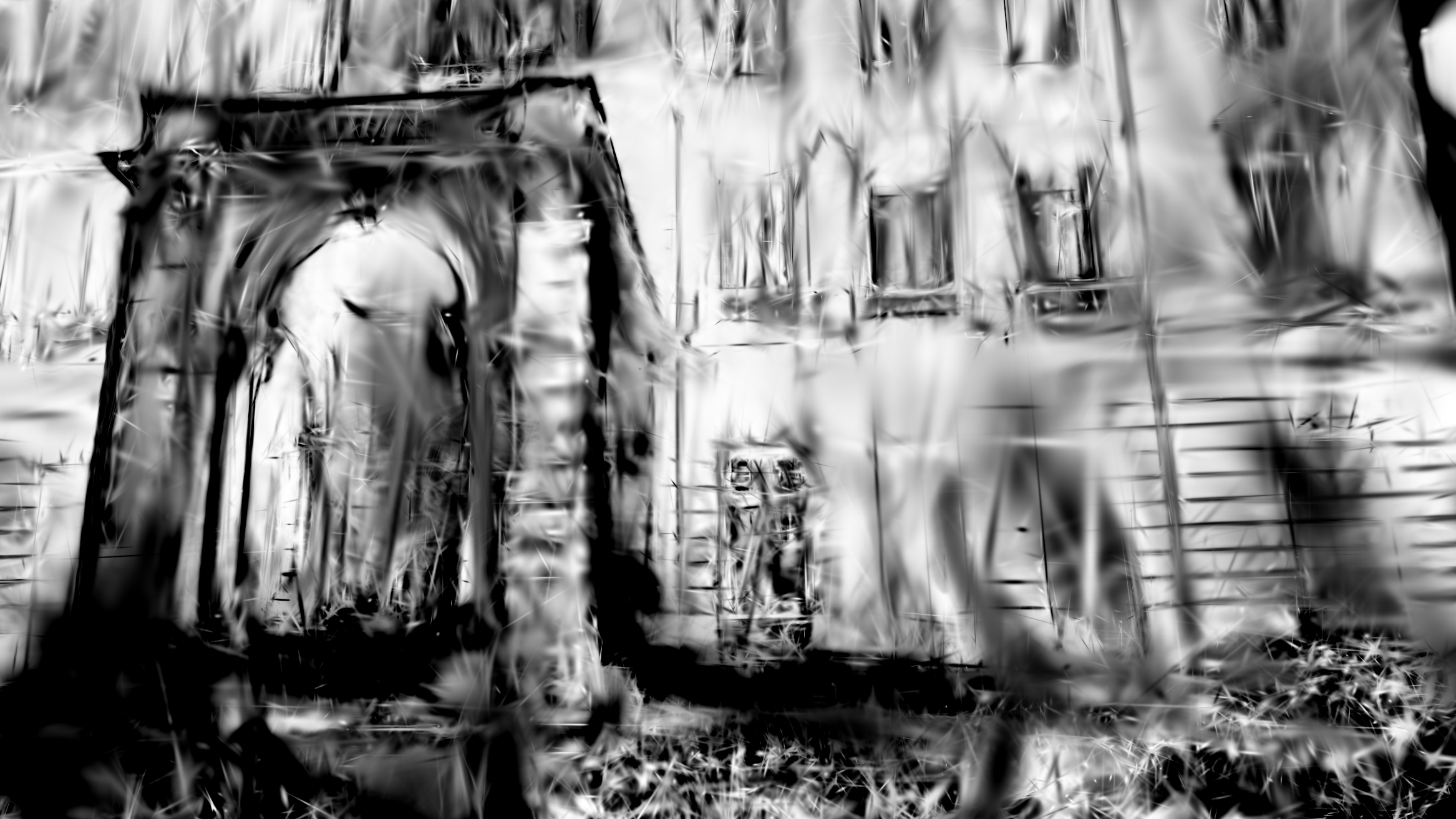}} &
\vcenterbox{\includegraphics[width=0.19\linewidth]{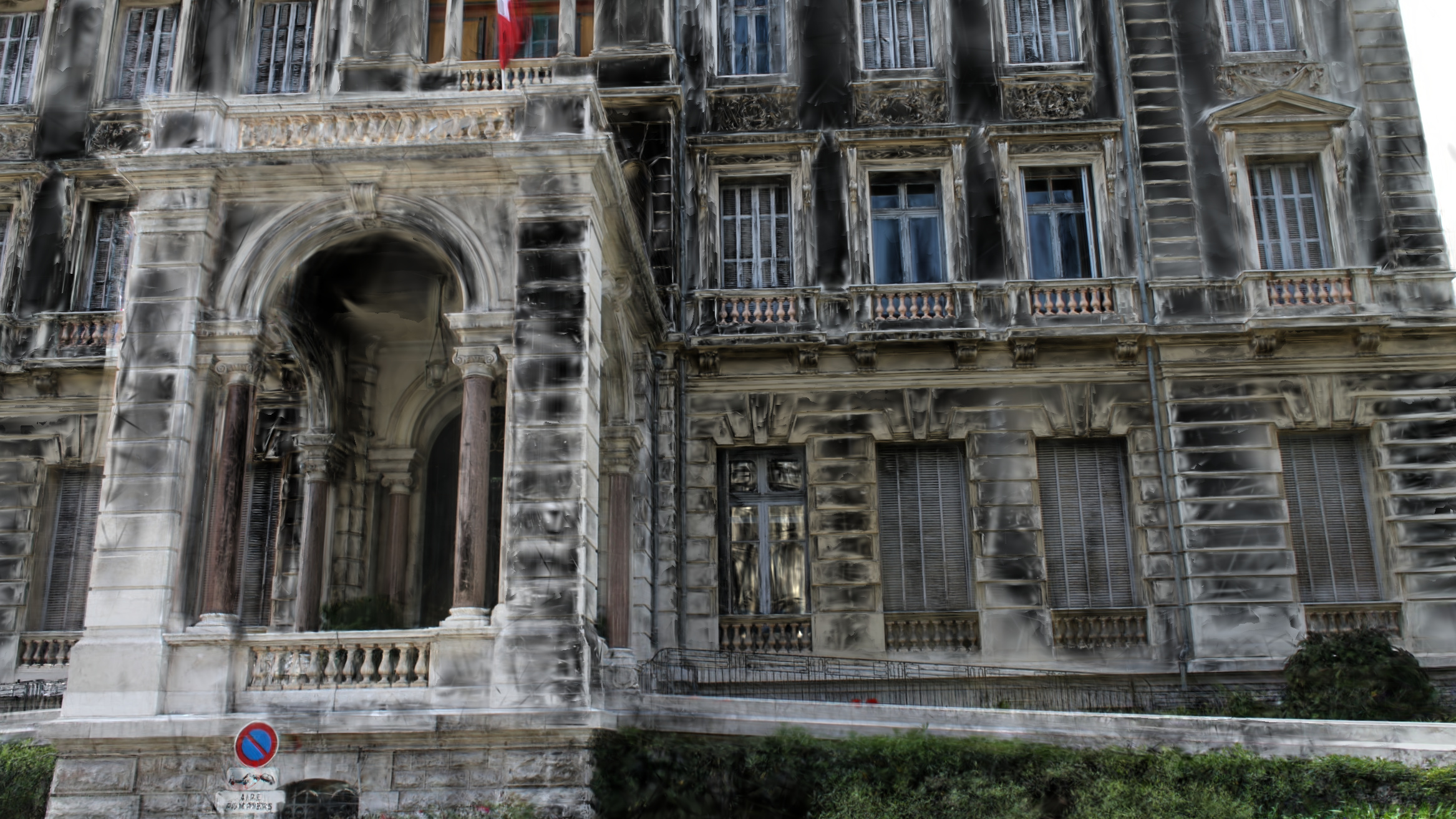}}
\\[28pt]

\vcenterbox{\rotatebox{90}{Playroom}} &
\vcenterbox{\includegraphics[width=0.19\linewidth]{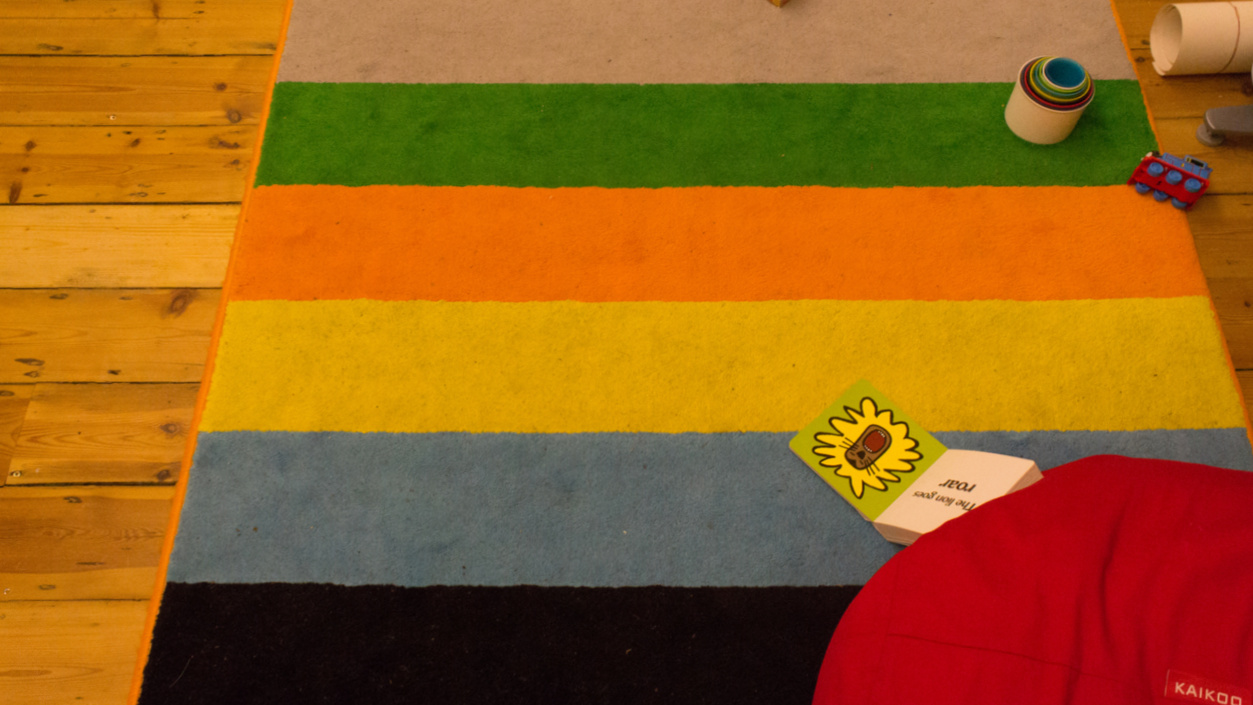}} &
\vcenterbox{\includegraphics[width=0.19\linewidth]{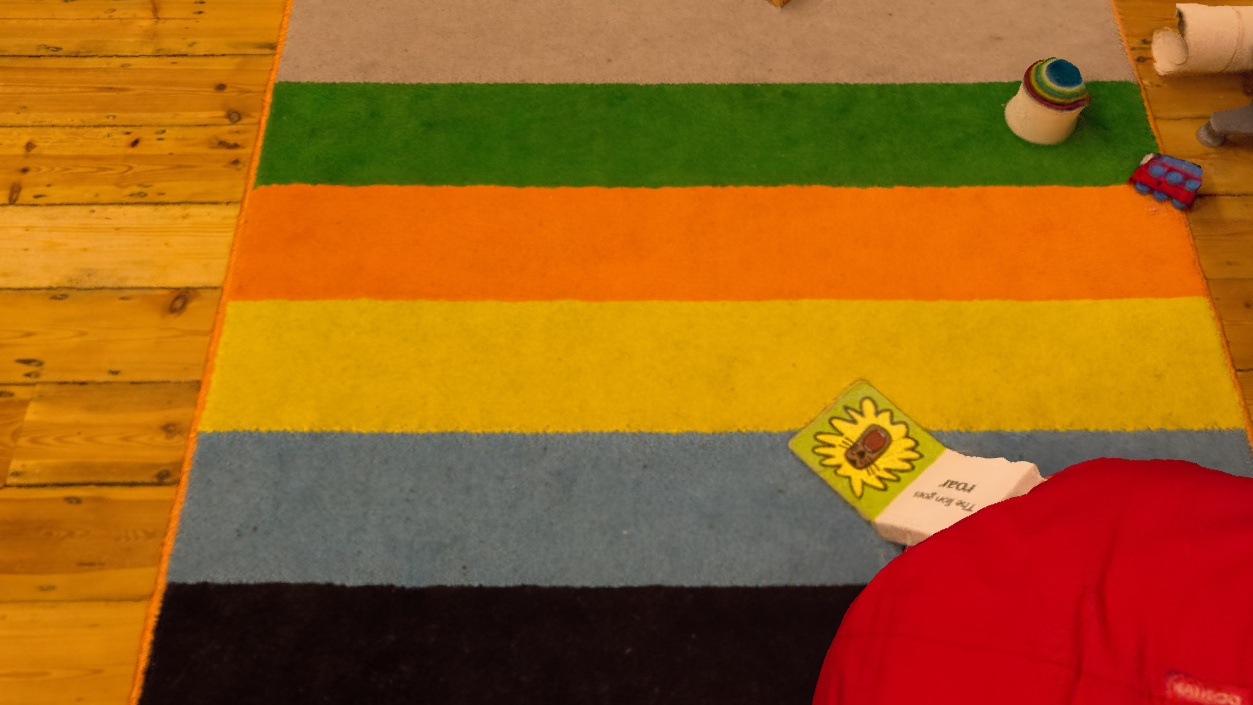}} &
\vcenterbox{\includegraphics[width=0.19\linewidth]{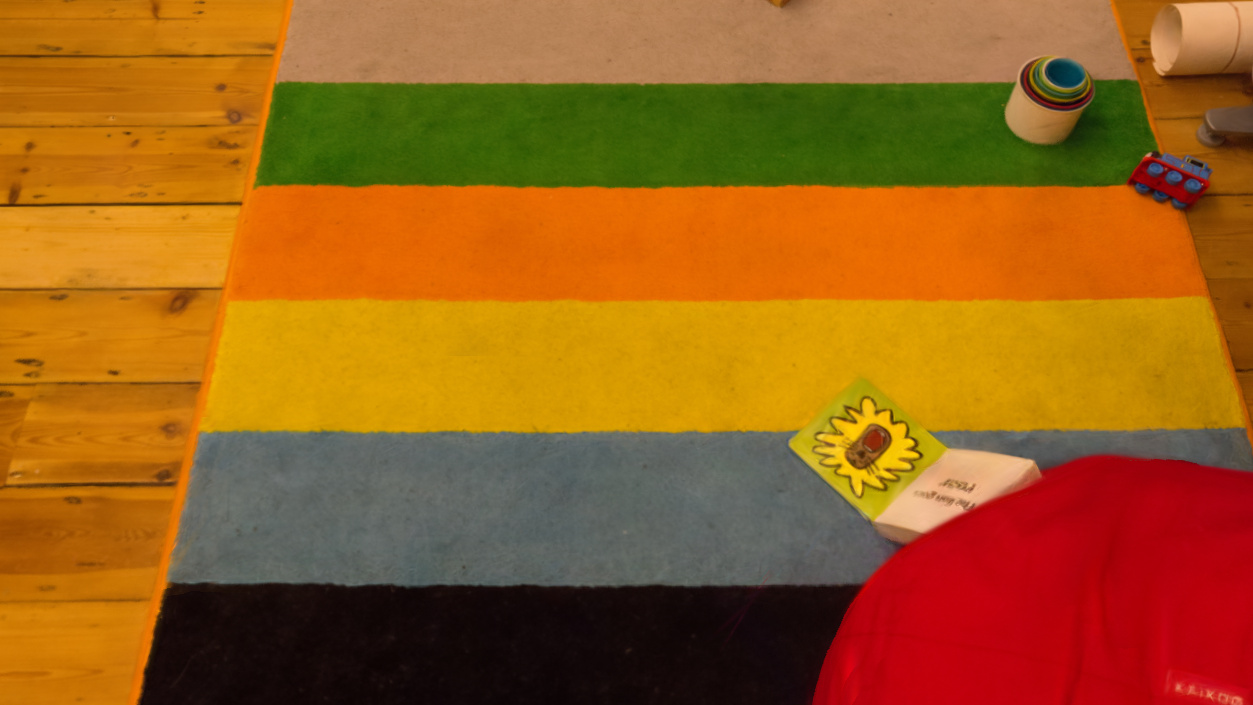}} &
\vcenterbox{\includegraphics[width=0.19\linewidth]{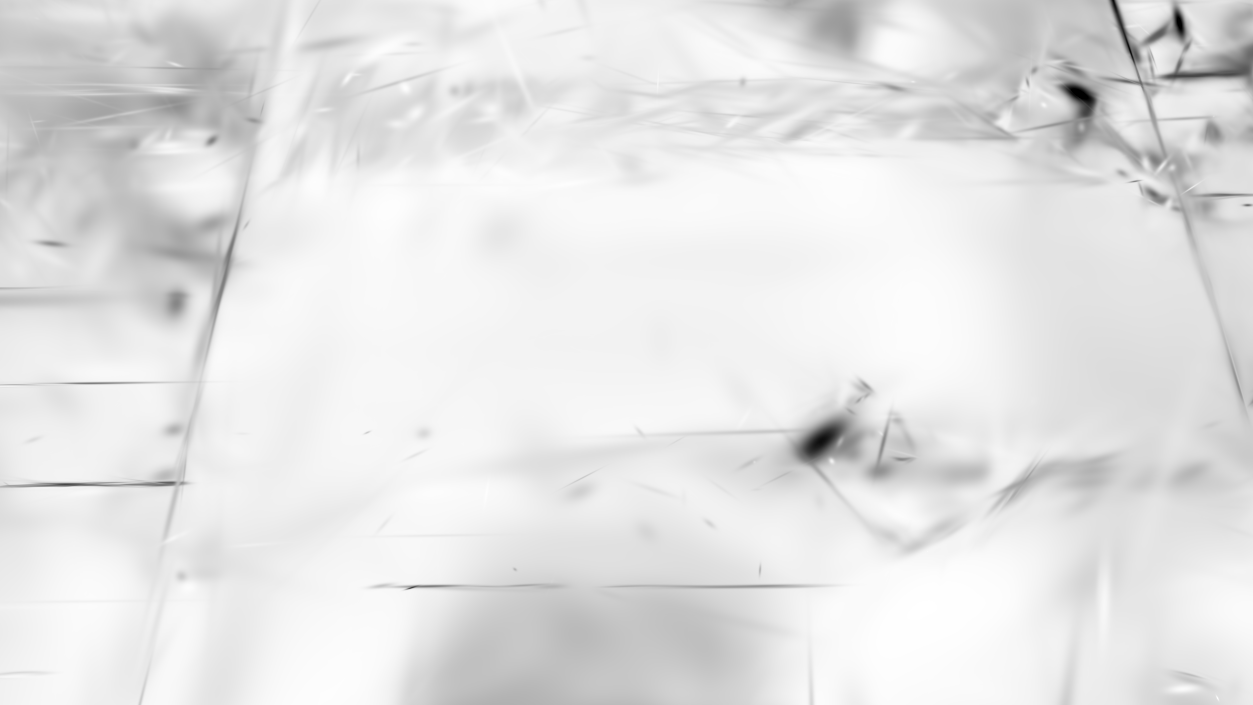}} &
\vcenterbox{\includegraphics[width=0.19\linewidth]{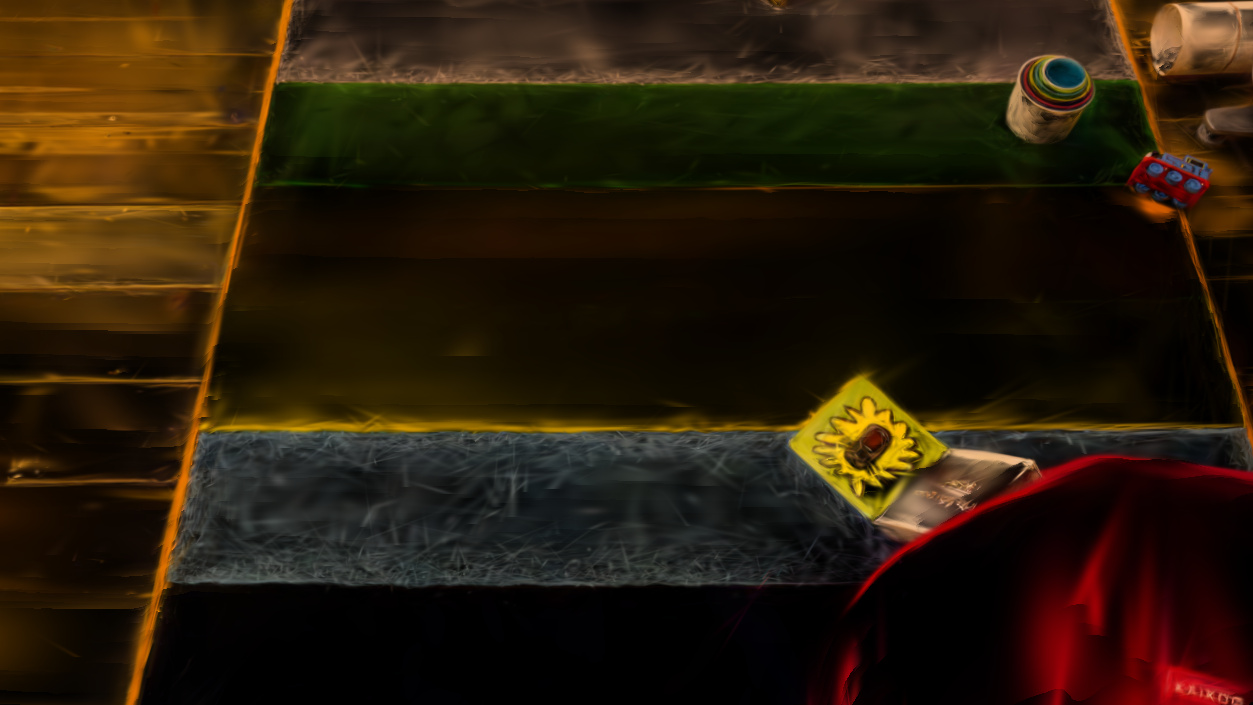}}
\\[28pt]

\vcenterbox{\rotatebox{90}{SaintAnne}} &
\vcenterbox{\includegraphics[width=0.19\linewidth]{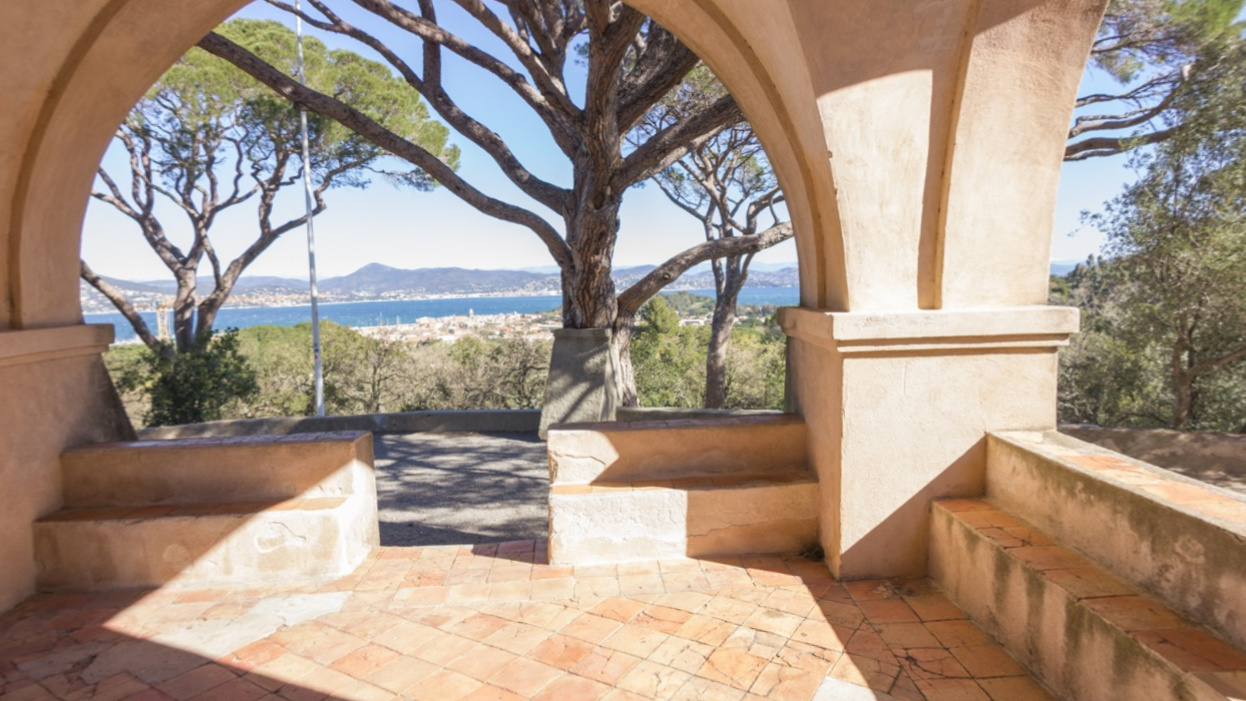}} &
\vcenterbox{\includegraphics[width=0.19\linewidth]{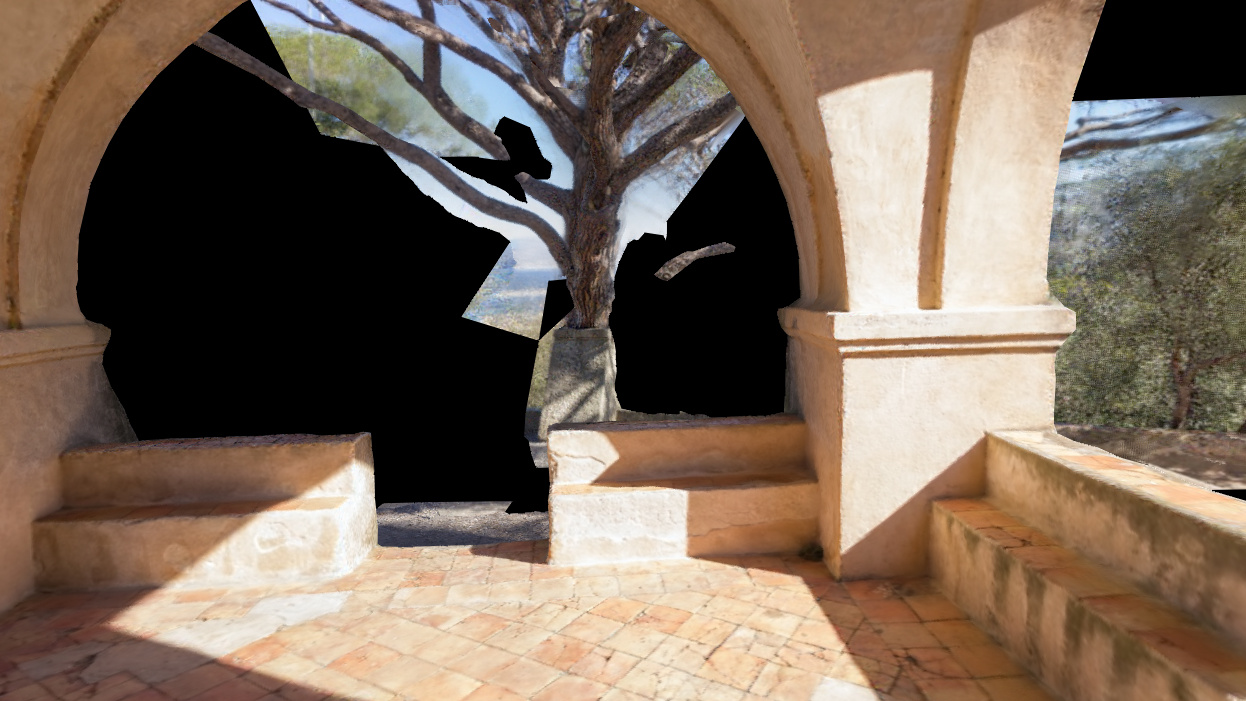}} &
\vcenterbox{\includegraphics[width=0.19\linewidth]{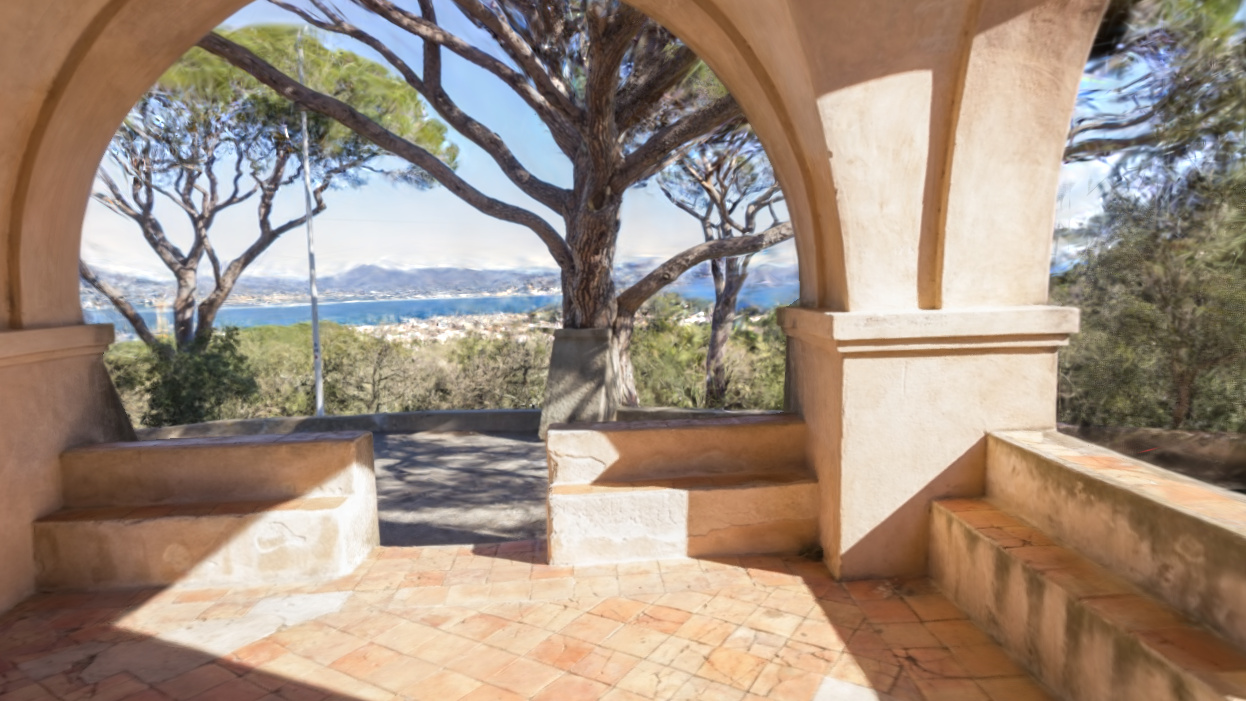}} &
\vcenterbox{\includegraphics[width=0.19\linewidth]{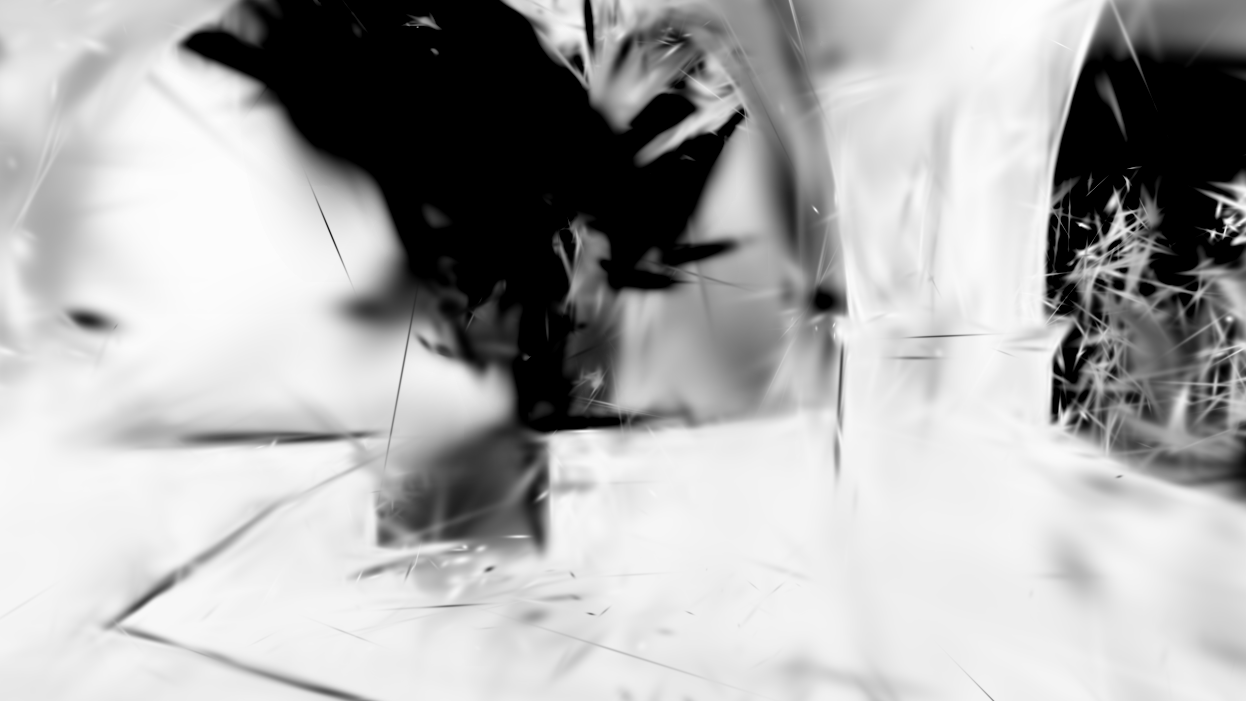}} &
\vcenterbox{\includegraphics[width=0.19\linewidth]{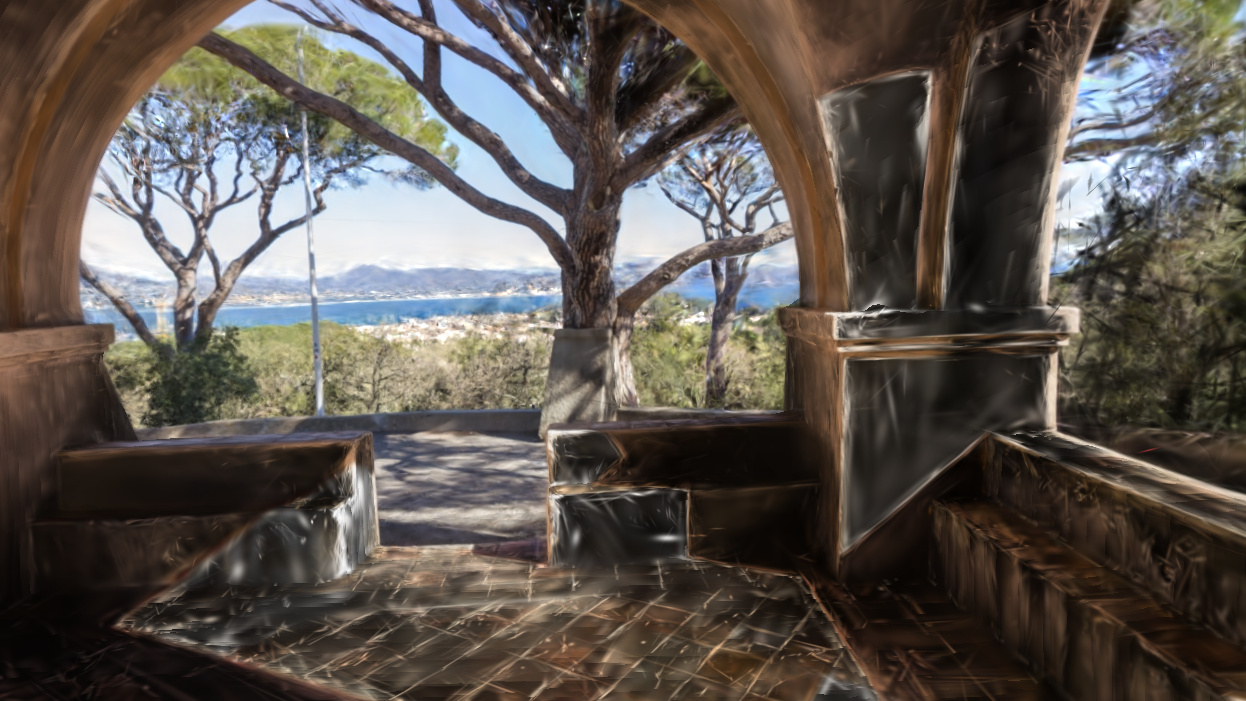}}
\\[28pt]

\vcenterbox{\rotatebox{90}{DrJohnson}} &
\vcenterbox{\includegraphics[width=0.19\linewidth]{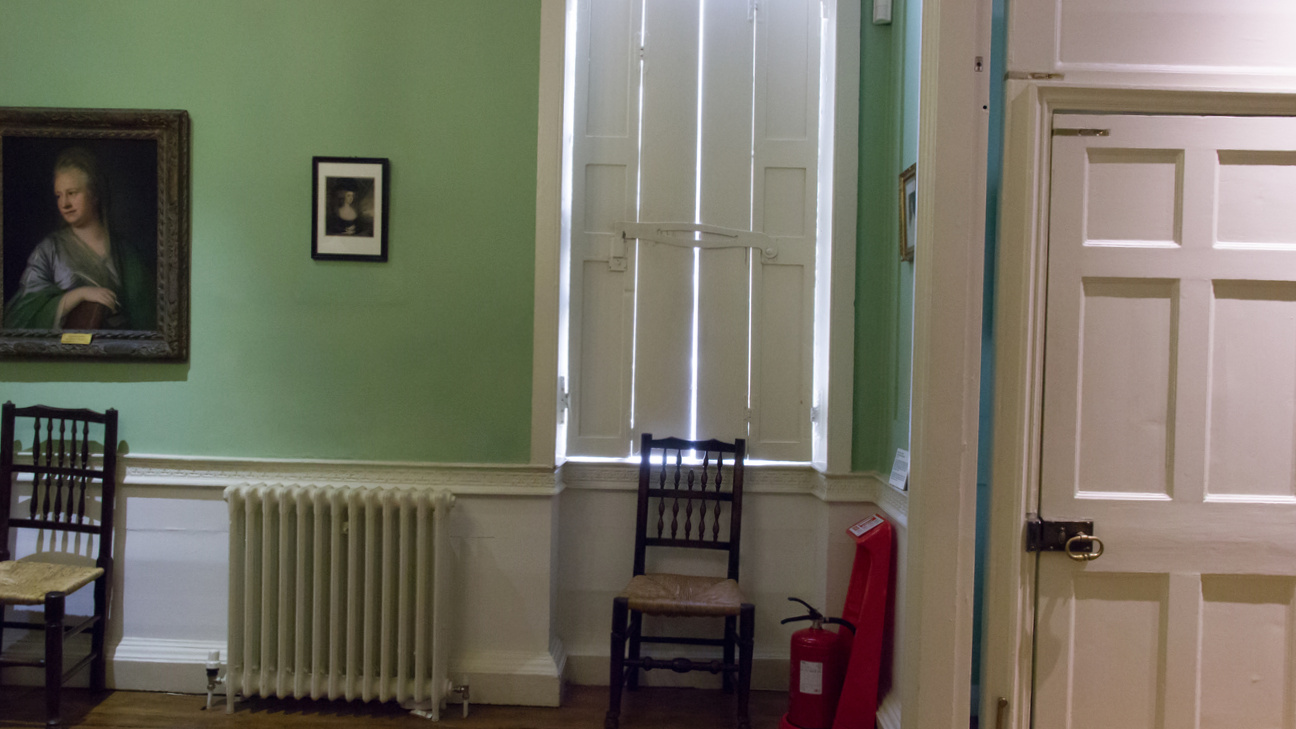}} &
\vcenterbox{\includegraphics[width=0.19\linewidth]{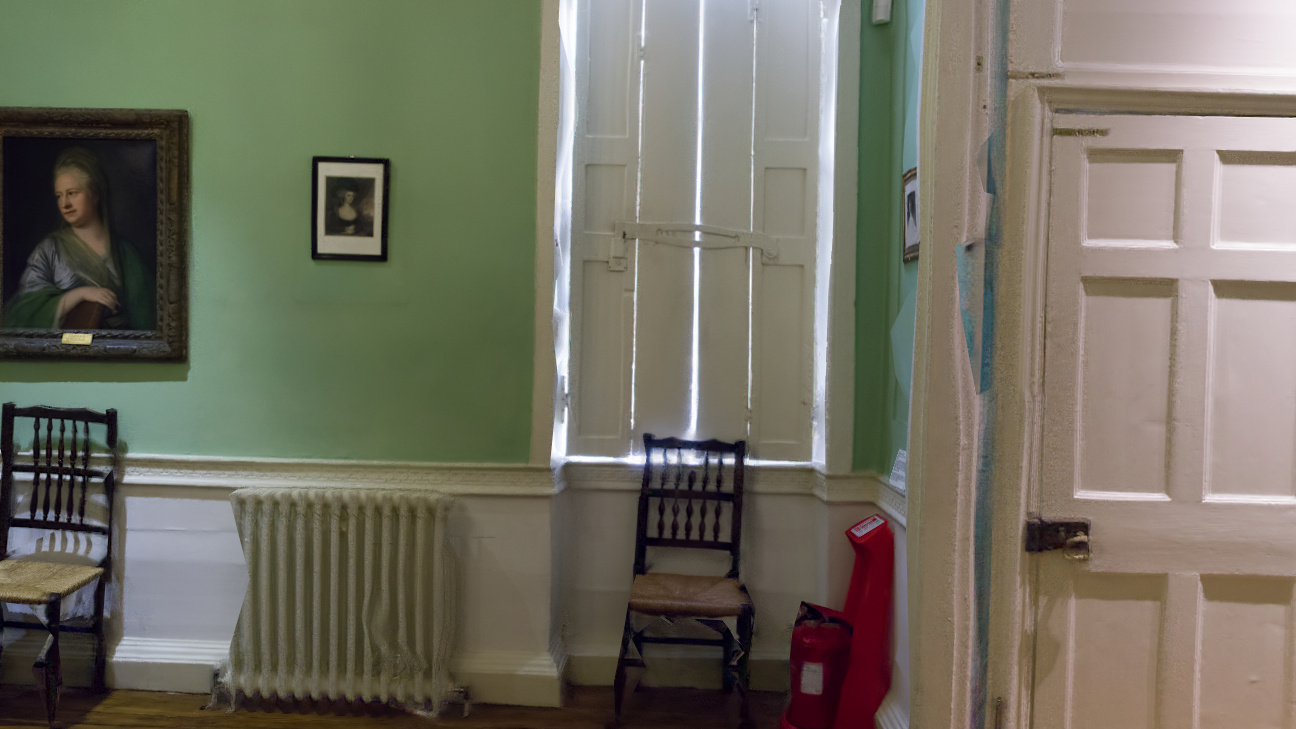}} &
\vcenterbox{\includegraphics[width=0.19\linewidth]{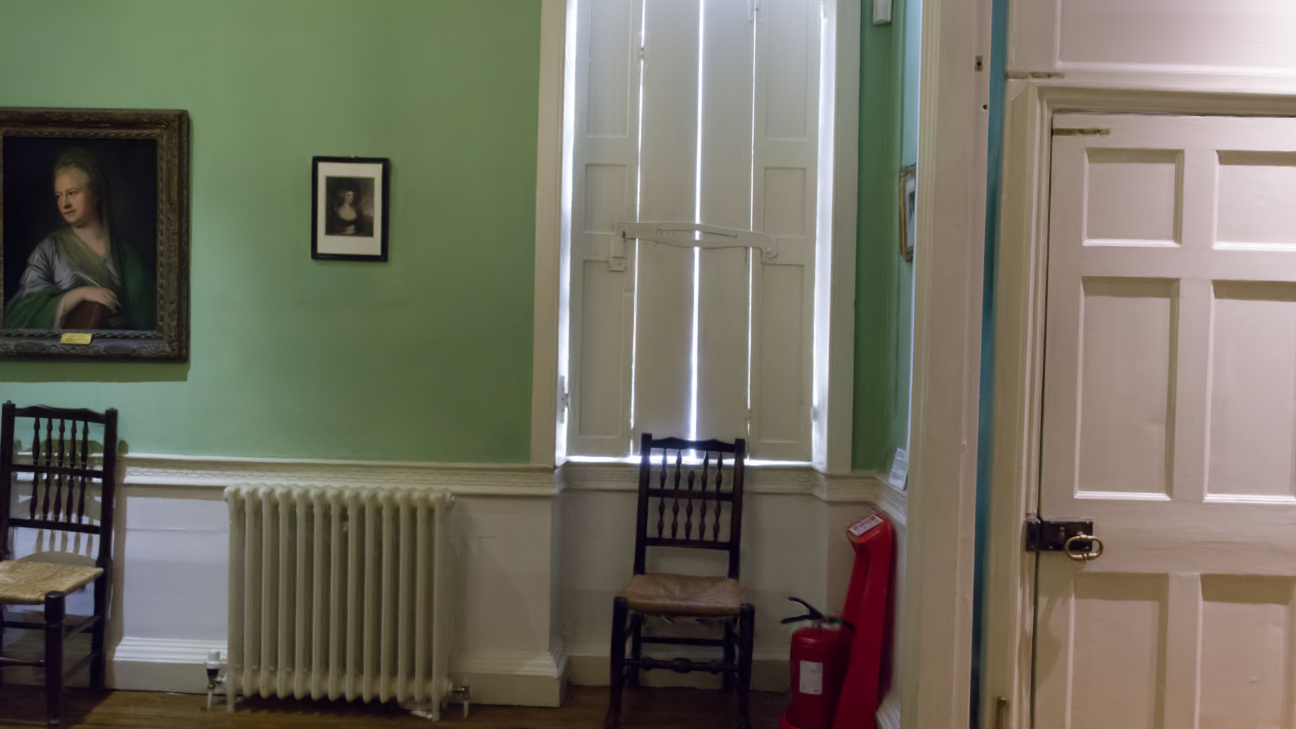}} &
\vcenterbox{\includegraphics[width=0.19\linewidth]{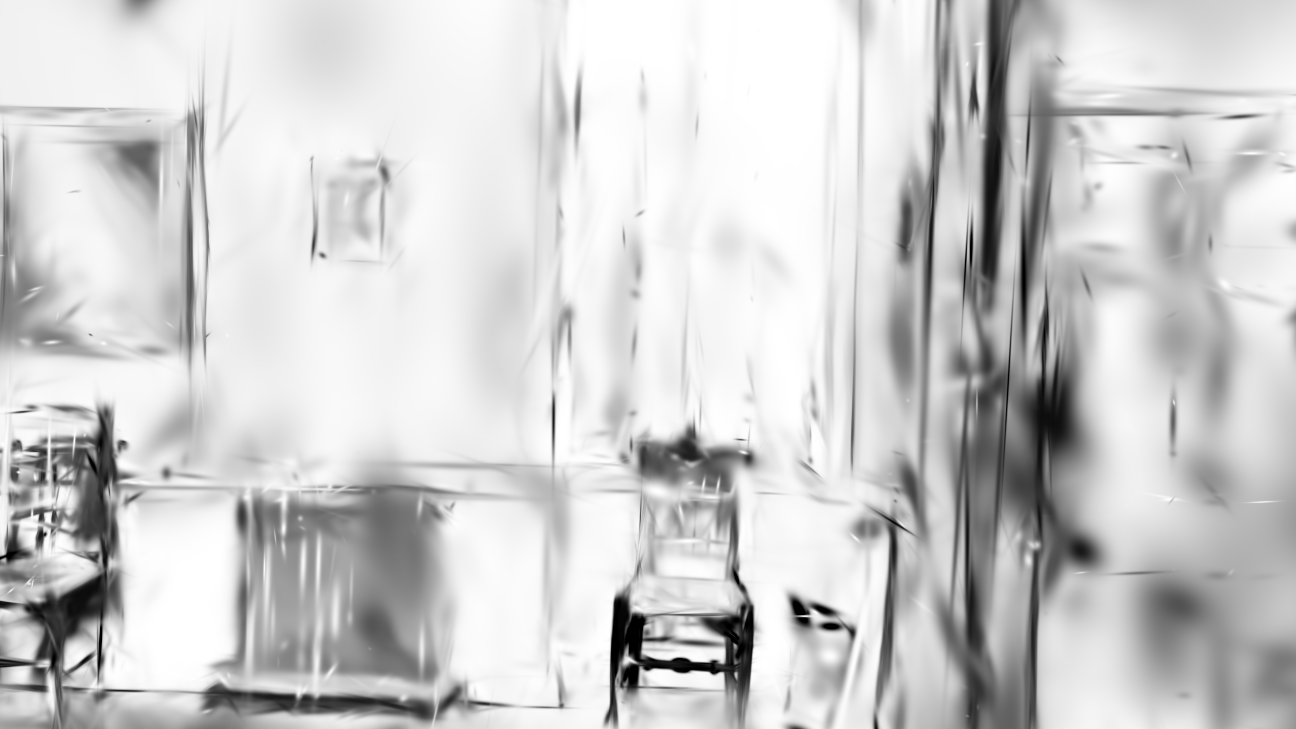}} &
\vcenterbox{\includegraphics[width=0.19\linewidth]{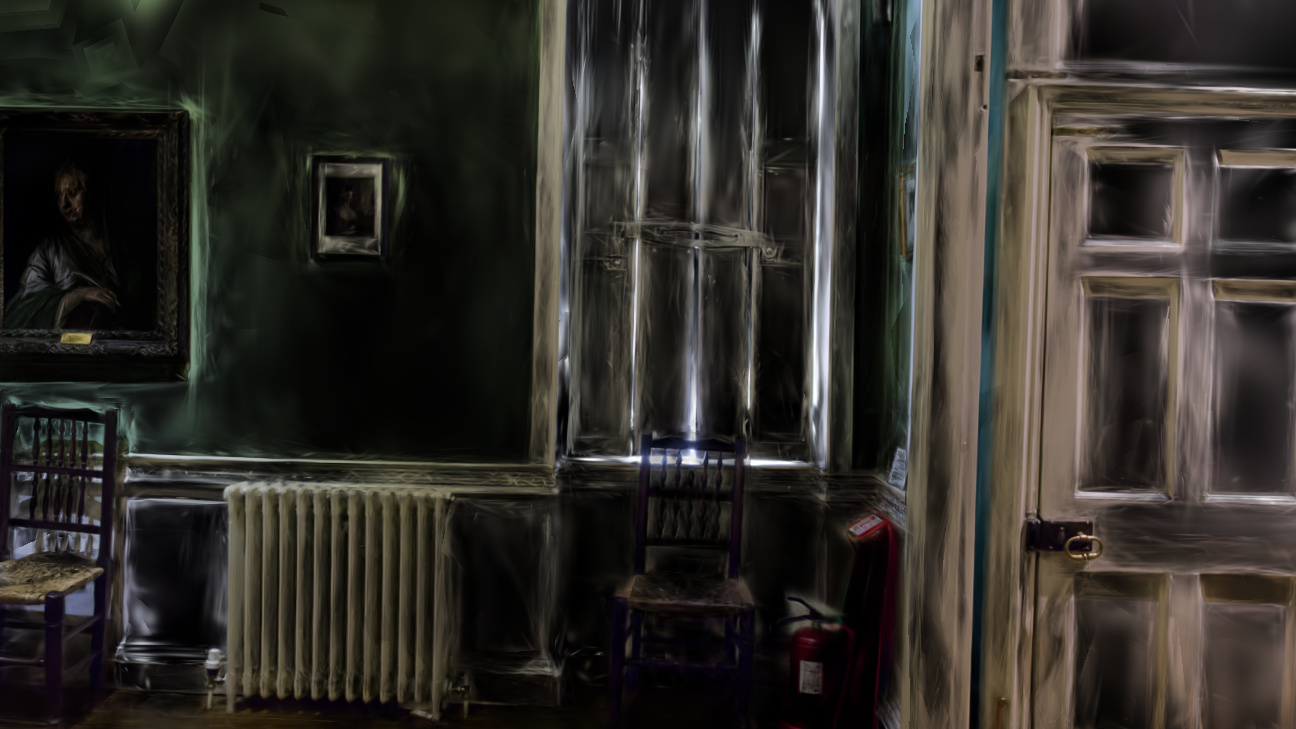}}
\end{tabular}
}{%
\fbox{\parbox[c][0.16\textheight][c]{0.90\linewidth}{\centering Missing view-consistency images. Copy the \texttt{consistency/} folder here.}}
}

\caption{\textbf{Per-scene component visualizations.}
The decomposition shows how the dissolve field suppresses unreliable mesh
evidence and how residual Gaussians refill the corresponding regions.
Here, ML-Hy denotes the ML-Hypersim-AI scene family.}
\label{fig:supp_component_visualization}
\end{figure*}

\begin{figure*}[t]
\centering
\newcommand{\vcenterbox}[1]{\raisebox{-0.5\height}{#1}}

\IfFileExists{consistency/hy/1_cropped.jpg}{%
\setlength{\tabcolsep}{1pt}\renewcommand{\arraystretch}{0.70}
\begin{tabular}{@{}ccccc@{}}
\vcenterbox{\rotatebox{90}{Scene 1}} &
\vcenterbox{\includegraphics[width=0.24\linewidth]{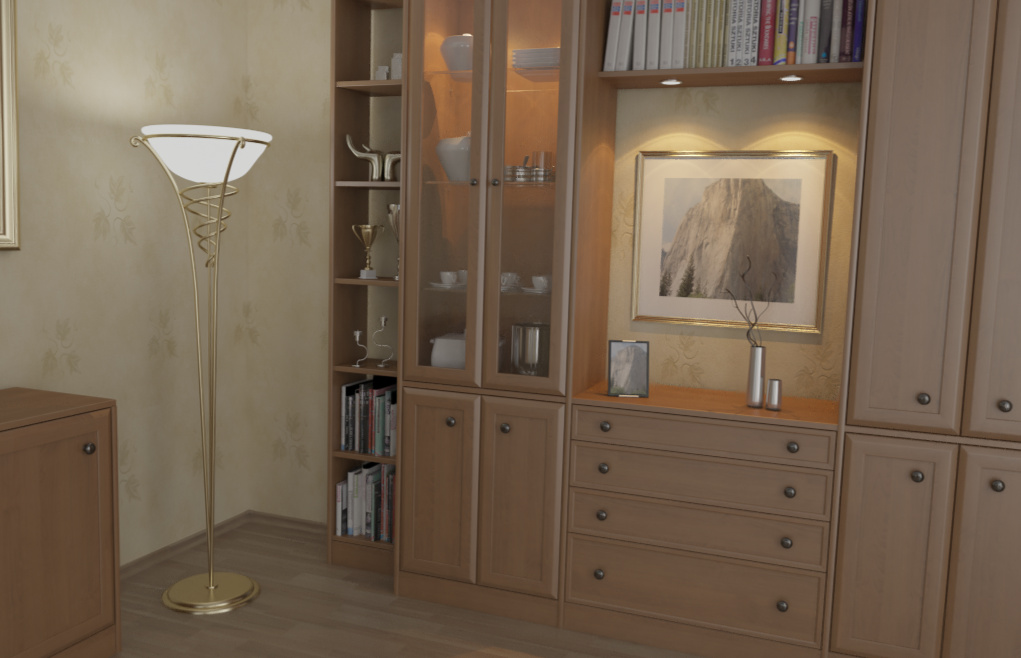}} &
\vcenterbox{\includegraphics[width=0.24\linewidth]{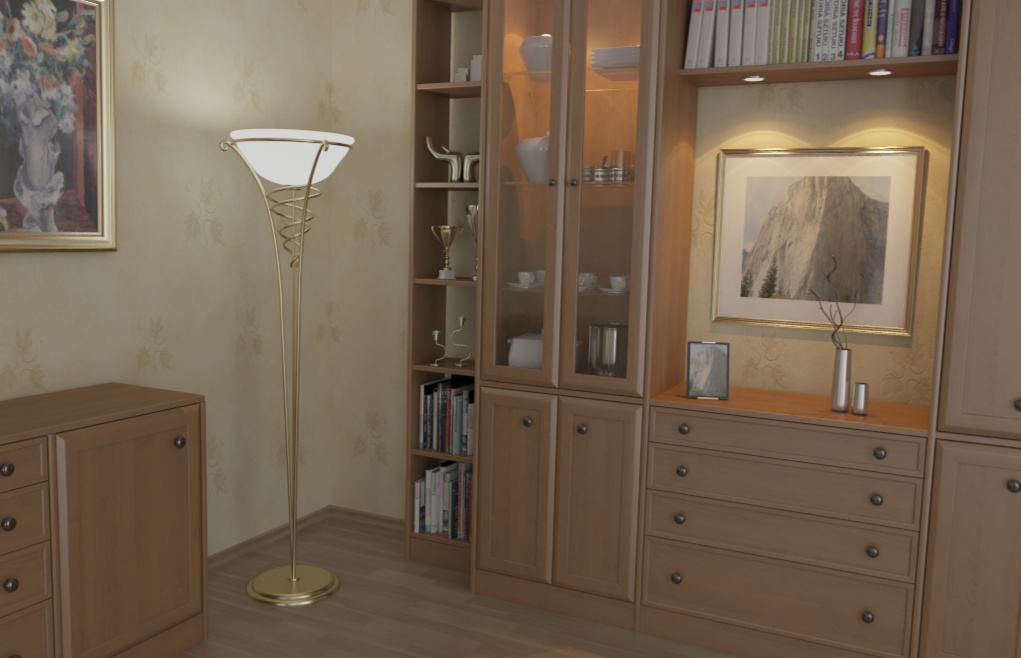}} &
\vcenterbox{\includegraphics[width=0.24\linewidth]{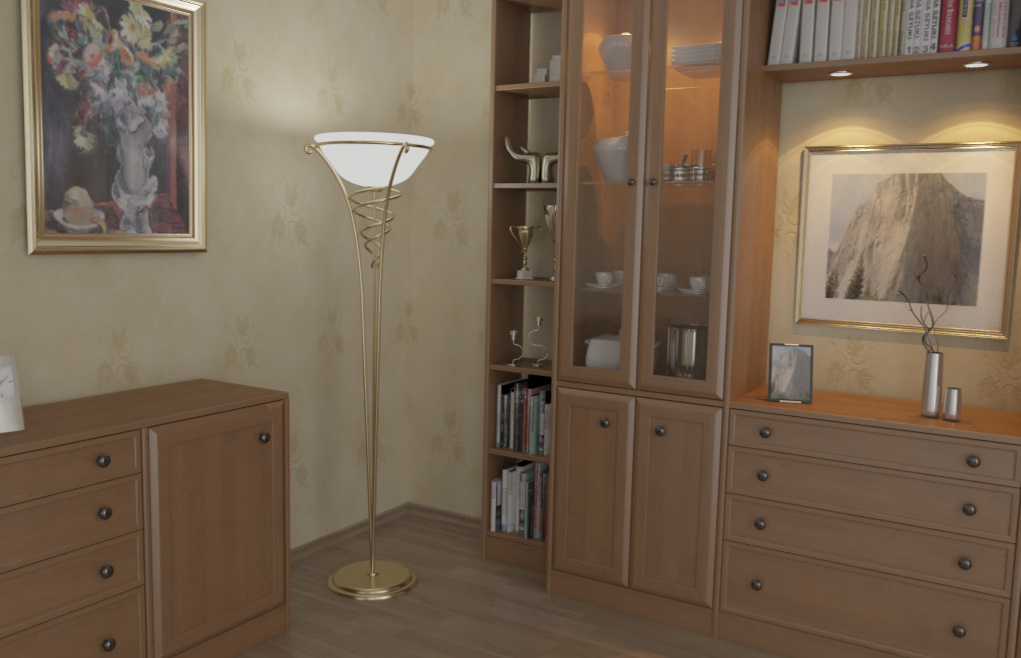}} &
\vcenterbox{\includegraphics[width=0.24\linewidth]{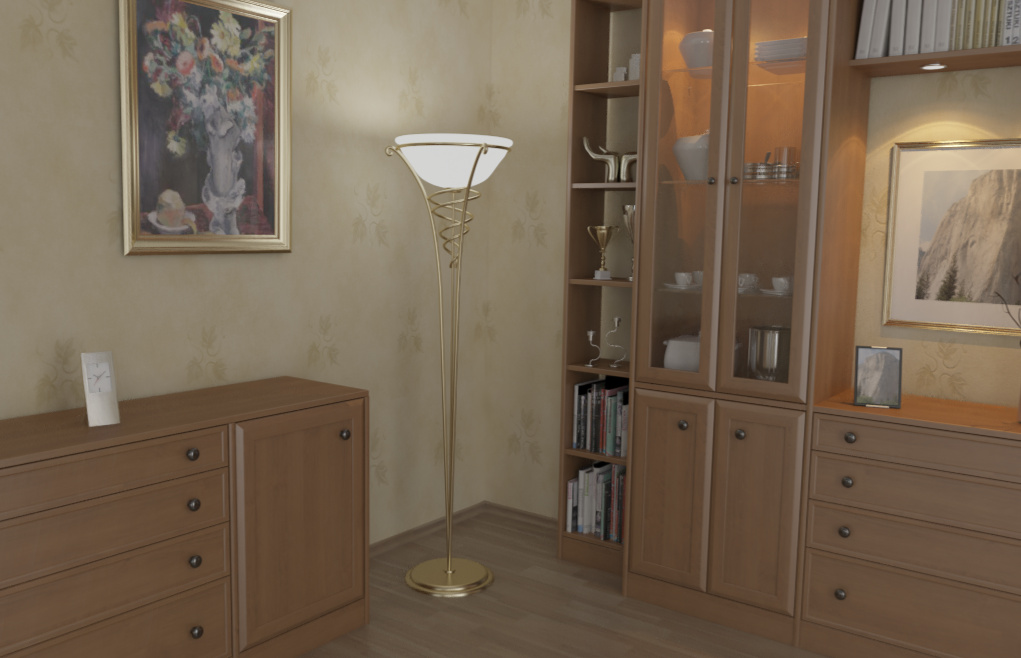}} \\[38pt] %

\vcenterbox{\rotatebox{90}{Scene 2}} &
\vcenterbox{\includegraphics[width=0.24\linewidth]{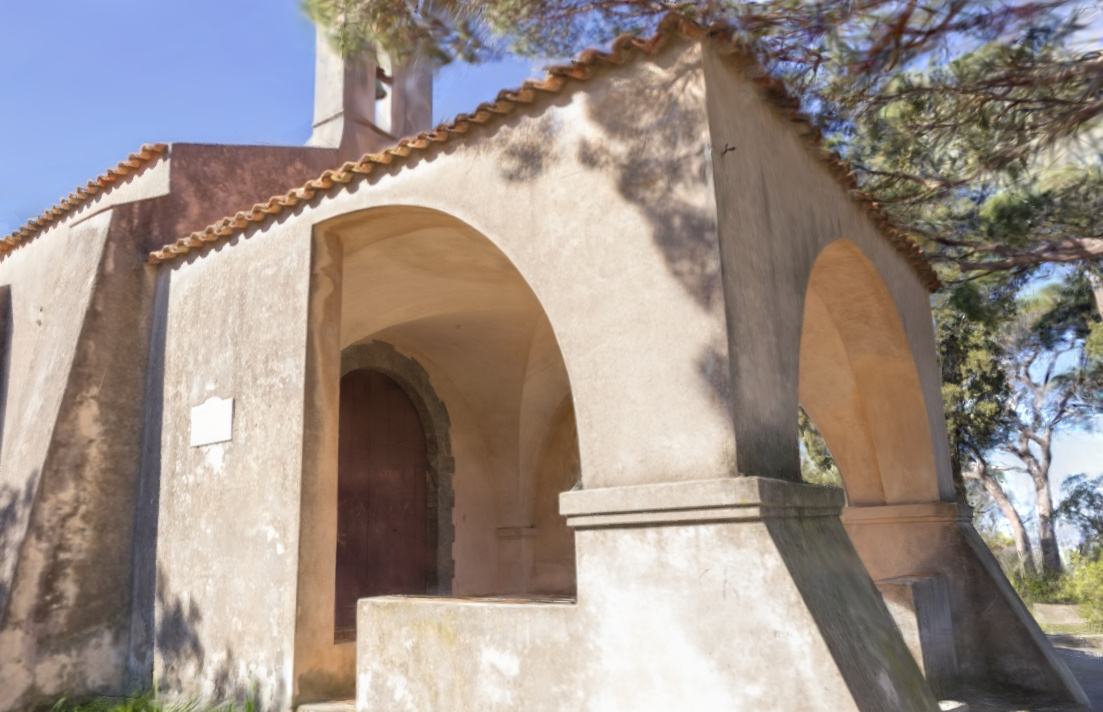}} &
\vcenterbox{\includegraphics[width=0.24\linewidth]{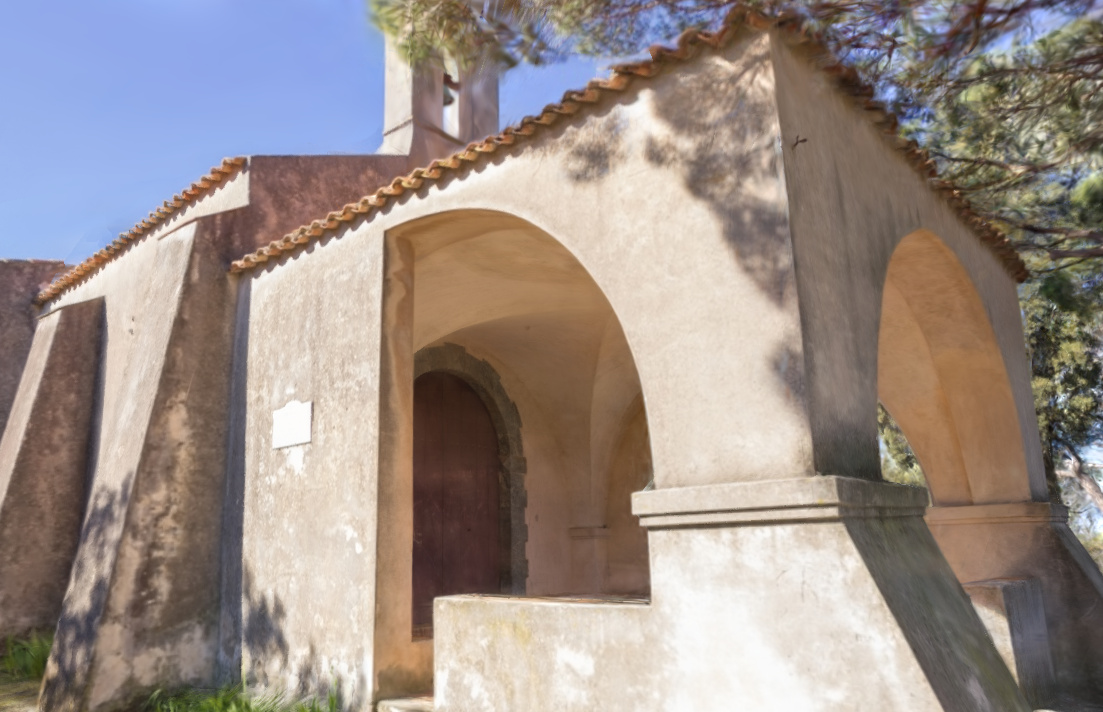}} &
\vcenterbox{\includegraphics[width=0.24\linewidth]{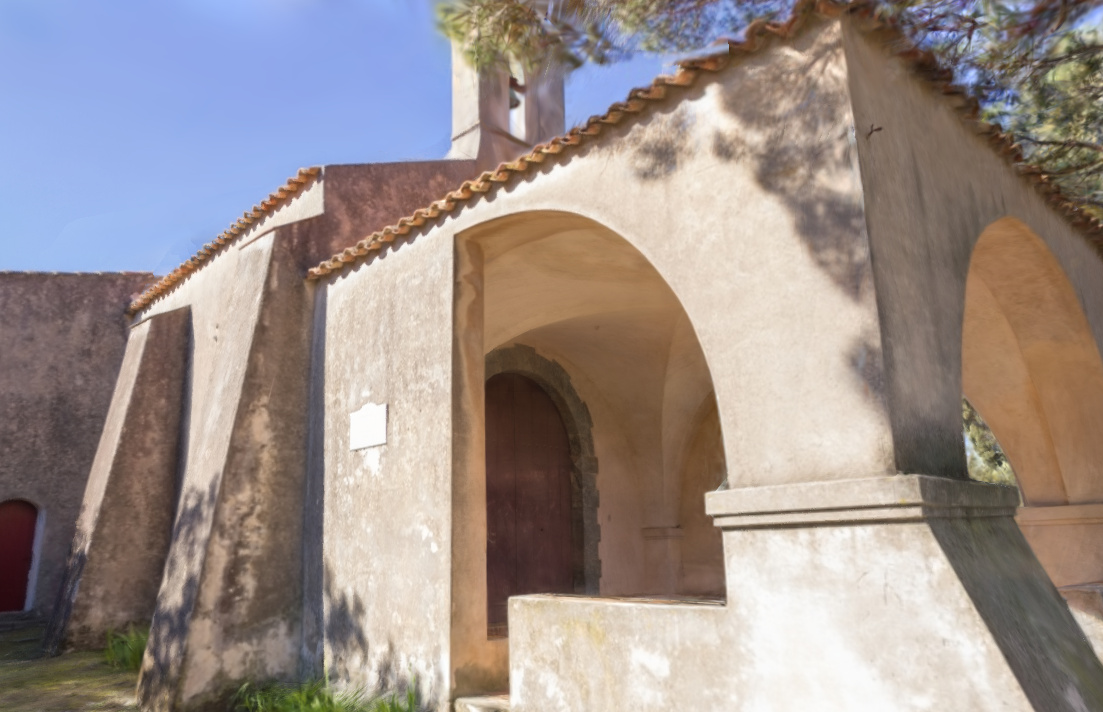}} &
\vcenterbox{\includegraphics[width=0.24\linewidth]{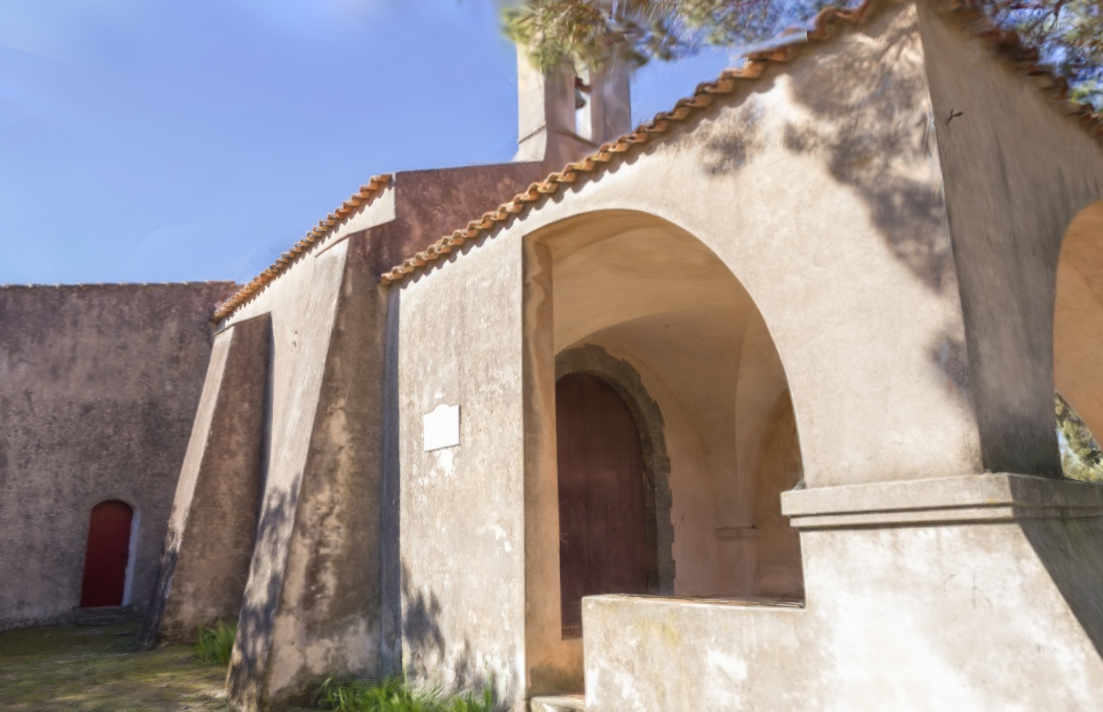}}
\end{tabular}
}{%
\fbox{\parbox[c][0.16\textheight][c]{0.90\linewidth}{\centering Missing view-consistency images. Copy the \texttt{consistency/} folder here.}}
}
\caption{\textbf{Additional adjacent-view consistency.}
Consecutive rendered crops remain stable across nearby views, including regions where residual Gaussians correct unreliable mesh evidence.}
\label{fig:supp_view_consistency}
\end{figure*}

\subsection{Loss-Driven Dual Densification}

The initial NDF captures mesh--image conflicts that are already visible before hybrid
training. However, additional conflicts may emerge as the hybrid
representation evolves. We therefore periodically render the current fused
result for all training views and compute its per-pixel mean absolute RGB
error. Pixels with negligible errors are discarded.

For a training view $v$, let $R_v(p)$ denote the reconstruction error at
pixel $p$, and let $\Omega_v$ denote the set of pixels whose errors exceed
the selection threshold. We sample pixels from an error-proportional categorical distribution,

\begin{equation}
P_v(p)=
\frac{R_v(p)}
{\sum_{p'\in\Omega_v}R_v(p')},
\qquad p\in\Omega_v,
\end{equation}

Thus, pixels with larger reconstruction errors have a higher probability of
being selected.

For each sampled pixel, we use the rasterized mesh depth to obtain a
mesh-guided 3D anchor. Let $D_m(p)$ denote the mesh depth at pixel $p$ and
$\operatorname{BackProj}_v(p,D_m(p))$ denote its back-projection in view
$v$. The anchor position is

\begin{equation}
\mathbf{x}(p)=
\operatorname{BackProj}_v\left(p,D_m(p)\right).
\end{equation}

At this position, we initialize one dissolve candidate and one residual RGB
candidate. The dissolve candidate increases the capacity to attenuate
unreliable mesh evidence, while the residual candidate increases the
capacity to reconstruct the content revealed after dissolution.

Candidates collected from different training views are first retained based on visibility across multiple
distinct cameras, and then residual candidates are merged using the 3D voxel aggregation described above. Retained residual candidates are initialized from the corresponding
ground-truth colors, whereas dissolve candidates start from a small scalar
dissolve value. Although both candidates share the same initial
mesh-guided anchor, they are optimized independently and are not constrained
to remain on the mesh surface.

\begin{table*}[t]
\centering
\scriptsize
\setlength{\tabcolsep}{2.1pt}
\renewcommand{\arraystretch}{1.08}
\caption{\textbf{Quantitative breakdown of hybrid methods on the Deep Blending dataset.}
We select four representative scenes encompassing both complex indoor layouts and structured outdoor architectural environments. 
\textbf{Bold} indicates the best performance, and cell colors highlight the \colorbox{best}{top}, \colorbox{second}{second-best}, and \colorbox{third}{third-best} results. $^{\dagger}$ Indicates catastrophic geometry collapse.}
\label{tab:deep_blending_breakdown}

\begin{tabularx}{\textwidth}{@{}l|cc|YYYY|YYYY|YYYY|YYYY@{}}
\toprule
\multirow{2}{*}{Method}
& \multirow{2}{*}{Mesh} 
& \multirow{2}{*}{Hybrid}
& \multicolumn{4}{c|}{Indoor: DrJohnson} 
& \multicolumn{4}{c|}{Indoor: Playroom} 
& \multicolumn{4}{c|}{Outdoor: SaintAnne} 
& \multicolumn{4}{c}{Outdoor: Bridge} \\
\cmidrule(lr){4-7}\cmidrule(lr){8-11}\cmidrule(lr){12-15}\cmidrule(lr){16-19}
& & 
& PSNR$\uparrow$ & SSIM$\uparrow$ & LPIPS$\downarrow$ & $\Delta$GS$\uparrow$ 
& PSNR$\uparrow$ & SSIM$\uparrow$ & LPIPS$\downarrow$ & $\Delta$GS$\uparrow$ 
& PSNR$\uparrow$ & SSIM$\uparrow$ & LPIPS$\downarrow$ & $\Delta$GS$\uparrow$ 
& PSNR$\uparrow$ & SSIM$\uparrow$ & LPIPS$\downarrow$ & $\Delta$GS$\uparrow$ \\
\midrule

3DGS 
& $\times$ & $\times$
& 29.33 & \cellcolor{best}\textbf{0.897} & 0.271 & 0\% 
& 30.62 & \cellcolor{best}\textbf{0.897} & 0.266 & 0\% 
& 25.54 & 0.838 & 0.280 & 0\% 
& \cellcolor{second}26.44 & \cellcolor{third}0.854 & 0.260 & 0\% \\

Reduced 3DGS 
& $\times$ & $\times$
& 29.30 & 0.894 & 0.276 & 53\% 
& 30.50 & \cellcolor{third}0.896 & 0.266 & \cellcolor{third}61\% 
& 25.50 & 0.835 & 0.287 & \cellcolor{second}39\% 
& 25.59 & 0.837 & 0.286 & \cellcolor{third}59\% \\

3DGS-MCMC 
& $\times$ & $\times$
& 29.40 & 0.889 & 0.274 & 1\% 
& 30.65 & 0.896 & 0.252 & -36\% 
& \cellcolor{second}25.62 & 0.840 & 0.279 & 0\% 
& \cellcolor{best}\textbf{26.46} & \cellcolor{best}\textbf{0.856} & 0.257 & 0\% \\

\midrule

SuGaR 
& $\checkmark$ & $\times$
& 28.30 & 0.870 & 0.326 & \cellcolor{second}59\% 
& 30.25 & 0.884 & 0.289 & 37\% 
& 24.20 & 0.820 & 0.301 & -38\% 
& 24.50 & 0.791 & 0.320 & 5\% \\

PGSR 
& $\checkmark$ & $\times$
& 29.25 & \cellcolor{third}0.895 & 0.273 & 37\% 
& 30.65 & 0.868 & 0.318 & 28\% 
& 25.42 & 0.827 & 0.315 & 19\% 
& 25.46 & 0.834 & 0.291 & 44\% \\

MeshSplatting 
& $\checkmark$ & $\times$
& 27.28 & 0.837 & 0.367 & -- 
& 24.17 & 0.808 & 0.365 & -- 
& 22.69 & 0.780 & 0.346 & -- 
& 22.79 & 0.748 & 0.395 & -- \\

\midrule

3D-GES 
& $\times$ & $\checkmark$
& 29.42 & 0.891 & 0.262 & 54\% 
& 30.66 & 0.895 & 0.253 & 45\% 
& 25.18 & 0.839 & 0.270 & \cellcolor{third}23\% 
& 11.75$^{\dagger}$ & 0.521$^{\dagger}$ & 0.608$^{\dagger}$ & -- \\

\midrule

Hard-Z Proxy 
& $\checkmark$ & $\checkmark$
& 29.15 & 0.891 & 0.244 & 49\% 
& 30.55 & 0.890 & \cellcolor{third}0.226 & 49\% 
& 25.27 & 0.836 & 0.231 & 20\% 
& 25.97 & 0.847 & 0.241 & \cellcolor{second}59\% \\

\rowcolor{gray!10}
\textbf{Ours (3DGS)} 
& $\checkmark$ & $\checkmark$
& \cellcolor{second}29.69 & 0.894 & \cellcolor{best}\textbf{0.240} & \cellcolor{third}58\% 
& \cellcolor{third}30.93 & 0.893 & \cellcolor{second}0.223 & \cellcolor{second}61\% 
& 25.61 & \cellcolor{third}0.842 & \cellcolor{best}\textbf{0.221} & 21\% 
& 26.23 & \cellcolor{second}0.854 & \cellcolor{best}\textbf{0.234} & 58\% \\

\rowcolor{gray!10}
\textbf{Ours (Reduced 3DGS)} 
& $\checkmark$ & $\checkmark$
& \cellcolor{third}29.60 & 0.895 & \cellcolor{third}0.243 & \cellcolor{best}\textbf{74\%} 
& \cellcolor{second}30.81 & 0.892 & 0.229 & \cellcolor{best}\textbf{76\%} 
& \cellcolor{third}25.62 & \cellcolor{second}0.843 & \cellcolor{third}0.224 & \cellcolor{best}\textbf{45\%} 
& 26.29 & 0.850 & \cellcolor{second}0.236 & \cellcolor{best}\textbf{76\%} \\

\rowcolor{gray!10}
\textbf{Ours (3DGS-MCMC)} 
& $\checkmark$ & $\checkmark$
& \cellcolor{best}\textbf{29.95} & \cellcolor{second}0.896 & \cellcolor{second}0.241 & -2\% 
& \cellcolor{best}\textbf{31.10} & \cellcolor{second}0.897 & \cellcolor{best}\textbf{0.218} & -3\% 
& \cellcolor{best}\textbf{25.73} & \cellcolor{best}\textbf{0.847} & \cellcolor{second}0.222 & -5\% 
& \cellcolor{third}26.39 & 0.850 & \cellcolor{third}0.238 & -6\% \\

\bottomrule
\end{tabularx}
\end{table*}

\begin{table*}[t]
\centering
\scriptsize
\setlength{\tabcolsep}{2.1pt}
\renewcommand{\arraystretch}{1.08}
\caption{\textbf{Irregular-scene stress test on Mip-NeRF 360.}
We report results on four outdoor scenes with complex geometry, vegetation, and weakly structured regions.
\textbf{Bold} indicates the best performance, and cell colors highlight the \colorbox{best}{top} and \colorbox{second}{second-best}.}
\label{tab:mipnerf360_stress}

\begin{tabularx}{\textwidth}{@{}p{2.4cm}|cc|YYYY|YYYY|YYYY|YYYY@{}}
\toprule
\multirow{2}{*}{Method}
& \multirow{2}{*}{Mesh} 
& \multirow{2}{*}{Hybrid}
& \multicolumn{4}{c|}{Outdoor: Treehill} 
& \multicolumn{4}{c|}{Outdoor: Garden} 
& \multicolumn{4}{c|}{Outdoor: Flowers} 
& \multicolumn{4}{c}{Outdoor: Bicycle} \\
\cmidrule(lr){4-7}\cmidrule(lr){8-11}\cmidrule(lr){12-15}\cmidrule(lr){16-19}
& & 
& PSNR$\uparrow$ & SSIM$\uparrow$ & LPIPS$\downarrow$ & $\Delta$GS$\uparrow$ 
& PSNR$\uparrow$ & SSIM$\uparrow$ & LPIPS$\downarrow$ & $\Delta$GS$\uparrow$ 
& PSNR$\uparrow$ & SSIM$\uparrow$ & LPIPS$\downarrow$ & $\Delta$GS$\uparrow$ 
& PSNR$\uparrow$ & SSIM$\uparrow$ & LPIPS$\downarrow$ & $\Delta$GS$\uparrow$ \\
\midrule

3DGS
& $\times$ & $\times$
& \cellcolor{second}20.42
& \cellcolor{second}0.684
& \cellcolor{second}0.258
& \cellcolor{second}0\%
& \cellcolor{second}28.55
& \cellcolor{second}0.883
& 0.123
& 0\%
& \cellcolor{best}\textbf{22.48}
& \cellcolor{best}\textbf{0.641}
& \cellcolor{second}0.314
& 0\%
& \cellcolor{best}\textbf{26.24}
& \cellcolor{best}\textbf{0.809}
& \cellcolor{second}0.187
& 0\% \\

\midrule

Hard-Z Proxy
& $\checkmark$ & $\checkmark$
& 19.21
& 0.655
& 0.290
& -1\%
& 27.41
& 0.866
& \cellcolor{second}0.119
& \cellcolor{best}\textbf{7\%}
& 20.77
& 0.573
& 0.334
& \cellcolor{second}5\%
& 24.33
& 0.764
& 0.211
& \cellcolor{second}13\% \\

\rowcolor{gray!10}
\textbf{Ours (3DGS)}
& $\checkmark$ & $\checkmark$
& \cellcolor{best}\textbf{20.58}
& \cellcolor{best}\textbf{0.689}
& \cellcolor{best}\textbf{0.255}
& \cellcolor{best}\textbf{14\%}
& \cellcolor{best}\textbf{29.00}
& \cellcolor{best}\textbf{0.897}
& \cellcolor{best}\textbf{0.093}
& \cellcolor{second}5\%
& \cellcolor{second}22.34
& \cellcolor{second}0.635
& \cellcolor{best}\textbf{0.295}
& \cellcolor{best}\textbf{13\%}
& \cellcolor{second}26.21
& \cellcolor{second}0.808
& \cellcolor{best}\textbf{0.175}
& \cellcolor{best}\textbf{17\%} \\

\bottomrule
\end{tabularx}
\end{table*}

\section{Additional Implementation Details}
\subsection{Training and Reproduction Settings}

All locally reproduced methods use the same calibrated cameras, train/test
splits, input resolution, and scene coordinate system. Unless otherwise
specified, training is performed for 30k iterations on a single NVIDIA RTX
5000 Ada GPU. Test views are not used for mesh reconstruction, texturing,
hyperparameter selection, or training.

For each baseline, we retain its official training objective and native
Gaussian update strategy. The Gaussolve variants use the same mesh, NDF initialization,
continuous fusion, and training schedule; they differ only in the update
strategy of the residual Gaussian backbone, as described in Sec.~\ref{sec:supp_plugin_backbones}. Exact
commands and configuration files are included in the supplementary code.

The MeshGS and Hybrid Mesh-Gaussian results in the main paper are taken
directly from their respective publications and are not included among our
local reproductions.

\paragraph{MeshSplatting reproduction.}
The official MeshSplatting implementation constructs its mesh through a
Restricted Delaunay (RDel) stage. Its topology code assumes that each
triangular face is shared by at most two tetrahedra. On several large scenes,
the raw Delaunay output contained duplicate, degenerate, or non-manifold
tetrahedra that violated this assumption and caused the official code to
terminate.

For these scenes, we used a deterministic robustness fix before the existing
RDel topology construction. We isotropically normalize the input coordinates,
canonicalize and deduplicate the generated tetrahedra, remove degenerate
tetrahedra, and retain at most the two highest-quality incident tetrahedra for
each remaining non-manifold face. The repair is accepted only when the removed
fraction is below $0.1\%$ and the final face incidence is at most two. All
subsequent MeshSplatting training and rendering stages remain unchanged.
Results produced with this fix are identified as the robust-RDel variant.

\subsection{Plug-in Gaussian Backbones}
\label{sec:supp_plugin_backbones}

\paragraph{3DGS.}
The default variant uses a 3DGS checkpoint obtained after the Gaussian-only
warm-up. At the hybrid handoff, we retain the Gaussian positions, scales, and
rotations, while resetting their color coefficients and opacities. The
residual field then continues with the standard 3DGS clone, split, and
opacity-based pruning operations. The NDF uses an independent scalar-Gaussian
optimizer and densification process.

\paragraph{3DGS-MCMC.}
For the MCMC variant, the Gaussian-only warm-up is performed using the
official 3DGS-MCMC update strategy. Unlike the default 3DGS handoff, the MCMC Gaussian colors and
opacities are retained because they determine the current particle
distribution and relocation probabilities.

During hybrid optimization, the MCMC operations are applied only to the
residual Gaussian field. Low-opacity residual Gaussians are relocated to
samples drawn from the alive population, their opacity and scale are adjusted
using the original MCMC relocation rule, and new residual Gaussians are added
until the prescribed scene capacity is reached. The original
opacity-dependent SGLD perturbation is also retained. 

The NDF, the textured mesh and continuous fusion remain unchanged.

Because 3DGS-MCMC does not use the opacity-based pruning mechanism of
standard 3DGS, we retain its original update strategy, as our goal is to
evaluate whether the fusion framework improves the backbone without altering
its native optimization behavior. Consequently, this variant can yield
slightly negative $\Delta$GS, as reported in the main paper.

\paragraph{Reduced 3DGS.}
For the Reduced 3DGS variant, the Gaussian-only warm-up is performed using
the official Reduced 3DGS update strategy. Because the Gaussolve renderer expects
a standard SH tensor layout, the grouped Reduced 3DGS checkpoint is converted
to an equivalent full-SH representation at the hybrid handoff. This
conversion changes only the storage layout and preserves the Gaussian
geometry and appearance.

During hybrid optimization, the redundancy-aware mercy pruning and
dead-point removal of Reduced 3DGS are applied only to the residual Gaussian
field. Moreover, we retain its
non-DC SH sparsity regularization. At the end of the densification window,
multi-view SH-band culling determines the lowest sufficient SH degree for
each residual Gaussian and removes the unused higher-order coefficients.

The NDF, the textured mesh and continuous fusion remain unchanged.

Compression codebooks and entropy coding are offline storage mechanisms and
are not required for the plug-in training comparison. We therefore transfer
the backbone's Gaussian initialization and population-reduction strategy,
while storing the final hybrid fields in the common representation used by
our renderer.

\section{Surface-Based Dissolve Alternatives}

A natural simplification of the \ndf{} is to directly optimize opacity on the mesh surface. We study two such alternatives: per-vertex alpha and texture-map alpha. Per-vertex alpha is constrained by mesh tessellation and has difficulty isolating thin or localized failure regions. Texture-map alpha is less constrained by vertex count, but the learned pattern is often diffuse and fragmented because texture-space errors do not correspond cleanly to view-space photometric conflicts. By contrast, the volumetric \ndf{} is localized in 3D and can be view-dependent, allowing the residual Gaussian branch to take over after unreliable mesh evidence is removed.

\subsection{Additional Component Visualizations}

Fig.~\ref{fig:supp_component_visualization} visualizes the
mesh, residual, and dissolve components across representative structured
and semi-structured scenes. The dissolve field suppresses unreliable mesh
evidence, while the residual branch reconstructs the revealed content.

\section{Additional View Consistency}

Fig.~\ref{fig:supp_view_consistency} shows adjacent-view crops under small viewpoint changes. The rendered appearance remains stable around mesh-image conflict regions, indicating that the dissolve-and-refill decomposition does not introduce noticeable view-to-view flicker.

\section{Limitations on Irregular Scenes}
\label{sec:supp_limitations}

\gaussolve{} is designed for scenes where a lightweight textured mesh can
replace broad structured regions, while residual Gaussians handle complex
geometry and appearance details. In scenes dominated by foliage, thin
branches, and severe geometric ambiguity, the extracted mesh may be too
inaccurate to provide a reliable surface prior. The mesh and residual
branches may then compete to explain the same image evidence, limiting the
reduction in Gaussian primitives.

This defines the intended operating regime of our method: its largest
compactness gains arise when broad regular regions can be represented
efficiently by the textured mesh. Geometry-aware routing for highly
irregular regions is an interesting direction for future work.

\end{document}